\documentclass{article} %

\usepackage{iclr2027_conference,times}

\usepackage{amsmath,amsfonts,bm}

\def\eqref#1{equation~\ref{#1}}

\def\1{\bm{1}}

\DeclareMathAlphabet{\mathsfit}{\encodingdefault}{\sfdefault}{m}{sl}
\SetMathAlphabet{\mathsfit}{bold}{\encodingdefault}{\sfdefault}{bx}{n}

\usepackage{hyperref}
\usepackage{url}
\usepackage{microtype}
\usepackage{graphicx}
\usepackage{subcaption}
\usepackage{booktabs}
\usepackage{amsmath}

\usepackage{amssymb}
\usepackage{mathtools}
\usepackage{amsthm}
\usepackage[capitalize,noabbrev]{cleveref}
\usepackage[textsize=tiny]{todonotes}

\usepackage{multirow}
\usepackage{pifont}   %
\usepackage{xcolor} 
\usepackage{tabularx}

\newcommand{\cmark}{\textcolor{green!60!black}{\textbf{O}}}

\newcommand{\xmark}{\textcolor{red}{\ding{55}}}
\newcommand{\header}[2]{\begin{tabular}[b]{@{}c@{}}\textbf{#1}\\\textbf{#2}\end{tabular}}
\newcommand{\modelname}{LIF-Stream}

\usepackage{float}
\usepackage{algorithm}
\usepackage{algorithmic}

\makeatletter
\let\ALC@orig@listi\@listi
\def\@listi{\ALC@orig@listi \itemsep\z@ \parsep\z@ \topsep\z@ \partopsep\z@}
\let\@listii\@listi \let\@listiii\@listi
\let\@listiv\@listi \let\@listv\@listi \let\@listvi\@listi
\makeatother

\theoremstyle{plain}
\newtheorem{theorem}{Theorem}[section]
\newtheorem{proposition}[theorem]{Proposition}
\newtheorem{lemma}[theorem]{Lemma}
\newtheorem{corollary}[theorem]{Corollary}

\theoremstyle{definition}

\theoremstyle{remark}
\newtheorem{remark}[theorem]{Remark}

\title{Spiking Neural Networks Incrementally Track Firing-Rate Networks}

\author{Hyunseok Oh \& Youngki Lee\\
Department of Computer Science and Engineering\\
Seoul National University\\
1 Gwanak-ro, Gwanak-gu, Seoul 08826, Republic of Korea\\
\texttt{\{ohsai,youngki.lee\}@snu.ac.kr} 
}

\iclrfinalcopy %
\begin{document}

\maketitle

\begin{abstract}
Understanding how spiking neural networks (SNNs) efficiently process data streams is key to (i) studying how biological brains conserve energy for lifelong predictions and (ii) enabling efficient continuous AI inference. In this work, we propose a theory on how an SNN incrementally tracks its corresponding firing-rate network, a continuous-valued neural network (ANN). We prove that the output of a discrete SNN continuously tracks the instantaneous ANN output implicitly determined by the SNN input. 
Based on this principle, we develop \modelname{}, a conversion framework for efficient SNN inference over unbounded data streams. It enables the converted SNN to reuse its prior activation state and compute only the activation updates between frames, thereby reducing the average number of time steps needed per frame. Specifically, \modelname{} controls the SNN to continuously track the reference ANN output, which gives a high-accuracy prediction for every frame. 
Evaluation on three large-scale streaming inference benchmarks with distinct modalities shows that \modelname{} achieves the best efficiency-accuracy trade-off compared to high-accuracy SNN techniques.
Code will be released at \url{https://github.com/snuhcs/LIF-Stream}.
\end{abstract}

\section{Introduction}

Understanding how biological brains efficiently process continuous sensory streams throughout their lifetimes can advance the design of efficient, always-on AI systems.
Brains save metabolic energy on static scenes while selectively scaling cognitive compute on surprising events~\citep{attwell2001energy,friston2010free}. 
Spiking neural networks (SNNs) computationally model these biological neural systems to (i) study their underlying information processing mechanisms and (ii) implement extremely energy-efficient AI systems with neuromorphic hardware~\citep{roy2019towards}. SNNs are  networks of spiking neurons that mimic neuronal dynamics and communicate through binary spikes~\citep{Maass1996ThirdGen,gerstner2002spiking}. Since they fundamentally process the time series of binary events (spike trains) in a biophysical manner, they are naturally expected to process streaming data efficiently. However, existing high-accuracy SNN techniques, including direct SNN training and ANN-to-SNN conversion methods, overlook the need to leverage historical states to efficiently process new inputs, especially for unbounded streams.

We first derive a theoretical principle governing the transient dynamics of discrete SNNs, which underlies how an SNN can efficiently process information through time. It is challenging to analyze the discrete SNNs since they are discontinuous dynamical systems. We overcome this by extending the ANN-to-SNN conversion theory~\citep{rueckauer2016theory,xiao2021trainingfsnn,oh2024sign} to the transient regimes through the lens of reference tracking.
Specifically, we prove that SNN consisting of discrete LIF neurons (LIF-SNN) continuously attracts toward the instantaneous output of its corresponding continuous-valued clipped-ReLU network (ANN).
This ANN is a firing-rate network~\citep{gerstner2014neuronal}: its activations are the targets of the exponentially filtered firing rates of the SNN neurons.
Thus, the ANN output is a time-varying reference target that SNN dynamics tracks. LIF-SNN dynamics is a damped fixed-point iteration with a constant step size that permanently tracks this target. For static inputs, it oscillates within the small neighborhood of the static equilibrium. Although SNNs with discrete Integrate-and-Fire (IF) neurons match the ANN output exactly at steady state, they cannot guarantee tracking of a changing input, because their step sizes decay through time and their membranes accumulate charge debt (Sec.~\ref{sec:approach}).

Based on our theory, we introduce \emph{\modelname{}}, a conversion framework for efficient SNN inference over unbounded data streams, in which the SNN incrementally tracks its continuous-valued ANN counterpart across an unbounded input stream, producing a prediction for every frame.
Concretely, \modelname{} is a two-time-scale SNN control process: The outer loop continuously encodes the input stream to \emph{implicitly} drive the reference ANN for accurate frame-wise prediction. The inner loop allocates a few time steps per frame for the SNN to continuously track this reference ANN output, which is a time-varying target, while retaining its states across frame boundaries. The framework is a recipe into which any frame-wise ReLU1 ANN can be plugged, configured only by the leak $\tau$, the number of ticks (SNN time steps) per frame $T$, the input encoder, and an optional warm-up.

\modelname{} efficiently processes streams, since SNN retains prior activation values and incrementally computes to close the gap between ANN and SNN activations. For streams where consecutive frames are similar, SNNs need only a few ticks per frame on average to maintain high frame-wise accuracy. 
This finding challenges the common belief by showing that ANN-to-SNN conversion can be as efficient as direct training methods on streaming data without specialized training or fine-tuning. For such incremental computation to persist indefinitely, we argue that ANN-to-SNN conversion techniques should pivot from IF to LIF neuron. While the IF neuron's time-diminishing step size and charge debt prevent it from tracking continuously, the LIF neuron's constant step size ensures permanent and autonomous alertness.

We comprehensively evaluate our framework across diverse streaming inference tasks spanning vision, audio, and neuromorphic (event-stream) benchmarks. Comparisons with state-of-the-art SNN techniques show that \modelname{} generally achieves the best efficiency-accuracy trade-off. 
\modelname{} is significantly more tick-efficient than conversion techniques, while surpassing directly trained SNNs in peak accuracy with comparable low-tick accuracy.
Furthermore, \modelname{} is natively compatible with standard digital neuromorphic hardware out-of-the-box, as it operates without post-training model calibration, customized neuronal dynamics, or non-binary spike communication. In-depth analyses demonstrate that this tick efficiency stems from our incremental computing approach, which naturally exploits small inter-frame prediction deltas by computing incremental state updates rather than full activations.

\section{Computing Dynamics of Spiking Neural Networks}

\label{sec:computational_dynamics}
We study the standard Leaky Integrate-and-Fire (LIF) neuron model~\citep{gerstner2014neuronal}, which is widely researched and implemented in neuromorphic hardware.
The discrete-time dynamics of an LIF neuron with membrane potential $u(t)$ and spike output $s(t)$ consists of  three consecutive steps:
\begin{align}
u_{\text{pre}}(t) &= \bigg(1 - \frac{1}{\tau}\bigg)~u(t-1) + \frac{1}{\tau}~u_{\text{rest}} + \frac{R}{\tau}~I(t)\label{eq:lif_charge} & \text{(Integration)}\\
s(t) &= \mathbb{H}(u_{\text{pre}}(t) - \theta)\label{fire}& \text{(Thresholding)}\\
u(t) &= u_{\text{pre}}(t) - \theta~s(t)&\text{(Reset)}\label{reset}
\end{align}
Here, $u_{\text{pre}}(t)$ and $u_{\text{rest}}$ denote the pre-firing and resting potential,  $\mathbb{H}(\cdot)$ is the Heaviside step function, $\theta$ is the threshold, $\tau$ is the leakage factor, and $R$ is membrane resistance. We adopt a soft reset ( \eqref{reset}),  the  subtraction-based discretization of neuronal reset. We choose it instead of hard reset for its superior accuracy and analytical tractability~\citep{han2020deeptsc}.

\emph{Integrate-and-Fire (IF) neuron} further simplifies the integration dynamics by removing the leakage behavior through $\tau \to \infty, R = \tau$. 

We use two decoding schemes to map binary spike trains $s(t)$ into real-valued signals $y(t)$. First, the standard rate coding averages all spikes equally through time. 
Second, the exponential moving average (EMA) coding~\citep{oh2024sign} averages spikes with weights that exponentially decay through time. For analytical tractability, we align the decay factor of EMA coding with the leakage factor $\tau$ of the LIF neuron. 
Both admit streaming formulation, starting from $y(0) = 0$: 
\begin{equation}
y(t+1) = \frac{\tau - 1}{\tau} y(t) + \frac{1}{\tau} s(t)\quad \text{\big(EMA\big)} \qquad\qquad y(t+1) = \frac{t - 1}{t} y(t) + \frac{1}{t} s(t)\quad\text{\big(Rate\big)}  \label{eq:stream_exponentialcoding_lif}
\end{equation}

We consider a general fully connected network of $d$ neurons driven by $n$ input sources. An input matrix \(F \in \mathbb{R}^{d \times n}\) weighs input signals $x(t)$, and a recurrent matrix \(W \in \mathbb{R}^{d \times d}\) weighs previous spike outputs $s(t-1)$.
Assuming linear synapse, the total current $I(t) \in \mathbb{R}^d$ is given by:
\begin{align}
I(t) = Fx(t) + W s(t-1). \label{eq:current_lif}
\end{align}

\begin{figure*}[t]
\vspace{-0.5em}
\centering
\begin{subfigure}[b]{0.48\textwidth}
    \centering
    \includegraphics[width=\linewidth]{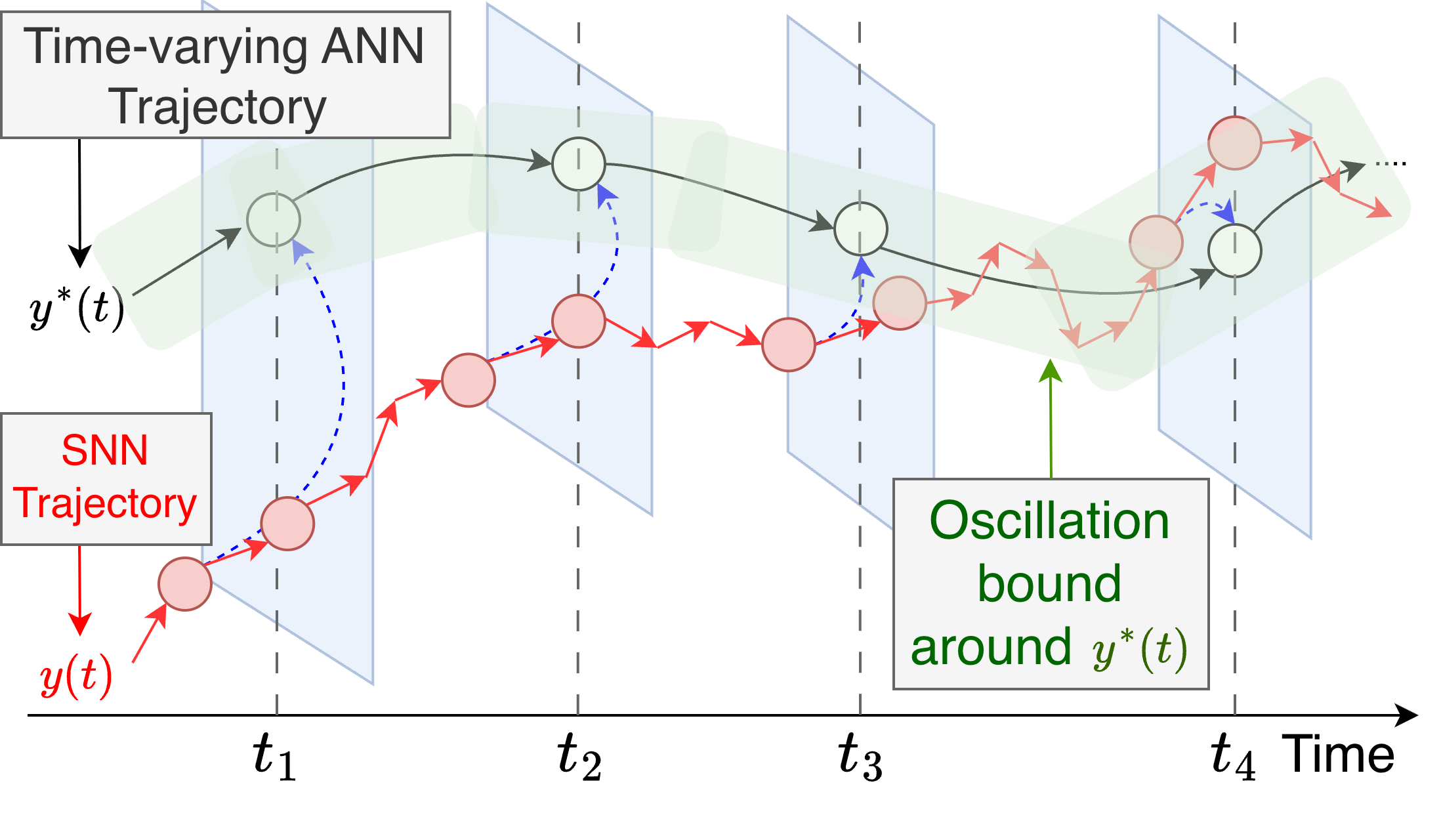}
    \caption{Conceptual overview of our theory: LIF-SNN dynamics constantly tracks time-varying ANN outputs.}
    \label{fig:sub_attraction}
\end{subfigure}
\hfill
\begin{subfigure}[b]{0.48\textwidth}
    \centering
    \includegraphics[width=\linewidth]{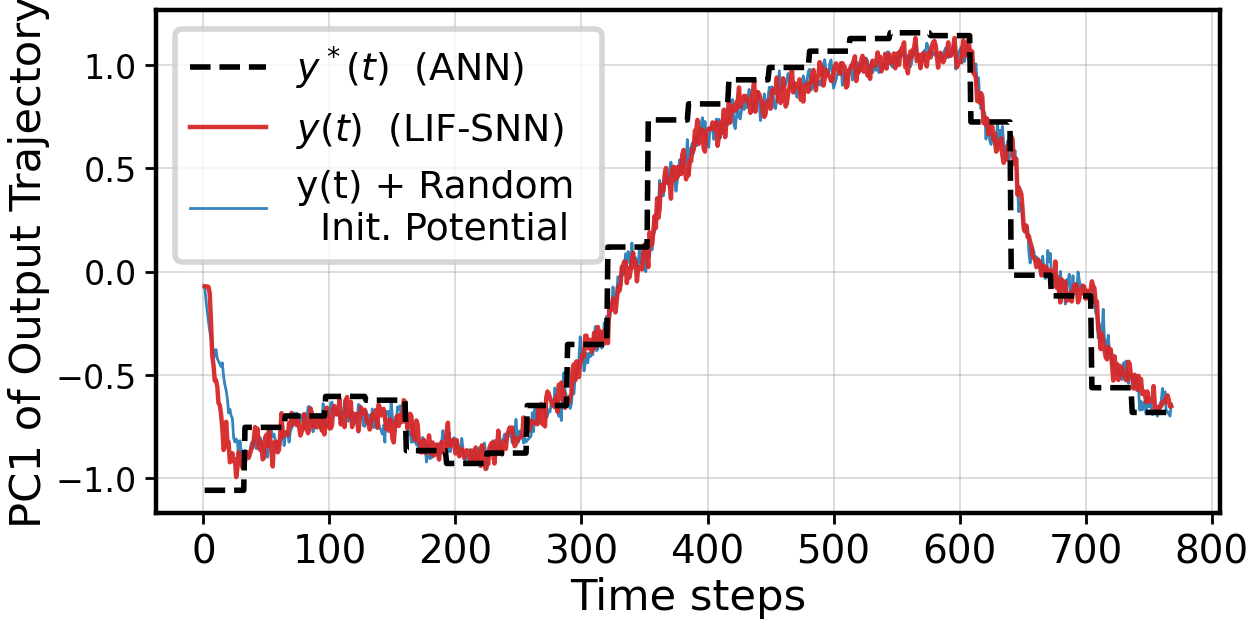}
    \caption{Empirical validation of our theory: Time-domain tracking of LIF-SNN towards its ANN Trajectory.}
    \label{fig:sub_static}
\end{subfigure}

\vspace{0.8em} %

\begin{subfigure}[b]{0.48\textwidth}
    \centering
    \includegraphics[width=\linewidth]{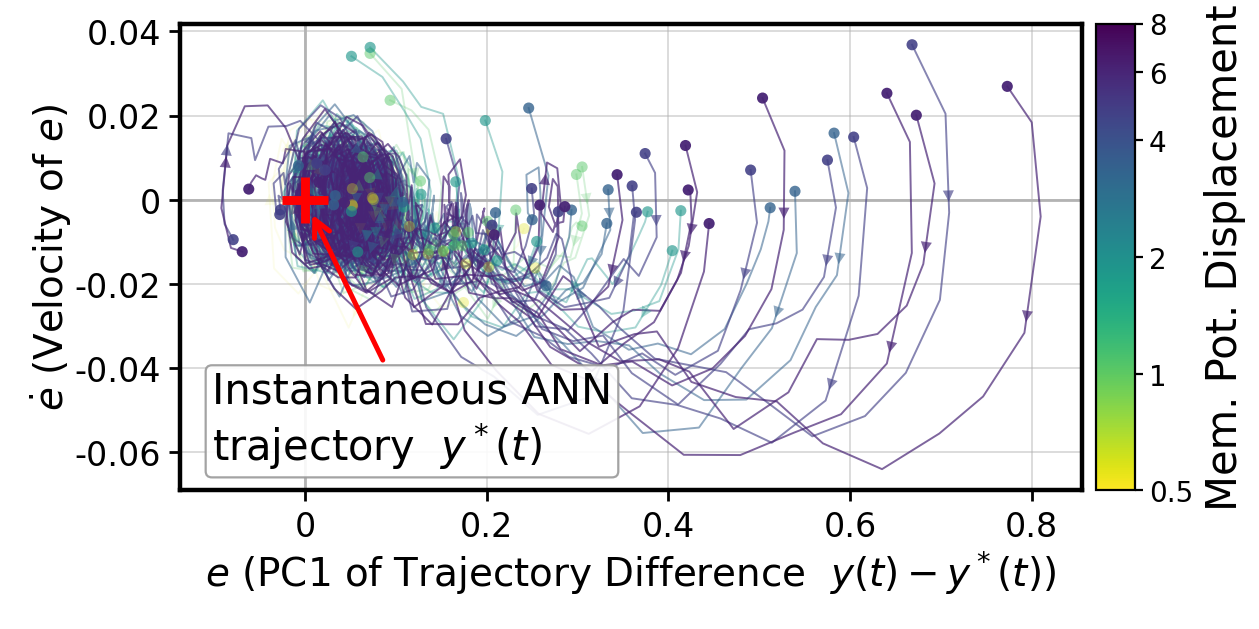}
    \caption{Multiple LIF-SNN trajectories starting from random initial conditions converge toward the time varying ANN trajectory given identical input sequences.}
    \label{fig:sub_dynamic}
\end{subfigure}
\hfill
\begin{subfigure}[b]{0.48\textwidth}
    \centering
    \includegraphics[width=\linewidth]{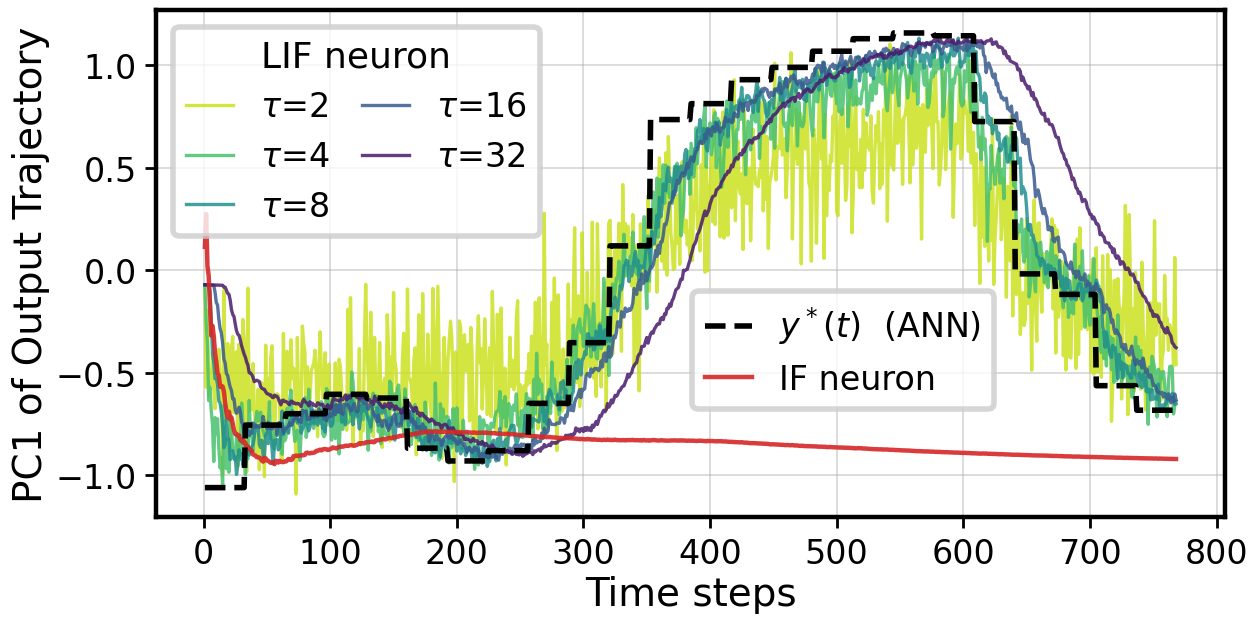}
    \caption{
    Impact of step size on SNN tracking dynamics. Unlike constant $1/\tau$ (LIF neuron), time-decaying step size  $1/(t+1)$ (IF neuron) fails to track continuously.}
    \label{fig:sub_fourth}
\end{subfigure}

\vspace{-0.5em}
\caption{
Conceptual overview and empirical validation of theoretical findings. (b) LIF-SNNs are continuously attracted to instantaneous ReLU1-ANN outputs, tracking the ANN trajectory through time. (c) Basin of attraction: trajectories starting from arbitrary initial states and histories converge toward the same target trajectory given identical inputs afterwards. (d) Trade-offs in step size ($\tau$): larger step sizes (smaller $\tau$) achieve faster convergence with larger oscillation, while smaller step sizes (larger $\tau$) reduce oscillation at the expense of convergence speed.
}
\label{fig:snn_trajectory_visualization}
\vspace{-1em}
\end{figure*}

\subsection{Theoretical Analysis}
The LIF-SNN dynamics in~\eqref{eq:lif_charge}-\ref{eq:current_lif} is a complex nonlinear system of difference equations that is difficult to analyze directly. We first show that these dynamics reduce to a single damped fixed-point iteration. 
Throughout the paper, we list complete proofs in Appendix~\ref{sec:proofs}.~~$ \mathbf{I}$ denotes the identity matrix, $\mathbf{1}$ the vector of ones, and $\text{ReLU1}: \mathbb{R} \to [0,1]$ defined as $\text{ReLU1}(x)= \min(\max(x,0),1)$.

\begin{lemma}
\label{thm:dynamics_lif} 
Let the input signals $x(t)$ and spike outputs $s(t)$ be EMA-coded into $\tilde{x}(t)$ and $y(t)$, respectively.
  The output trajectory $y(t)$ of a  discrete-time LIF-SNN exactly equals the trajectory of the damped fixed point iteration with a constant factor $\frac{1}{\tau}$:
\begin{equation}
y(t) =  \bigg(1 - \frac{1}{\tau} \bigg)~y(t-1) +  \frac{1}{\tau}~ \mathbb{H}\bigg(F~\tilde{x}(t)
 + W_0~y(t-1) + b(t) - \theta\cdot\mathbf{1}\bigg)  \label{eq:fixedpoint_iteration_y}
\end{equation}
where $\alpha = 1 - \frac{1}{\tau}$, $W_0 \coloneqq W - (\tau -1) \theta \mathbf{I}$, and $b(t) \coloneqq u_{\text{rest}}(1 - \alpha^{t})\mathbf{1} + \alpha^{t} u_{\text{pre}}(0)$. 
\end{lemma}
\begin{proof}[Proof sketch]
We eliminate the membrane potentials $u(t), u_{\text{pre}}(t)$ in terms of  $\tilde{x}(t)$ and $ y(t)$. 
Recursively unrolling the integration \eqref{eq:lif_charge} backwards to $t=0$ expresses $u_{\text{pre}}(t)$ as an exponentially weighted sum of $I(k)$. Substituting $I(k)$ using~\eqref{eq:current_lif} reduces $u_{\text{pre}}(t)$ to a linear combination of $y(t-1)$ and $\tilde{x}(t)$:
\begin{equation}
u_{\text{pre}}(t) = F\,\tilde{x}(t) + W_0\,y(t-1) + b(t), \label{eq:upre_closed_form}
\end{equation}
Plugging this~\eqref{eq:upre_closed_form} and the firing rule~\eqref{fire} into the coding definition \eqref{eq:stream_exponentialcoding_lif} yields the desired results  
(see complete proof in  Appendix~\ref{pf:dynamics_lif}).
\end{proof}
We call $\frac{1}{\tau}$ the \emph{step size} because~\eqref{eq:fixedpoint_iteration_y} is a relaxed fixed-point iteration, $y(t) = y(t-1) + \frac{1}{\tau}\big(\mathbb{H}(\cdot) - y(t-1)\big)$, whose coefficient sets how far $y$ moves toward its current target per tick; the same coefficient is the step size in stochastic approximation~\citep{robbins1951stochastic}.

Based on this, we derive our key finding governing the computing direction of transient SNN dynamics: the SNN output $y(t)$ continuously tracks its corresponding continuous-valued ReLU1 network \emph{instantaneously} determined by the input $\tilde{x}(t)$ at each time step $t$, as shown in  Figure~\ref{fig:sub_attraction}.

\begin{theorem}
\label{thm:equilibrium_lif}
For each time step $t$, the LIF-SNN output $y(t)$ is attracted with a step size $\frac{1}{\tau}$ toward the fixed-point set $\mathcal{E}(t)$ of a continuous-valued neural network, instantaneously  defined by $\tilde{x}(t)$:
\begin{equation}
    \mathcal{E}(t) \coloneqq \bigg\{y \in [0,1]^d~\Big|~y = \text{ReLU1}\Big(\tfrac{1}{(\tau-1)\theta}\big(F\tilde{x}(t+1) + Wy  + b(t+1) - \theta \cdot \mathbf{1}\big)\Big)\bigg\}
\end{equation}
The set $\mathcal{E}(t)$ is always nonempty because its ReLU1 map is a continuous self-map on the compact convex domain $[0,1]^d$, guaranteeing the existence of a fixed point by Brouwer's fixed-point theorem.
\end{theorem}
\vspace{-1em}
\begin{proof}[Proof sketch]
The update term~\eqref{eq:fixedpoint_iteration_y} can be rewritten as $y(t+1) = y(t) - \frac{1}{\tau}(y(t) - \mathbb{H}( D(t; y(t))-y(t)))$, where $D(t; y) \coloneqq \frac{1}{(\tau-1)\theta}\big(F\tilde{x}(t+1)+b(t+1)+Wy-\theta\cdot\mathbf{1}\big)$. A coordinate-wise analysis shows that each coordinate $[y(t)]_k$ always moves toward $\text{ReLU1}([D(t;y(t))]_k)$ with a constant step size $\frac{1}{\tau}$. This contracts the gap between $[y(t)]_k$ and $\text{ReLU1}([D(t;y(t))]_k)$ at a geometric rate $\alpha=1-\frac{1}{\tau}$, towards the set of points satisfying $y=\text{ReLU1}(D(t;y))$, which is  $\mathcal{E}(t)$. Hence, $y(t)$ is attracted towards the instantaneous fixed-point set $\mathcal{E}(t)$  regardless of the initial condition or prior history $y(<t)$. (See full proof in Appendix~\ref{pf:equilibrium_lif}.)
\end{proof}
\vspace{-0.5em}
$\mathcal{E}(t)$ can be identified as a Deep Equilibrium Model (DEQ)~\citep{bai2019deq}, which defines the output as the fixed point of a single transformation driven by input. A single-layer DEQ is equivalent to infinite-depth, weight-tied feedforward network, and thus can represent any finite-depth network.

The constant step size makes $y(t)$ continuously pursue the moving target $\mathcal{E}(t)$, but also prevents exact convergence. Using Theorem~\ref{thm:equilibrium_lif}, we can derive a stronger version of ANN-to-SNN conversion than prior work~\citep{xiao2021trainingfsnn}: When input stabilizes and weight is contractive, 
$y(t)$ converges geometrically into an $\mathcal{O}(\frac{1}{\tau})$-neighborhood of the static equilibrium and oscillates within it thereafter.

\begin{corollary}[ANN-to-SNN Conversion]
\label{cor:practical_stability}
Let $\gamma \coloneqq \frac{\Vert W\Vert_\infty}{(\tau-1)\theta} < 1$. If $\tilde{x}(t) \to \mathbf{x}$, then $\mathcal{E}(t) \rightarrow \mathcal{E}^*$, i.e.,
\begin{equation}
    \mathcal{E}^* := \big\{y \in [0,1]^d~\big|~y = \text{ReLU1}\big(\tfrac{1}{(\tau-1)\theta}(F\mathbf{x} + Wy  + (u_{rest} - \theta)\mathbf{1})\big)\big\}
\end{equation}
is a singleton $\{y^*\}$, and $y(t)$ approaches its $\frac{1}{\tau(1-\gamma)}$-neighborhood at the geometric rate $\max\{\alpha, \gamma\}$, remaining there thereafter:
$\limsup_{t\to\infty}\Vert y(t) - y^*\Vert_\infty \le \frac{1}{\tau(1-\gamma)}$ (Proof at Appendix~\ref{pf:practical_stability}.)
\end{corollary}

Note that $\mathcal{E}(t)$ and $\mathcal{E}^*$ are both outputs of the same ReLU1 network evaluated at different inputs $\tilde{x}(t)$ and $\mathbf{x}$. Their only other difference is the initial-state term $\alpha^t u_{\text{pre}}(0)$ in $b(t)$, which vanishes within a few frames. The SNN output $y(t)$ tracks $\mathcal{E}(t)$ within an $\mathcal{O}(\frac{1}{\tau})$ band, and the error stays bounded on a moving stream as well (proof at Appendix~\ref{pf:tracking}):

\begin{corollary}[Tracking error on a stream]
\label{cor:tracking}
Let $\gamma < 1$ as in Corollary~\ref{cor:practical_stability}, so that $\mathcal{E}(t) = \{y^*_t\}$, let $\rho \coloneqq \max\{\alpha,\gamma\}$, and suppose that after some tick $t_0$ the target moves at most $\delta$ per tick, $\Vert y^*_t - y^*_{t-1}\Vert_\infty \le \delta$ for $t > t_0$. Then for every $k \ge 0$, whatever the state $y(t_0)$ was at $t_0$,
\begin{equation}
\underbrace{\Vert y(t_0+k) - y^*_{t_0+k}\Vert_\infty}_{\text{SNN tracking error}} \;\le\; R^* + \rho^{k},
\qquad\text{where}\quad
R^* \coloneqq \max\Big\{\underbrace{\tfrac{1/\tau + \gamma\delta}{1-\gamma}}_{\text{jitter}},\ \underbrace{(\tau-1)\,\delta}_{\text{lag}}\Big\}.
\end{equation}
So after any abrupt change, such as a scene cut at $t_0$, the error falls below $R^* + \epsilon$ within $\log(1/\epsilon)/\log(1/\rho)$ ticks, however long the stream has run.
\end{corollary}

\textbf{IF Neuron: steady-state equivalence does not imply tracking.} Under rate coding, IF-SNN dynamics simplify into a fixed-point iteration with a step size $\eta(t) = \frac{1}{t+1}$. The trajectory is attracted toward its corresponding ReLU1-DEQ model outputs $\mathcal{E}_{\text{IF}}(x(t))$. Unlike LIF-SNNs, the time-diminishing characteristic of the step size  $\eta(t)$~\citep{robbins1951stochastic} makes the IF-SNN trajectory converge \emph{exactly} towards  $\mathcal{E}_{\text{IF}}(\mathbf{x})$ once the input stabilizes $x(t) \to \mathbf{x}$, under 
$\Vert W\Vert_\infty < \theta$.
\begin{equation}
    \mathcal{E}_{\text{IF}}(x) = \big\{y \in [0,1]^d~|~y = \text{ReLU1}\big((Wy + Fx)/\theta\big)\big\} 
\end{equation}
This analytical mapping between the IF neuron and clipped ReLU at steady state~\citep{rueckauer2017conversion,han2020deeptsc} has been foundational for ANN-to-SNN conversion, and our findings reveal the transient dynamics underlying it. However, steady-state equivalence alone does not imply tracking: IF neurons satisfy it \emph{exactly}, yet cannot guarantee tracking of a changing input (Appendix~\ref{pf:if_memory}). Hence tracking (Theorem~\ref{thm:equilibrium_lif} and Corollary~\ref{cor:tracking}) is a strictly stronger property than steady-state conversion, and whether it holds depends on the neuronal dynamics.

\textbf{Empirical Validation.} 
We validate our theoretical findings through a toy experiment. We train a 7-layer MLP with ReLU1 activations to regress a smooth multivariate sinusoidal function, and convert it to SNN by replacing ReLU1s with LIF (or IF) neurons. 
Figure~\ref{fig:snn_trajectory_visualization} shows that LIF-SNN output continuously tracks its corresponding ReLU1-ANN's trajectory through time with a broad basin of attraction. Interestingly, the tracking dynamics governed by step sizes closely mirror online optimization dynamics: larger step sizes accelerate convergence at the expense of higher oscillation, whereas smaller step sizes yield smoother trajectories that lag behind. In contrast, time-diminishing step size (IF neuron) greatly lags behind the time-varying target (Detailed experimental setup is at Appendix~\ref{app:theory_validation_setup}.)

\section{Controlling SNNs to efficiently process data streams }
\label{sec:approach}

In the last section, we discovered a core principle governing the transient dynamics of SNN. We now leverage this principle to establish a framework that processes data streams accurately and efficiently with SNNs. Specifically, Theorem~\ref{thm:equilibrium_lif} shows that SNN constantly updates its output a step towards its corresponding ANN output. We \emph{control} the dynamics of SNN using this governing principle so that SNN continuously tracks its corresponding ANN as a reference model. 

Our \emph{\modelname{}} framework is a two-time-scale control that continuously adapts SNN to its time-varying reference model. It consists of (i) an inner loop, in which the SNN output closely tracks the output of the reference ANN model, and (ii) an outer loop, in which the reference ANN continuously evolves its output given input stream. Simply put, we continuously tune the reference ANN, which is easier to train and stably infers, to robustly control the unstable and discontinuous SNN computation.

\begin{figure}[t]
\centering
\vspace{-1em}
\begin{subfigure}[b]{0.65\textwidth}
    \centering
    \includegraphics[width=\linewidth]{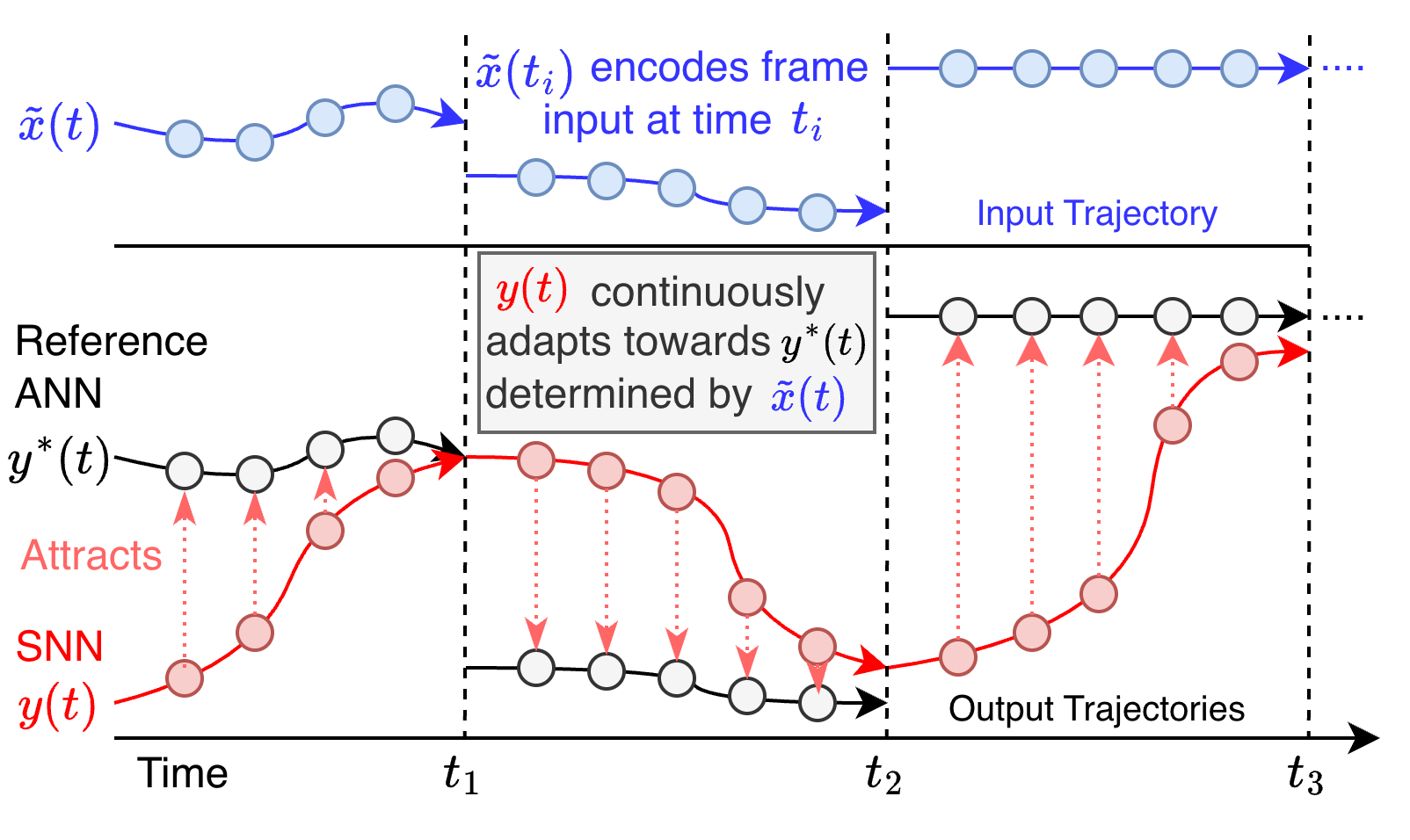}
\end{subfigure}%
\hfill
\begin{subfigure}[b]{0.35\textwidth}
    \centering
    \includegraphics[width=\linewidth]{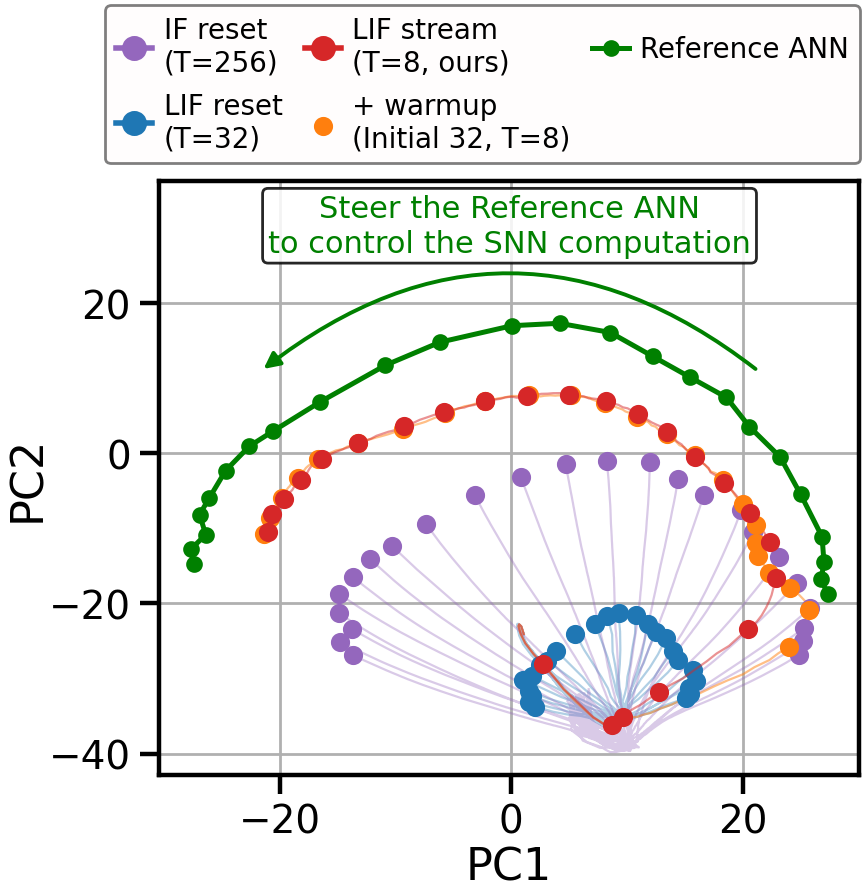}
\end{subfigure}
\vspace{-2em}
\caption{%
Conceptual overview and empirical trajectory analysis of \textbf{\modelname{}}. (Left) Conceptual diagram of our \modelname{} framework for streaming SNN inference. (Right) Empirical validation on MOT17 video sequences~\citep{milan2016mot16} evaluated as a streaming pedestrian detection task. \modelname{} efficiently tracks the continuous ANN output by updating only inter-frame delta, whereas standard reset-based IF and LIF-SNNs  suffer from re-convergence latency overhead for every frame.
}
\vspace{-2em}
\label{fig:conceptual_diagram}
\end{figure}

Figure~\ref{fig:conceptual_diagram} (left) illustrates the concept of our framework. SNN continuously updates its output to adapt to a new reference ANN output. 
Here, the inner loop can be framed as an online adaptation process of SNN $y$ to its reference model $y^*$ in a time window of $t \in [T_{i-1}, T_{i}]$, as follows:
\begin{itemize}
\item Estimator: The SNN output $y(t)$ encoded in spike trains.
\item Initial State: Prior SNN output $y(T_{i-1})$.
\item Goal: The ANN output $y^*(t)$ that stably evolves towards desired frame output $y^*(T_{i})$.
\item Update Rule: The forward propagation of SNN, a fixed point iteration (Lemma~\ref{thm:dynamics_lif}.) that attracts $y(t)$ towards the reference ANN output $y^*(t)$.
\item Step size: Constant $\frac{1}{\tau}$ (LIF) or time-decaying $\frac{1}{t+1}$(IF)
\end{itemize}

We pretrain a reference ReLU1-ANN with floating-point activations as a normal DNN. Unlike prior ANN-to-SNN conversion methods that utilize IF neurons, we convert ReLU1 functions into LIF neurons based on the steady-state equivalence in Corollary~\ref{cor:practical_stability}. Since the reference ANN achieves high accuracy with standard gradient-based optimizations, the SNN could also achieve high streaming accuracy as long as it stably and closely tracks the reference model. 

Algorithm~\ref{alg:stream_processing} simply implements our framework with LIF-SNNs. Such simplicity comes from two aspects. First, the principle we discovered intrinsically links SNN with its reference ANN through inputs, without additional component. Second, the step size of LIF dynamics is constant throughout the adaptation process. These preclude any intervention during stream processing. As in Algorithms~\ref{alg:float_encoding}--\ref{alg:spike_population_encoding}, input streams could be arbitrarily encoded, as long as the reference ANN stably and swiftly evolves towards the desired frame output. We empirically compare the encoding schemes in Appendix~\ref{app:encoder_comparison}.

\textbf{Streaming Efficiency} Our framework reveals how SNNs could be exceptionally efficient for processing temporally correlated data streams. We view SNN dynamics as an incremental computing algorithm of continuous-valued neural networks; It retains prior state and computes only the incremental updates in neural activations induced by input changes.
For continuous, temporally correlated streams, the activations change gradually rather than abruptly.
When the prior SNN state is already close to the target ANN output, and the reference ANN output changes minimally, the SNN requires only a small number of time steps to process a new frame.
Hence, SNN naturally exploits the temporal redundancy without sacrificing accuracy, whereas conventional DNN inference should fully recompute regardless of temporal correlation.

\textbf{Perpetual Inference} Incremental computation is most effective when it can operate continuously without external reset or  intervention. However, SNN inference systems frequently employ global reset that re-initialize all the neuronal states each frame to start with a clean slate. For our framework, this introduces three key drawbacks: (i) it completely discards the prior states that SNNs rely on for streaming efficiency, (ii) SNNs struggle to recover the accuracy after the reset, and  
(iii) it triggers global synchronization, which makes real-time systems less responsive to sudden events.

We observe that LIF neurons should be used instead of IF neurons to make neural computation perpetually incremental. Prior approaches using IF neuron (or its variants) are fundamentally limited because IF-SNNs have a \emph{time-diminishing step size}. The number of time steps (i.e., ticks) to track the reference model scales linearly over time, and thus it freezes after a few frames. Thus, in practice the IF conversions we compare against reset every frame. While recent IF-based conversion techniques show high accuracy in ultra-low ticks, they are only effective in the first few ticks after global reset. In contrast, LIF neuron's \emph{time-constant step size} makes the SNN permanently alert. LIF-SNN can continuously track the reference model's trajectory closely without any intervention. It is also naturally robust to long time horizon and scene cuts in input streams.

Furthermore, the IF neuron cannot guarantee continuous tracking under any decoder (including rate coding) because of \emph{charge debt}: negative input cannot leave an IF membrane as spikes and piles up as a debt that silences the neuron for a time proportional to the history. We prove in Appendix~\ref{pf:if_memory} that no decoder gives an IF neuron a lag bound that holds over a stream, whereas leak lets the debt of an LIF neuron fade and keeps its tracking error bounded (Corollary~\ref{cor:tracking}).

\textbf{Initial Warm-up.} SNNs 
can exhibit a cold start phenomenon: a temporary initial accuracy drop where the SNN output $y(t)$ requires several ticks to warm up from $y(0)$ to match the reference target $y^*(T_1)$.
This can be problematic in real-world scenarios when an agent should wake up fast and react.
To resolve this, we dynamically schedule more time-steps solely for the first frame, e.g., 32 ticks in Figure~\ref{fig:conceptual_diagram} (right). This design preserves both initial accuracy and a low average-case tick budget.

\textbf{Empirical validation on static-camera video streaming perception.} Figure~\ref{fig:conceptual_diagram} (right) demonstrates our framework on MOT17 video sequences~\citep{milan2016mot16} repurposed for streaming pedestrian detection. We use an FCOS-lite architecture~\citep{tian2019fcos} with a VGG backbone~\citep{simonyan2014very} restricted to affine layers and ReLU1, trained on COCO~\citep{lin2014microsoft}. We project the SNN output trajectories onto a 2D PCA subspace fitted to the ANN trajectory. \modelname{} smoothly tracks the reference ANN via local incremental updates within 8 ticks. In contrast, reset-based baselines reinitialize to the origin each frame, and thus incur significant time steps. Also, initial warm-up aligns \modelname{} with the ANN output in earlier frames.

\section{Evaluation}
\label{sec:eval}

To demonstrate generalizability, we extensively evaluate our framework across three distinct stream processing modalities, including vision, audio, and event streams. We benchmark against ANN-to-SNN conversion and direct SNN training baselines (see Appendix~\ref{app:baselines} for implementation details).

\textbf{Streaming Object Detection.} 
We assess streaming detection by zero-shot transferring models pretrained frame-by-frame on MS COCO~\citep{lin2014microsoft} directly to the Argoverse-HD benchmark~\citep{li2020towards}. COCO provides large-scale object detection pretraining across 80 classes, while Argoverse-HD provides high-frame-rate (30 FPS) egocentric video sequences annotated at 2D bounding-box level across eight road-object overlapping categories (24 validation streams, 15,062 frames). 
We report frame-wise mAP@50 evaluated sequentially over streaming frames. For conversion baselines, we adapt YOLOv8~\citep{jocher2023ultralytics} for SNN conversion (60.7\% AP50 on COCO, ReLU1). Full experimental setups are in Appendix~\ref{app:obj_det_eval}.

\textbf{Streaming Automatic Speech Recognition (ASR). } 
We train and evaluate streaming ASR on the LibriSpeech dataset ($\sim$960 hours of speech across 3–15 s clips)~\citep{panayotov2015librispeech}. To simulate a real-time streaming environment, audio signals are processed sequentially in small, fixed-size chunks rather than feeding full utterances simultaneously. Candidate transcriptions are served at a low latency every 33 ms to mirror real-time serving pipelines, with final output decisions committed every 1.28 s to ensure prediction stability. Accuracy is measured via Word Error Rate (WER). We base our conversion backbone on an adapted SummaryMixing architecture~\citep{parcollet2024summarymixing} (7.62\% WER on LibriSpeech);  Detailed model and training specifications are provided in Appendix~\ref{app:asr_eval}.

\textbf{Gesture Recognition on Event Streams.} We benchmark streaming event processing using the standard  DVS-Gesture dataset (1,342 event recordings across 11 gesture classes captured by a DAVIS240 sensor)~\citep{amir2017lowpower}. In contrast to offline clip-level classification, we evaluate online streaming inference: the model must continuously feed event streams and classify gestures using only the event history received so far. The task involves 12 target classes, including 11 gesture categories and an additional background (NULL) class. Performance is measured by frame-wise classification accuracy. As a conversion backbone, we adapt GestureNet~\citep{massa2020efficient} (a TCN-based model reaching 91.1\% streaming accuracy). Complete setup details are in Appendix~\ref{app:gesture_eval}. 
\begin{table*}[t]
\centering
\caption{Comparison of high-accuracy SNN techniques across ANN-to-SNN conversion and direct training approaches. \textbf{Calibration-Free}: requires no data-driven post-training calibration. \textbf{Binary Spike}: inter-neuron communication only with a single binary spike per tick; No burst or multi-precision spikes. \textbf{Standard neurons}: utilizes basic neuronal dynamics (e.g., IF/LIF) supported by conventional neuromorphic chips. \textbf{ANN Accuracy}: accuracy of the dense ANN each conversion starts from. \textbf{Bold} and \underline{underline} denote the best accuracy within the same time-step, across all methods and among conversion methods, respectively.
}
   \vspace{-0.5em}
\label{tab:snn_comparison_revised}
\resizebox{\textwidth}{!}{%
\begin{tabular}{l c c c c ccccccc}
\toprule
\multirow{ 2}{*}{\textbf{Method}}&
\multirow{ 2}{*}{\header{Calibration}{Free}}&
\multirow{ 2}{*}{\header{Binary}{Spike}}&
\multirow{ 2}{*}{\header{Standard}{Neurons}}&
\multirow{ 2}{*}{\header{ANN }{Accuracy}}& \multicolumn{7}{c}{\textbf{Task Accuracy ($T$ is time-steps per frame)}}
\\
\cmidrule(lr){6-12}
&  &  &  &  & \textbf{$T=2$} & \textbf{$T=4$} & \textbf{$T=8$} & \textbf{$T=16$} & \textbf{$T=32$} & \textbf{$T=64$} & Best \\
\midrule
\multicolumn{12}{c}{\textbf{Streaming Object Detection} (COCO $\to$ Argoverse-HD Zero-shot Transfer, Accuracy metric: AP50, higher the better)} \\
\midrule
EMS-YOLO~\citep{su2023deep}    & \cmark & \cmark & \cmark & -- & \textbf{20.2} & 22.1 & \textbf{22.5} & 22.7 & 23.2 & 23.6 & 23.7 \\
SpikeYolo~\citep{luo2024integer}     & \cmark & \xmark & \xmark & -- & - & \textbf{30.3} & - & - & - & - & 30.3 \\
\midrule
ReLUNorm~\citep{rueckauer2017conversion}     & \xmark & \cmark & \cmark & 38.8 & 0.0 & 0.0 & 0.0 & 0.0 & 0.0 &0.9 &  33.6\\
Spike-Calib~\citep{li2021freelunch} &\xmark & \cmark & \cmark & 38.8 & 0.0 & 0.0 & 0.0 & 0.1 & 2.5 & 10.8 & 34.6\\
QCFS~\citep{buqcfs}          & \xmark & \cmark & \cmark & 34.5 & 0.0 & 0.0 & 1.5&4.2    &9.3    &25.5& 34.1 \\
SlipReLU~\citep{jiang2023sliprelu}      & \xmark & \cmark & \cmark & 35.0 & 0.0 & 0.0 & 0.2 & 0.8   & 2.9   & 12.0 & 26.0 \\
RMP-SNN~\citep{hao2023reducingresmembranepotential}                                 & \xmark & \cmark & \cmark & 38.8 & 0.0 & 0.0 & 0.0 & 0.0 & 0.0 & 0.0 & 29.8 \\
DCGS~\citep{huang2025differential}                                    & \xmark & \cmark & \xmark & 38.8 & 0.0 & 0.0 & 0.0 & 0.7 & 3.8 & 12.6 & 37.3 \\
AdaFire~\citep{wang2025adaptive}                           & \xmark & \xmark & \xmark & 38.8 & 0.0 & 0.0 & 0.0  &0.4    &6.4    &20.7& 35.4 \\
AdaFire+SSC~\citep{wang2025adaptive}                             & \xmark & \xmark & \xmark & 38.8 & 0.0 & 0.0 & 0.1    &1.4    &7.4    &17.8 & 25.0 \\
LIF Neuron + Per-frame Reset                             & \cmark & \cmark & \cmark  & 37.9 & 0.0 & 0.0 & 0.0   & 0.0   & 0.0   & 25.7 & \underline{\textbf{37.4}} \\
\midrule
\textbf{\modelname{} (Ours)}            & \cmark & \cmark & \cmark & 37.9 & \underline{12.3} & \underline{16.8} & \underline{21.5} & \underline{\textbf{25.2}}  & \underline{\textbf{26.5}} & \underline{\textbf{27.8}} & \underline{\textbf{37.4}} \\
\midrule
\midrule
\multicolumn{12}{c}{\textbf{Streaming Automatic Speech Recognition} (LibriSpeech dev-clean [greedy, no LM], Accuracy metric: WER, lower the better)} \\
\midrule
ReLUNorm~\citep{rueckauer2017conversion}     & \xmark & \cmark & \cmark & 7.6 & 100.0 & 100.0 & 100.0 & 100.0 & 100.0 & 100.0 & 72.6  \\
RTS~\citep{deng2021optimal} & \xmark & \cmark & \cmark& 7.6 & 100.0 & 100.0 & 100.0 & 100.0 & 100.0 & 100.0 & 62.9\\
Spike-Calib~\citep{li2021freelunch} &\xmark & \cmark & \cmark & 7.6& 100.0 & 100.0 & 100.0 & 100.0 & 98.1  & 91.4  & 65.7\\
QCFS~\citep{buqcfs}          & \xmark & \cmark & \cmark & 17.9 & 114.9 & 95.9   & 61.7 &    30.2 &  21.1& 18.8 & 18.2 \\
SlipReLU~\citep{jiang2023sliprelu}      & \xmark & \cmark & \cmark & 14.5 & 122.9& 81.5 &   42.7& 23.9 & 18.1 & 16.2  & 14.9 \\
RMP-SNN~\citep{hao2023reducingresmembranepotential}                                 & \xmark & \cmark & \cmark & 7.6 & 100.0 & 100.0 & 100.0 & 100.0 &100.0 & 94.2 & 18.3 \\
DCGS~\citep{huang2025differential}                                    & \xmark & \cmark & \xmark & 7.6 & 143.9 & 108.7  & 95.9 &    66.2 &  27.9 & 15.1  & 11.9 \\
AdaFire~\citep{wang2025adaptive}                           & \xmark & \xmark & \xmark & 7.6 & 97.1  & 94.0 &51.0 &  19.1& 12.2 &    10.5 &  10.1  \\
AdaFire+SSC~\citep{wang2025adaptive}                             & \xmark & \xmark & \xmark & 7.6 & 99.8 & 99.2 & 58.4  & 19.4 &    \underline{\textbf{11.9}} & 10.5 & 9.8 \\
LIF Neuron + Per-frame Reset                             & \cmark & \cmark & \cmark  & 7.8 & 100.0 &    100.0 & 99.9 &  87.6 &  12.1 & 9.9 & 8.1 \\
\midrule
\textbf{\modelname{} (Ours)}            & \cmark & \cmark & \cmark & 7.8 & \underline{\textbf{44.6}} &  \underline{\textbf{27.8}} & \underline{\textbf{19.3}} & \underline{\textbf{15.5}} & 12.0 &\underline{\textbf{9.8}} &  \underline{\textbf{8.0}} \\
\midrule
\midrule
\multicolumn{12}{c}{\textbf{Event Stream-based Gesture Recognition} (DVS-Gesture, Accuracy metric: Multi-class Accuracy over 11+1 classes, higher the better)} \\
\midrule
STBP~\citep{wu2018spatio}       & \cmark & \cmark & \cmark& -- & 28.2&  87.5    &\textbf{90.2}  &\textbf{90.6}  &\textbf{90.5}  &90.3 & 90.6 \\
TET~\citep{deng2022temporal}  & \cmark & \cmark & \cmark & -- & 65.1 & \textbf{88.4} &  89.7 &  89.7    & 89.5 &    89.5 & 89.7 \\
OTTT~\citep{Xiao2022OnlineTT}            & \cmark & \cmark & \cmark & -- & 19.8 & 86.3 & 89.7 & 90.3 & 90.3 & 90.3 & 90.3 \\
SLTT~\citep{meng2023sltt}          & \cmark & \cmark & \cmark & -- & 37.6 & 86.7 & 88.5 & 88.8 & 88.8 & 88.9 & 88.9 \\
ASGL~\citep{wang2023asgl}       & \cmark & \cmark & \xmark & -- &  36.3 &   86.9 & 89.6 & 89.8 & 89.7 & 89.7 & 89.8 \\
SDT~\citep{yao2023sdt} & \cmark & \cmark & \cmark & -- & - & - & - & 89.4 & - & - & 89.4 \\
\midrule
ReLUNorm~\citep{rueckauer2017conversion}     & \xmark & \cmark & \cmark & 91.1 & 25.2&  25.0    &25.0   &4.8    &7.5    &28.4&  86.9 \\
RTS~\citep{deng2021optimal} & \xmark & \cmark & \cmark & 91.1 & 12.6 & 4.8 & 4.8 & 4.8 & 5.9& 25.2 & 86.9  \\
Spike-Calib (adv)~\citep{li2021freelunch} &\xmark & \cmark & \cmark & 91.1 & 25.0& 27.5 & 49.2 &    78.8 &  86.2 &  87.5 &  88.6  \\
QCFS~\citep{buqcfs}          & \xmark & \cmark & \cmark & 91.1 & 59.1   &69.5   &85.9   &88.3&  89.6&   90.4 & 90.8 \\
SlipReLU~\citep{jiang2023sliprelu}      & \xmark & \cmark & \cmark & 91.5 & 56.8    &63.7   &79.5   &87.4   &89.3   &90.4 & 90.8 \\
RMP-SNN~\citep{hao2023reducingresmembranepotential}                                 & \xmark & \cmark & \cmark & 91.1 & 25.0 &  14.3     &4.8 & 4.8 &   4.8 &   18.5 &  26.5 \\
DCGS~\citep{huang2025differential}                                    & \xmark & \cmark & \xmark & 91.1 &  25.0 &25.0&25.0&25.0 &27.0   &53.3   &   86.0 \\
AdaFire~\citep{wang2025adaptive}                           & \xmark & \xmark & \xmark & 91.1 & 12.6 &25.0   &24.8   &4.8    &9.0    &31.1   &   87.0 \\
AdaFire+SSC~\citep{wang2025adaptive}                             & \xmark & \xmark & \xmark & 91.1 & 12.6 & 25.0 & 25.0 & 4.8 & 4.8 & 16.2 &  85.8 \\
LIF Neuron + Per-frame Reset                             & \cmark & \cmark & \cmark  & 91.1 & 13.8 &     19.8 & 19.8 & 20.8 &   88.2 &  \underline{\textbf{90.5}} & \underline{\textbf{90.9}} \\
\midrule
\textbf{\modelname{} (Ours)}            & \cmark & \cmark & \cmark & 91.1 &\underline{\textbf{66.4}} & \underline{80.3} &   \underline{88.5} &  \underline{90.0} &  \underline{90.4} & \underline{\textbf{90.5}} & \underline{\textbf{90.9}} \\
\bottomrule
\end{tabular}%
}
\vspace{-2em}
\end{table*}

\textbf{Results.} 
Table~\ref{tab:snn_comparison_revised} shows that \modelname{} generally achieves the best efficiency-accuracy trade-off among SNN techniques: (i) it is substantially more tick-efficient than conversion methods while maintaining  peak accuracy, and (ii) it achieves higher peak accuracy with competitive low-tick performance compared to direct training methods. Prior conversion approaches degrade sharply in streaming tasks at low tick budgets ($T < 8$). Conversely, while direct training performs well at ultra-low ticks, its peak accuracy remains constrained, and it fails to train on complex backbones matching conversion baselines beyond DVS-Gesture (e.g., streaming object detection and ASR).

Importantly, \modelname{} is natively compatible with standard digital neuromorphic hardware supporting soft reset, as it operates out-of-the-box without post-training model calibration, customized neuronal dynamics, or non-binary spike communication. These structural properties also preserve the full ANN-level peak accuracy ceiling of our method. Prior conversion methods frequently sacrifice peak accuracy to compensate for low-tick latency (via parameter altering~\citep{li2021freelunch} or discontinuous activation functions~\citep{buqcfs,jiang2023sliprelu}). Similarly, direct SNN training suffers from an inherent accuracy gap with ANNs due to discontinuous spiking mechanisms and surrogate gradient approximations, which trade exactness for smoothness. %

This tick-efficiency stems from our incremental computing approach. 
Streaming inference tasks frequently exhibit high inter-frame similarity, which \modelname{} exploits to reduce the average computational cost required for the SNN to track the reference ANN prediction. Consequently, our framework achieves high accuracy with low frame-wise ticks. This reduction is vital for real-time streaming, where per-frame computing load directly bottlenecks the system throughput, making low tick latency essential for SNNs in streaming workloads. 

In terms of theoretical energy consumption (Appendix~\ref{app:indepth:ops}), at $T=16$ \modelname{} reaches higher AP50 on Argoverse-HD than AdaFire at $T=64$, QCFS at $T=128$ and SlipReLU at $T=128$, with $6.1\times$, $6.4\times$ and $6.6\times$ less energy per frame, respectively. At $T=32$, it also uses $2.1\times$ less energy than the same LIF network with a per-frame reset at $T=64$. We also tested its robustness by streaming three scenes back to back without a reset, randomly dropping 30\% of frames, and adding Gaussian noise to every frame. Scene cuts cost nothing, and frame drops and noise reduce AP50 by at most $2.7$ and $1.7$ points ($15\%$ and $8\%$ relative), respectively (Appendix~\ref{app:indepth:robust}).

\begin{figure*}[t]
\centering

\begin{minipage}[c]{0.48\textwidth}
\centering
\includegraphics[width=\linewidth]{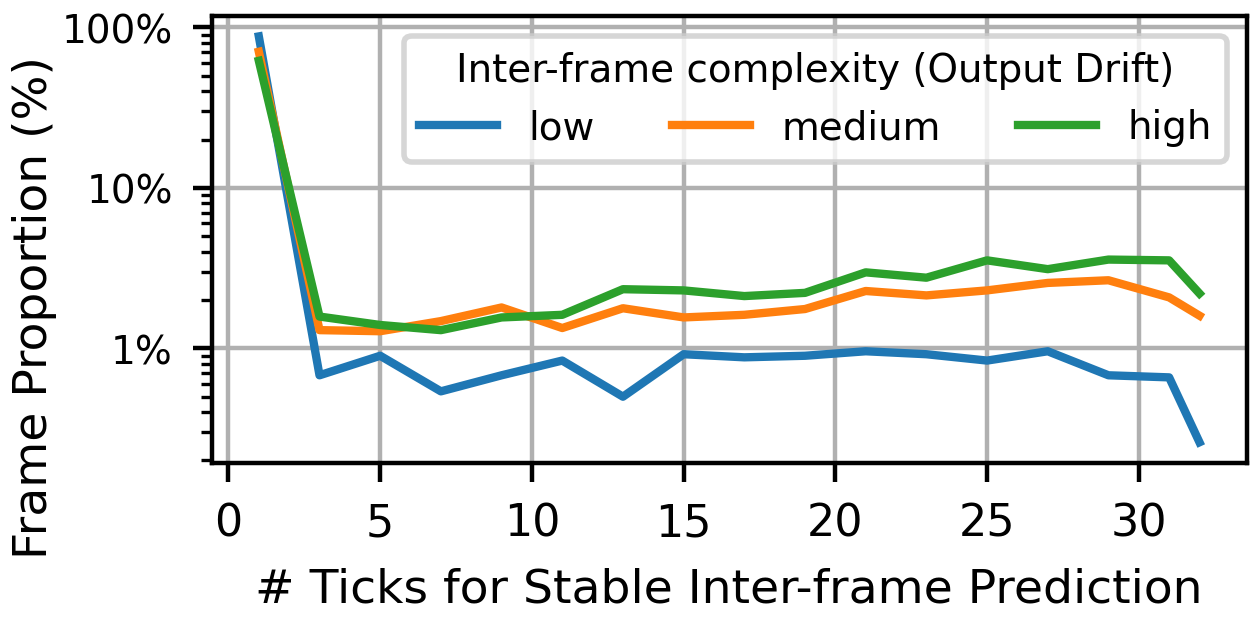}
\vspace{-2em}
\caption{Efficiency via inter-frame redundancy (post-hoc analysis; Table~\ref{tab:snn_comparison_revised} uses a fixed number of ticks).}
\vspace{0.5em}
\label{fig:efficiency}
\end{minipage}%
\hfill
\begin{minipage}[c]{0.48\textwidth}
\centering
    \includegraphics[width=\linewidth]{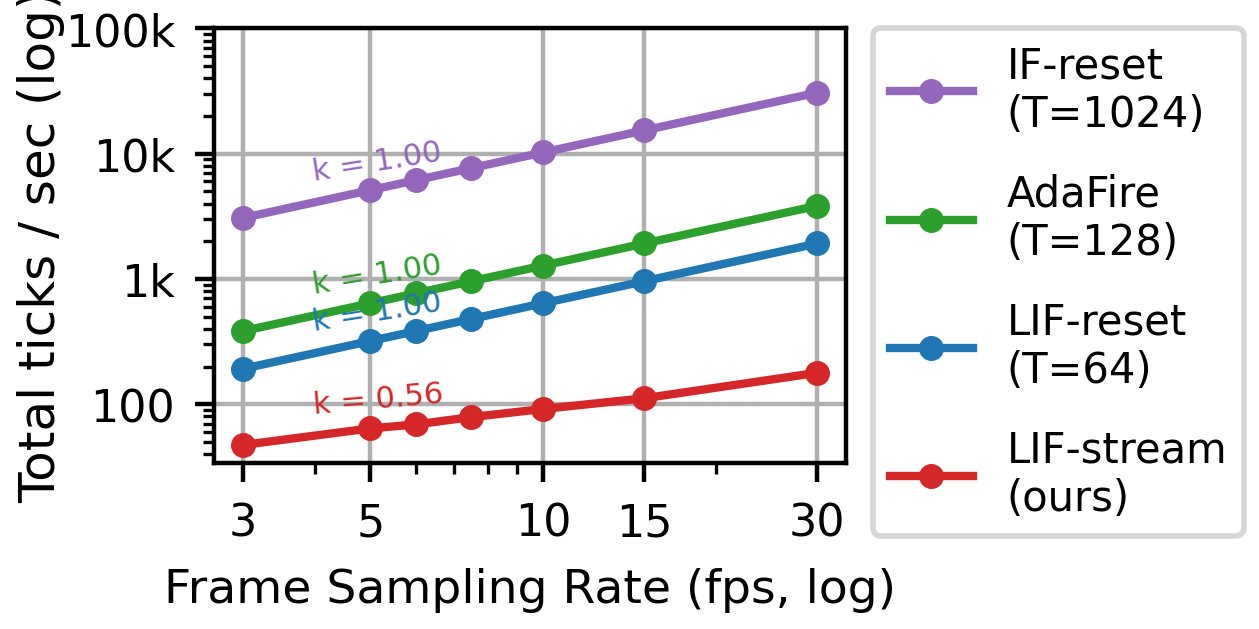}
\vspace{-2em}
\caption{Finer temporal resolution benefits ours (post-hoc analysis).}
\vspace{0.5em}
\label{fig:temporal_resolution}
\end{minipage}

\begin{minipage}[c]{0.50\textwidth}
\centering
\includegraphics[width=\linewidth]{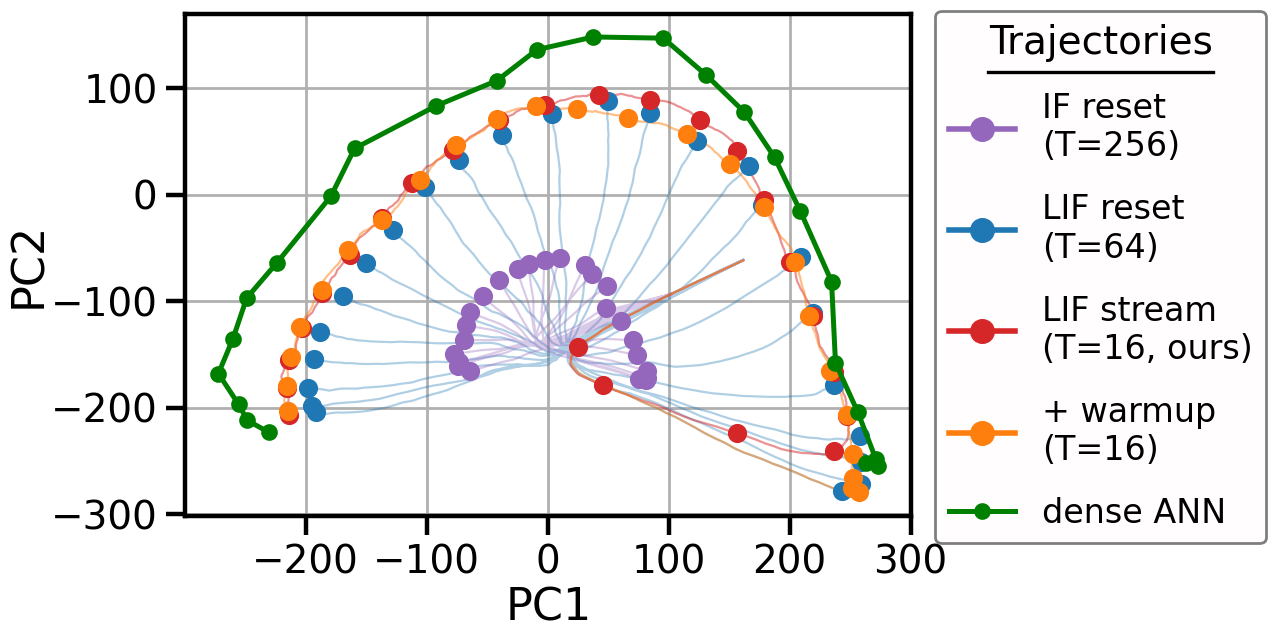}
\vspace{-2em}
\caption{SNNs on egocentric driving perception.}
\label{fig:qualitative}
\end{minipage}%
\hfill
\begin{minipage}[c]{0.48\textwidth}
\centering
\includegraphics[width=\linewidth]{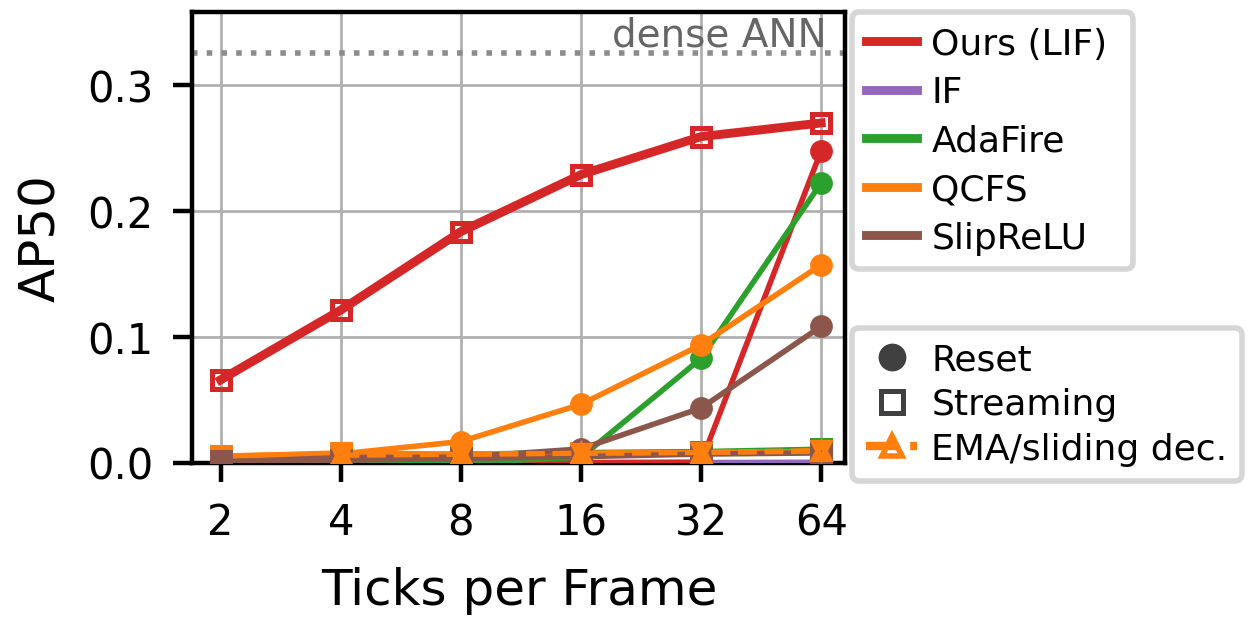}
\vspace{-2em}
\caption{Prior IF-conversions fail to stream.}
\label{fig:ablation_study}
\end{minipage}
\vspace{-1.5em}
\end{figure*}
\subsection{In-depth analysis}
\label{sec:in_depth}
We comprehensively analyze \modelname{} across multiple dimensions. Setup details are in Appendix~\ref{app:indepth}.

\textbf{Inter-frame Redundancy benefits \modelname{}. } 
We demonstrate the efficiency gains of our incremental computing SNN by measuring the distribution of inter-frame ticks required to reach a stable, confident prediction. Specifically, we conduct a post-hoc evaluation on \modelname{} using Argoverse dataset and measure, for each frame, the number of ticks needed to fall and remain within an $\epsilon = 0.15$ error tolerance of the peak SNN F1 score. We define frame complexity as the prediction drift of the reference ANN, given by the relative variation $\Vert y^* - y^*_{\text{prev}} \Vert / \Vert y^*_{\text{prev}} \Vert$. As shown in Figure~\ref{fig:efficiency}, a logarithmically dominant portion of frames converge in fewer than three ticks due to temporal redundancy. This effect becomes even more pronounced for lower-complexity frames, confirming that SNNs efficiently process consecutive frames by reusing prior prediction states.

\textbf{Finer temporal resolution benefits \modelname{}.} 
To further demonstrate computational efficiency, we examine how the tick cost required for accurate predictions scales with temporal resolution. Using the Argoverse dataset, we perform streaming SNN inference over fixed time intervals and aggregate the ticks needed per frame to reach the stable prediction region (within $\epsilon=0.15$ of the peak SNN F1 score). We then compute the cumulative tick consumption per second across varying temporal resolutions. As shown in Figure~\ref{fig:temporal_resolution}, reset-based SNN baseline scales linearly ($\mathcal{O}(T^{1.0})$) with temporal resolution because it discards prior state information and recomputes predictions from scratch for every frame. In contrast, \modelname{} exhibits sub-linear scaling ($\mathcal{O}(T^{0.56})$). Higher temporal resolution increases inter-frame similarity, reducing \modelname{}'s computation per frame.

\textbf{Qualitative Evaluation.} Figure~\ref{fig:qualitative} illustrates the streaming inference trajectory to highlight how \modelname{} efficiently handles consecutive frames unlike reset-based frame-wise approaches. Inference is evaluated on a stream from Argoverse-HD, featuring complex egocentric driving scenes in contrast to the static person-detection scenarios in Figure~\ref{fig:conceptual_diagram}. 
Reset-based methods reinitialize their hidden states to the origin at every frame, requiring them to cover a significantly longer distance in state space per frame. 
In contrast, \modelname{} computes small state updates between the prior state and the new  target prediction, effectively exploiting inter-frame prediction deltas. As a result, it converges faster with fewer ticks. Additionally, our initial warm-up mechanism positions the SNN state near the ANN output at the first frame, effectively preventing early mispredictions.

\textbf{Compatibility with IF-SNN conversions. } 
Our approach is conceptually orthogonal to advanced conversion techniques that calibrate or optimize IF-SNNs, as it is built upon a distinct theoretical foundation. 
We evaluate our streaming inference framework alongside existing ANN-to-SNN conversion methods to examine how these techniques interact. Specifically, we apply our framework across diverse conversion methods (while maintaining IF neurons) and measure their AP50 accuracy over ticks on Argoverse-HD to determine if incremental computing benefits advanced conversion techniques. As shown in Figure~\ref{fig:ablation_study}, our streaming SNN inference does not transfer to existing IF-neuron-based conversion methods. Although these techniques are engineered to accelerate early-step inference in static frames, their gains do not propagate across frames, as our streaming framework relies theoretically on LIF neuron dynamics. Consequently, this demonstrates that streaming SNN inference calls for tailored optimizations beyond standard static IF conversion approaches.

We also empirically analyze whether the streaming inference capability comes from stream-friendly decoders by applying them to the streamed IF network. With an EMA or a sliding-window decoder, every streamed IF conversion we tested (QCFS, SlipReLU, ReLU-Norm and AdaFire) stays below $1$ AP50 (Figure~\ref{fig:ablation_study}, Appendix~\ref{app:indepth:porting}), as the charge-debt analysis of Appendix~\ref{pf:if_memory} predicts.

\section{Related Works}

\textbf{Direct SNN Training.} A common approach for building high-accuracy SNNs is to directly train them with backpropagation-based methods~\citep{lee2016training}. The learning signal backpropagates across time steps, and surrogate gradients~\citep{Neftci2019SurrogateGL} relax the discontinuous firing rule with smooth approximations. Techniques such as OTTT~\citep{Xiao2022OnlineTT}, e-prop~\citep{bellec2020solution}, mixed-mode differentiation~\citep{zenke2021braininspiredlearning}, or event-driven BP~\citep{zhu2022training} further improve computational efficiency. BPTT extends to advanced sequence processing architectures such as spiking transformers~\citep{zhou2022spikformer,zhou2023spikingformer} or state-space SNN models~\citep{stan2024learning, shen2025spikingssms}. These methods do not compute states incrementally. In contrast, \modelname{} reuses prior states to compute only incremental activation updates without specialized training, and guarantees this property perpetually over unbounded streams. Furthermore, directly trained SNNs generally exhibit an accuracy gap relative to their ANN counterparts~\citep{fang2021sewresnet, sengupta2019going, zhou2022spikformer} due to the gradient mismatch introduced by surrogate gradient approximations.

\textbf{ANN-to-SNN Conversion} is an alternative of direct training to build a high-performance SNN that can match the ANN accuracy.  \citet{cao2015spiking} and \citet{diehl2015thresholdbalancing} first empirically demonstrated its feasibility. \citet{rueckauer2016theory,rueckauer2017conversion} established its theoretical basis by finding a steady-state equivalence between discrete IF neuron firing rates and clipped ReLU activations. Because early conversion methods took too many time steps even for a single static input, subsequent works focused on improving the static SNN accuracy in early steps~\citep{diehl2015thresholdbalancing, li2021freelunch, hao2023bridging}. \citet{deng2021optimal} formulated the layer-wise conversion errors, and  \citet{buqcfs} categorizes this error to propose a clip-floor function to replace ReLU1. \citet{huang2025differential} propose a new differential multi-threshold neuron that communicates rate delta, and \citet{hwang2024spikedattention} converted Transformers using a custom winner-takes-all neurons and one-spike phase coding. 

Few works keep neuron states across multiple frames. \citet{kugele2020efficient} convert an ANN to IF neurons and keep membrane potentials across frames as a warm start, but their ANN is a streaming-rollout network trained over multiple frames. Unlike us, they note that the carried state can slow convergence, and they neither analyze state reuse nor claim it as a contribution. Moreover, they bound the membrane from below to avoid long silence after negative input. As we prove in Remark~\ref{rem:if_bound}, such a bound breaks the conversion relation, since the firing rate no longer equals ReLU1 of the mean input. \citet{diehl2016conversion} keep IF states within a sentence and reset between sentences, and \citet{perez2013mapping} run leaky event-driven neurons continuously with a hand-tuned leak. None of these works analyzes why state reuse works or when it fails; our theory explains both, and shows that a plain LIF neuron on an ordinary frame-wise ANN suffices (Appendix~\ref{pf:if_memory}). 

We contribute to the conversion theory in two novel directions: (i) we reveal the principle underlying the transient SNN dynamics which fundamentally enabled the ANN-to-SNN conversion, and (ii) we make the converted SNNs efficiently process the sequence or streaming inputs without any additional training or fine-tuning. We also challenge the conventional dependence on IF neurons, showing why LIF neurons should be primarily used for stream processing.

\textbf{SNN Theories.} 
Continuous-time SNNs have been widely studied theoretically in terms of computational power~\citep{Maass1996Lowerbound,Maass1996ThirdGen} or dynamics~\citep{gerstner2014neuronal,zhang2021bifurcation, mancoo2020understandingconvexopt}. However, directly applying these results to discrete-time SNNs causes errors that accumulate over time~\citep{mancoo2020understandingconvexopt, niesen2004globaldiscretizationerror}. 
Discontinuous discrete dynamics are more difficult to analyze, and hence relatively few works have studied discrete-time SNNs;  through symbolic dynamics~\citep{cessac2008discretespikeneurontheory}, ergodic theory~\citep{cessac2011discrete}, geometric expressivity~\citep{nguyen2025time}, or optimization theory~\citep{oh2024sign}. We  provide a new perspective by viewing the transient dynamics of discrete SNNs as continuous attraction toward an instantaneous ANN. This bridges transient SNN dynamics, reference tracking, and the conversion theory. Our view is closest to spike-coding networks in computational neuroscience~\citep{boerlin2013predictive,deneve2016efficient}, in which exponentially filtered spike trains track a target signal and each neuron spikes only when this reduces the readout error, so that computation is incremental by construction. The targets have grown from linear dynamical systems to nonlinear dynamics~\citep{thalmeier2016learning,alemi2018learning} and convex piecewise-linear input-output functions akin to ReLU layers~\citep{mancoo2020understandingconvexopt}. In all of these, the connectivity and thresholds are derived from, or learned toward, the target, which requires specific structure such as fast recurrent inhibition or dendritic nonlinearities. In contrast, we start from an arbitrary trained deep ReLU1 network, convert it with standard conversion into standard LIF neurons, and prove that the resulting dynamics already track the instantaneous fixed point of the corresponding firing-rate network, a static map of the current filtered input, under a contraction condition that every feedforward network satisfies. Incremental computation thus comes for free from conversion, which lets us run deep networks incrementally on data streams. 

Sigma-delta modulation~\citep{de1952delta, Steele1975DeltaMS, o2016sigmadelta} conceptually links to our work. In detail, \citet{yoon2016lif} interpreted LIF neuron as a variant of sigma-delta modulator, an analog-to-digital converter that communicates only quantized temporal differences between nodes. Following works applied this idea for sequence processing DNNs by computing only quantized activation deltas~\citep{ o2016sigmadelta,o2017temporally,zambrano2019sparse}.
In contrast, we view SNN computation through the lens of reference tracking.  The concept of closing the temporal activation gap is a byproduct of controlling SNN dynamics for efficient stream processing. 

\textbf{Incremental Computation of Neural Activations.} Delta networks~\citep{neil2017delta,gao2018deltarnn,Parger2022DeltaCNNEC, yousefzadeh2022delta} reuse prior activations and incrementally compute only activation differences. However, the efficiency gain only comes when the activation deltas are sparsified through quantization, and thus the ANN should be trained to be aware of this quantization effect to reduce its impact on stream processing accuracy. Techniques for vision tasks exploit the temporal redundancy in their features by computing feature changes from events~\citep{cavigelli2017cbinfer,dutson2022event,buckler2018eva2,li2022efficient}. More coarse-grained approaches adapt neural computation through early exit~\citep{teerapittayanon2016branchynet}, test-time block scaling~\citep{fang2018nestdnn,han2021legodnn}, or caching~\citep{huynh2017deepmon, narayanan2018deepcache}. Unlike them, our work performs fine-grained, neuron-level incremental computation with SNNs, and computes to narrow the ANN-SNN activation gap instead of the inter-frame ANN activation gap.

\section{Conclusion}
We reveal a core principle underlying discrete SNN dynamics: an instantaneous ReLU1-ANN serves as a moving reference target that drives the output trajectory of the discrete SNN. Upon this, we build \modelname{}, an efficient stream processing framework that controls a discrete SNN to continuously adapt to the reference ANN's output trajectory, thereby incrementally computing the streaming perception. 
Comprehensive evaluations across three large-scale benchmarks with distinct modalities show that our method achieves the best efficiency-accuracy trade-off. \modelname{} is deployable out-of-the-box only with a trained ReLU1-ANN, requiring no post-training model calibration, customized neuron dynamics, or non-binary spike formats. Broader discussions, including limitations regarding hardware evaluations and sequence modeling, are detailed in Appendix~\ref{app:discussions}.

\subsubsection*{Acknowledgments}
This work was supported by the Institute for Information \& Communications Technology Planning \& Evaluation (IITP) grant funded by the Korea government (MSIT) (No.~RS-2026-25507282, Development of AI learning technologies based on brain cognitive science and biological learning principles). This work was supported by the National Research Foundation of Korea(NRF) grant funded by the Korea government(MSIT) (No.~RS-2024-00463802).

\bibliography{references,arxiv_extra}
\bibliographystyle{iclr2027_conference}

\appendix

\section*{Technical Appendices}
\section{Discussions and Limitations}
\label{app:discussions}
\paragraph{On-device Always-on Inference.} A prominent application of this technique is on-device continuous inference. These workloads are typically embodied and event-driven, meaning that sensory inputs are temporally correlated rather than batched. Crucially, surprising events are usually rare, and a proper, composed reaction is often more adequate than immediate precision.  Computational resources are scarce on mobile devices; the static amortized cost must remain low while dynamically scaling up when sudden events occur abruptly.

Our technique precisely satisfies these conditions: our average-case cost reduction, combined with the energy efficiency of neuromorphic computing, leads to ultra-low cost in static scenes with high temporal redundancy, while adaptively allocating higher computation only during surprising events.

In contrast, standard ANNs consume significant energy on mobile devices, making true always-on execution infeasible unless they are (i) constantly plugged into power, or (ii) triggered on-demand by humans or lightweight wake-up models. However, the trigger-based approach (ii) either loses immediate always-on responsiveness or inherently sacrifices the accuracy of the main model during non-triggered intervals. By contrast, our technique enables SNNs to maintain permanent responsiveness with ultra-low latency, without sacrificing model accuracy. It is thus a candidate enabling technique for ubiquitous, low-cost, always-on sensing applications; demonstrating it on a neuromorphic chip is future work.

\paragraph{Biological Implications.} Since SNNs serve as simplified models of biological neural networks, our theory offers valuable insights into how biological brains efficiently sense, perceive, and react continuously throughout their lifespan. Biological brains do not reprocess full-resolution sensory inputs for every inference, nor do they exhibit global state resets. Like SNNs, the brain may persistently retain neural activations and compute only to minimize the error between current spike-coded activations and underlying continuous-valued signals. This mechanism conserves energy in static environments while maintaining constant readiness to respond to unexpected events.

In this perspective, the intrinsic leakage in LIF neurons enables unbounded stream processing without state resets, analogous to how biological forgetting via neuronal leakage clears capacity for new information, thereby facilitating lifelong inference in biological agents. Nevertheless, biological neuronal dynamics are continuous and substantially more complex, and hence extending our theoretical framework to continuous-time and biophysically detailed models, such as the Hodgkin-Huxley~\citep{hodgkin1952quantitative}, adaptive LIF~\citep{gerstner2014neuronal}, or Izhikevich~\citep{izhikevich2003simple} models, remains an essential direction for future work to validate this hypothesis.

\paragraph{Predictive Coding.} Our theoretical framework aligns closely with the principles of predictive coding~\citep{friston2009predictive}. Predictive coding views the brain as an active \emph{prediction engine}: rather than constructing perception purely bottom-up by processing raw sensory inputs, the brain uses internal generative models to continuously predict incoming signals. Within our formulation, the top-down predictions correspond to the prior SNN output $y(t)$, while the bottom-up prediction error represents the activation gap $y^*(t) - y(t)$ relative to the target ANN output $y^*(t)$. The internal state is subsequently updated through adaptation dynamics that drive $y(t)$ toward $y^*(t)$.In this sense, the neural network computes only the prediction error rather than re-evaluating the full input scene, directly analogous to how our SNN operates exclusively to close the ANN-SNN activation gap. This error-driven scheme ensures extreme bandwidth and energy efficiency by transmitting prediction errors instead of full-resolution sensory data, while enabling rapid response through temporal anticipation. Consequently, our framework provides a formal mechanism demonstrating how predictive coding operates within and benefits SNNs from an ANN-SNN equivalence perspective. An important extension of predictive coding pertains to learning, which we discuss next.

\paragraph{Online Learning for SNNs.} Extending our theoretical framework to train the corresponding ANN online over continuous data streams, and thereby train the SNN dynamically, represents a promising research direction. The traditional paradigm of ANN-to-SNN conversion relies on pre-training an ANN offline prior to conversion. However, extending this framework to online learning disentangles ANN-SNN correspondence from a purely static conversion pipeline, establishing this equivalence as a general paradigm for studying SNNs and potentially biological neural networks.

Our theory naturally connects to local, bio-plausible learning methods that assign credit based on per-neuron activation error signals, including predictive coding~\citep{rao1999predictive, friston2009predictive}, equilibrium propagation~\citep{scellier2017equilibrium}, and contrastive Hebbian learning~\citep{movellan1991contrastive}. While STDP~\citep{kheradpisheh2018stdp} also measures activation differences over time rather than direct value encodings, its precise theoretical link warrants further investigation.

Prior conversion techniques offer limited implications for online SNN training since they lack an online adaptation perspective; at equilibrium, the corresponding ANN retains few signal components to exploit beyond static state values. Consequently, training traditionally resorts to standard backpropagation or, for feedback architectures, Deep Equilibrium methods such as implicit differentiation~\citep{bai2019deq, xiao2021trainingfsnn} and phantom gradients~\citep{geng2021phantom}. While DEQ training concepts offer valuable insights, these approaches remain computationally expensive and require global coordination, highlighting the advantages of our adaptive formulation.

\paragraph{Hardware Implementation.} Our stream processing SNN framework uses only operations supported by neuromorphic hardware platforms with LIF neurons and soft reset. As a cornerstone model in computational neuroscience, the LIF neuron is widely supported across modern neuromorphic platforms, including Intel Loihi~\citep{davies2018loihi, orchard2021loihi2}, IBM TrueNorth~\citep{akopyan2015truenorth}, SpiNNaker~\citep{furber2014spinnaker, gonzalez2024spinnaker2}, and custom FPGA-based implementations~\citep{han2020hardware}. Crucially, our method avoids the need for global state resets, a requirement inherently imposed by prior conversion techniques, allowing each neuron to operate independently without global synchronization beyond a shared clock.

This offers a significant hardware advantage over recent trends aiming for high-accuracy SNNs. Such approaches typically introduce non-standard components alongside basic neuronal dynamics, including synchronized bitwise operations across spike trains~\citep{zhou2022spikformer}, customized neuron dynamics~\citep{oh2024sign}, group dynamics~\citep{hwang2024spikedattention}, or multi-bit spiking mechanisms~\citep{guo2024ternary, xing2024spikelm}. These additions incur substantial silicon complexity overheads without yielding proportional energy efficiency gains, making custom chip fabrication impractical. We evaluate energy efficiency theoretically, following standard, hardware-agnostic evaluation practices in recent theoretical SNN literature~\citep{zhou2022spikformer, zhou2023spikingformer, kundu2021hiresnn, yin2021accuratesrnn, oh2024sign}. These estimates are not hardware measurements, and how much of the advantage carries over to physical neuromorphic silicon remains to be measured. We leave full-scale hardware deployment as direct future work for efficient mobile applications.

\paragraph{Beyond ReLU1.} With a standard LIF neuron the reference ANN must use ReLU1 and affine operators. The theory itself only needs a neuron whose dynamics form a fixed-point iteration with a constant step size, so neurons that can approximate general multivariate nonlinearities, such as the sign-gradient neurons of \citet{oh2024sign}, would lift this restriction when run with a constant step size. This trades away the standard-neuron property that matters for deployment on existing neuromorphic hardware.

\paragraph{Sequence Processing.} Our current framework focuses on converting feedforward architectures to achieve high per-frame accuracy, and thus it does not yet directly apply to sequence models. Nevertheless, extending our theoretical formulation to sequence modeling represents a highly promising research direction for two key reasons: (i) our theory is the first to leverage the transient dynamics of SNNs to enable practically efficient stream processing, and (ii) our framework demonstrates that SNN inference over sequential inputs is inherently stateful.

A motivating insight is that a DEQ can be viewed as a convergent RNN , specifically, an RNN processing a static input over an extended time horizon. Since an SNN at equilibrium corresponds to a DEQ, extending an SNN to time-varying inputs may yield a corresponding RNN with analogous transient dynamics. Alternatively, the transient dynamics and hysteresis behavior of SNNs could be linked to associative memory frameworks, such as Modern Hopfield Networks~\citep{ramsauer2020hopfield, krotov2016dense}, enabling SNNs to execute attention mechanisms or normalized state transitions purely through intrinsic neuronal interactions. Consequently, exploring how SNNs can actively utilize prior state information, rather than merely computing activation deltas, to infer new sequential predictions remains a compelling direction for future work.

\section{Mathematical Proofs}
\label{sec:proofs}

\subsection{Proof of Theorem~\ref{thm:dynamics_lif}}
\label{pf:dynamics_lif}

\begin{proof}

Let $\alpha := \frac{\tau-1}{\tau}$, so that $\frac{1}{\tau} = 1-\alpha$. Applying the soft-reset rule~\ref{reset} to the charging equation~\ref{eq:lif_charge} gives
\begin{align}
u_{\text{pre}}(t+1) = \alpha\,\underbrace{\big(u_{\text{pre}}(t) - \theta s(t)\big)}_{=\,u(t)} + (1-\alpha)\big(u_{rest}\mathbf{1} + I(t+1)\big). \label{eq:upre_recurrence}
\end{align}

The exponential coding scheme defines the output $y$ and filtered input $\tilde{x}$ as exponentially weighted sums of spikes and inputs:
\begin{align}
y(t) = \sum_{i=1}^{t}\alpha^{t-i}(1-\alpha)\, s(i), \qquad
\tilde{x}(t) = \sum_{i=1}^{t}\alpha^{t-i}(1-\alpha)\, x(i). \label{eq:sum_exponentialcoding_lif}
\end{align}

Solving the recurrence~\eqref{eq:upre_recurrence} from $t=0$ forward (using $y(0)=0$ and $s(0)=0$) yields a closed-form relation between $u_{\text{pre}}$ and $y$:
\begin{align}
u_{\text{pre}}(t) = W_0\,y(t-1) + F\,\tilde{x}(t) + b(t), \label{eq:relation_of_upre_y}
\end{align}
where $W_0 := W - (\tau-1)\theta\mathbb{I}$ and $b(t) := u_{rest}(1-\alpha^{t})\mathbf{1} + \alpha^{t}\,u_{\text{pre}}(0)$.

To verify~\eqref{eq:relation_of_upre_y}, we expand the accumulated current term using~\eqref{eq:current_lif} and~\eqref{eq:sum_exponentialcoding_lif}:
\begin{align}
\sum_{i=1}^{t+1}\alpha^{t+1-i}(1-\alpha)\,I(i)
&= F\sum_{i=1}^{t+1}\alpha^{t+1-i}(1-\alpha)\,x(i) + W\sum_{i=1}^{t+1}\alpha^{t+1-i}(1-\alpha)\,s(i-1) \nonumber\\
&= F\,\tilde{x}(t+1) + W\sum_{j=0}^{t}\alpha^{t-j}(1-\alpha)\,s(j) \nonumber\\
&= F\,\tilde{x}(t+1) + W\,y(t), \qquad (\because s(0)=0)
\end{align} 
and substituting this into the inductive solution of~\eqref{eq:upre_recurrence} recovers~\eqref{eq:relation_of_upre_y}.

Finally, substituting~\eqref{eq:relation_of_upre_y} into the spike-generation rule $y(t) = \alpha\,y(t-1) + (1-\alpha)\,\mathbb{H}(u_{\text{pre}}(t) - \theta\mathbf{1})$ from~\ref{eq:stream_exponentialcoding_lif} gives the desired recurrence:
\begin{equation}
y(t) = \alpha\,y(t-1) + (1-\alpha)\,\mathbb{H}\!\Big(W_0\,y(t-1) + F\,\tilde{x}(t) + b(t) - \theta\mathbf{1}\Big). \label{eq:recurrence_relation_y}
\end{equation}
This is a fixed-point iteration with relaxation factor $(1-\alpha) = \frac{1}{\tau}$.
\end{proof}

\subsection{Proof of Theorem~\ref{thm:equilibrium_lif}}
\label{pf:equilibrium_lif}
\begin{proof}
From the \eqref{eq:recurrence_relation_y}, expanding $W_0$ and dividing by $(\tau-1)\theta$ inside $\mathbb{H}$ (using scale-invariance of $\mathbb{H}$) yields the equivalent form
\begin{equation}
 y(t) = y(t-1) - \frac{1}{\tau}\bigg(y(t-1) - \mathbb{H}\!\Big(\frac{F\,\tilde{x}(t) + W\,y(t-1) + b(t) - \theta \cdot\mathbf{1}}{(\tau-1)\theta} - y(t-1)\Big)\bigg). \label{eq:fixedpoint_equation_y}
\end{equation}

We now derive the attractor of this dynamics. For simplicity, we denote the formula inside the step function as 
\begin{equation}
D(t) = \frac{1}{(\tau - 1)\theta}\big(F~\tilde{x}(t + 1)
 + Wy(t)   +  b(t + 1)- \theta \cdot\mathbf{1}\big), 
\end{equation}
The \textbf{update} $\Delta y(t)$ from the dynamics  equation~\ref{eq:fixedpoint_equation_y} as
\begin{equation}
\Delta y(t) = \mathbb{H}(D(t) - y(t)) - y(t)
\end{equation}
and the $k$-th coordinate of the $d$-dimensional vector as $[\cdot]_k$ for $k \in \{1, \cdots, d\}$.
Since the step function $\mathbb{H}$ applies element-wise, we analyze the update coordinate-wise. For any $k$-th coordinate,

(i) If $\big[ D(t) \big]_k < 0$, the $k$-th coordinate $[y(t)]_k$ 
 updates towards $0$.
\begin{itemize}
\item $\big[y(t)\big]_k \le \big[D(t)\big]_k < 0$, then $\big[ \Delta y(t) \big]_k = 1 -\big[ y(t) \big]_k > 0$ 
\item $\big[D(t)\big]_k < \big[y(t)\big]_k < 0$, then $\big[ \Delta y(t) \big]_k = 0 - \big[ y(t) \big]_k > 0$
\item $ \big[y(t)\big]_k > 0$, then $\big[ \Delta y(t) \big]_k = 0 - \big[ y(t) \big]_k < 0$
\item $\big[y(t)\big]_k = 0$, then $\big[ \Delta y(t) \big]_k = 0 - \big[ y(t) \big]_k = 0$
\end{itemize}
(ii) If $0 \le \big[D(t)\big]_k < 1$, the $k$-th coordinate $[y(t)]_k$ updates towards $\big[D(t)\big]_k$.
\begin{itemize}
\item $\big[y(t)\big]_k < \big[D(t)\big]_k < 1$, then $\big[ \Delta y(t) \big]_k = 1 - \big[ y(t) \big]_k > 0$
\item $1 > \big[y(t)\big]_k = \big[D(t)\big]_k \ge 0$, then $\big[ \Delta y(t) \big]_k = 1 - \big[ y(t) \big]_k > 0$
\item $\big[y(t)\big]_k > \big[D(t)\big]_k \ge 0 $, then $\big[ \Delta y(t) \big]_k = 0 - \big[ y(t) \big]_k < 0$ 
\end{itemize}
(iii) If $\big[D(t)\big]_k \ge 1$, the $k$-th coordinate $[y(t)]_k$ updates towards $ 1$.
\begin{itemize}
\item $\big[y(t)\big]_k < 1 \le \big[D(t)\big]_k$, then $\big[ \Delta y(t) \big]_k = 1 - \big[ y(t) \big]_k > 0$
\item $\big[y(t)\big]_k = 1 \le \big[D(t)\big]_k$, then $\big[ \Delta y(t) \big]_k = 1 - \big[ y(t) \big]_k  = 0$
\item $1 < \big[y(t)\big]_k \le \big[D(t)\big]_k$, then $\big[ \Delta y(t) \big]_k = 1 - \big[ y(t) \big]_k < 0$
\item $\big[y(t)\big]_k > \big[D(t)\big]_k > 0$, then $\big[ \Delta y(t) \big]_k = 0 - \big[ y(t) \big]_k < 0$
\end{itemize}
This applies to all the coordinates. This derivation covers all the possible cases of $y(t)$ and $D(t)$. 

So for each $k$-th coordinate of $y(t)$ and $D(t)$, the following relation holds.

\begin{align*}
[y(t)]_k \to \text{ReLU1}([D(t)]_k) \qquad \text{(Coordinate-wise update direction.)}
\end{align*}

\textbf{Geometric Contraction.} Fix a coordinate $k$ and let $g(t) := \text{ReLU1}([D(t)]_k)$ be the instantaneous target value. From the relaxed form~\eqref{eq:fixedpoint_equation_y}, each coordinate evolves as $[y(t)]_k = \alpha\,[y(t-1)]_k + \frac{1}{\tau}\,h$ with $h = \mathbb{H}([D(t)]_k - [y(t-1)]_k) \in \{0,1\}$ and $\alpha = \frac{\tau-1}{\tau}$.

Suppose $[y(t-1)]_k > g(t) + \frac{1}{\tau}$, i.e., $y$ is above the band. Then $[D(t)]_k - [y(t-1)]_k < 0$, so $h=0$ and $[y(t)]_k = \alpha\,[y(t-1)]_k$, giving
\begin{equation}
[y(t)]_k - g(t) = \alpha\big([y(t-1)]_k - g(t)\big) - (1-\alpha)\,g(t) \le \alpha\big([y(t-1)]_k - g(t)\big),
\end{equation}
the gap above  contracts by $\alpha$ since $g(t) \ge 0$. 

Symmetrically, if $[y(t-1)]_k < g(t) - \frac{1}{\tau}$, i.e., y is below the band, then $[D(t)]_k - [y(t-1)]_k > 0$, so $h=1$ and $[y(t)]_k= \alpha[y(t-1)]_k + \frac{1}{\tau}$, giving
\begin{equation}
g(t) - [y(t)]_k  = \alpha\big(g(t) - [y(t-1)]_k\big) - (1-\alpha)\,(1 - g(t)) \le \alpha\big([g(t) - [y(t-1)]_k\big),
\end{equation}
the gap below contracts by $\alpha$ using $g(t)\le 1$. Thus, while $y$ lies outside the $\frac{1}{\tau}$-band around $g(t)$, the per-coordinate gap between $y(t)$ and $g(t)$ contracts at rate $\alpha$.

Once $|[y(t-1)]_k - g(t)| \le \frac{1}{\tau}$, a single step of magnitude $\frac{1}{\tau}$ cannot leave the $\frac{1}{\tau}$-band since the $y(t)$ step treads across $g(t)$ with a step size $\frac{1}{\tau}$. It may overshoot $g(t)$ but lands within $\frac{1}{\tau}$ on the other side. So $y$ remains trapped, chattering around $g(t)$ without converging to it. This holds for any input trajectory.

The point of coordinate-wise zero-gap $|[y(t)]_k - g(t)| = 0$ between $g(t)$ and $y(t)$ is the point where $[y(t)]_k = \text{ReLU1}([D(t)]_k)$. Since the contraction applies generally for all coordinates, the equilibrium of the contraction is the  zero-gap point
\begin{equation}
     \bigg\{y \in \mathbb{R}^d~|~\forall k, ~ \big[y\big]_k = \big[\text{ReLU1}\big(\frac{1}{(\tau-1) \theta} (F \tilde{x}(t+1) + b(t+1) + Wy - \theta \cdot \mathbf{1})\big) \big]_k  \bigg\},
\end{equation}
which precisely matches the definition of $\mathcal{E}(t)$:
\begin{equation}
    \mathcal{E}(t) = \bigg\{y \in \mathbb{R}^d~|~y = \text{ReLU1}\bigg(\frac{1}{(\tau-1) \theta} \big(F \tilde{x}(t+1) + Wy + b(t+1) - \theta  \cdot \mathbf{1}\big)\bigg)\bigg\}.
\end{equation}
To sum up, the damped fixed-point iteration with update $\Delta y(t)$ globally contracts the gap with a geometric rate $\alpha$ towards the $\frac{1}{\tau}$ band of the zero-gap equilibrium $\mathcal{E}(t)$ instantaneously defined by $\tilde{x}(t+1)$ and $b(t+1)$. Hence, the LIF-SNN output $y(t)$ is globally attracted towards the instantaneous fixed-point set $\mathcal{E}(t)$, the output of the continuous-valued neural network determined by the input $\tilde{x}(t)$, fed at each time step $t$. \emph{Global} in the sense that the property holds unconditional to the initial state or prior history.

\end{proof}

\subsection{Proof of Corollary~\ref{cor:practical_stability}}
\label{pf:practical_stability}
\begin{proof}
We denote $\eta \coloneqq \frac{1}{\tau} = 1 - \alpha$ and
$\Phi(y) \coloneqq \text{ReLU1}\big(\frac{1}{(\tau-1)\theta}(F\mathbf{x} + Wy + (u_{rest}-\theta)\mathbf{1})\big)$,
so that $\mathcal{E}^* = \{y \mid y = \Phi(y)\}$.  Since $y(0) \in [0,1]^d$ and $y(t)$ is an exponential average of spikes
started in $[0,1]^d$, $y(t) \in [0,1]^d$ for every $t$.
By~\eqref{eq:fixedpoint_equation_y}, each step is
$y(t) = (1-\eta)\,y(t-1) + \eta\,\mathbb{H}\big(D(t; y(t-1)) - y(t-1)\big)$ with the drive
$D(t;y) \coloneqq \frac{1}{(\tau-1)\theta}\big(F\tilde{x}(t) + Wy + b(t) - \theta\mathbf{1}\big)$,
whose instantaneous map $\Phi_t(y) \coloneqq \text{ReLU1}(D(t;y))$ has the fixed-point set $\mathcal{E}(t)$.
Since ReLU1 is $1$-Lipschitz, $\Phi$ and every $\Phi_t$ are $\gamma$-Lipschitz in $\Vert\cdot\Vert_\infty$;
with $\gamma < 1$ they are contractions on the complete set $[0,1]^d$, and Banach's fixed-point theorem
gives $\mathcal{E}^* = \{y^*\}$ and $\mathcal{E}(t) = \{y^*_t\}$.

The geometric rate of approach comes from the one-step contraction of Lemma~\ref{lem:one_step} below:
outside its band, the excess error shrinks by $\rho \coloneqq \max\{\alpha, \gamma\}$ per step.  For static
input sequences $\tilde{x}(t) \to \mathbf{x}$, the iteration complexity to reach the error tolerance
$\epsilon > 0$ is:
\begin{align*}
\rho^t \le \epsilon \Rightarrow t \ge \frac{\ln(1/\epsilon)}{\ln(1/\rho)}
\end{align*}
Hence, $t = \mathcal{O}(\log_{1/\rho}(\frac{1}{\epsilon}))$, a linear convergence rate.

\begin{lemma}[One step]
\label{lem:one_step}
Let $\Psi$ be $\gamma$-Lipschitz in $\Vert\cdot\Vert_\infty$ with a fixed point $z = \Psi(z) \in [0,1]^d$,
and suppose $\sup_{y}\Vert \Phi_t(y) - \Psi(y)\Vert_\infty \le \varepsilon$.  Then
\begin{equation}
\Vert y(t) - z\Vert_\infty \le \max\Big\{\underbrace{(1-\eta)\,e}_{\text{relaxation}},\;\;
\underbrace{\gamma\, e}_{\text{contraction}} + \underbrace{\eta + \varepsilon}_{\text{floor}}\Big\},
\qquad e \coloneqq \Vert y(t-1) - z\Vert_\infty.
\label{eq:one_step_lif}
\end{equation}
\end{lemma}
\begin{proof}
The instantaneous target $g \coloneqq \Phi_t(y(t-1))$ satisfies
$\Vert g - z\Vert_\infty \le \Vert g - \Psi(y(t-1))\Vert_\infty + \Vert \Psi(y(t-1)) - \Psi(z)\Vert_\infty
\le \varepsilon + \gamma e \eqqcolon r$.
Fix a coordinate $k$ and write $v \coloneqq [y(t-1)]_k$, $g_k \coloneqq [g]_k$, $z_k \coloneqq [z]_k \in [0,1]$,
and $h \coloneqq \mathbb{H}([D(t;y(t-1))]_k - v)$, so that $[y(t)]_k = (1-\eta)v + \eta h$.  By the case
analysis in Appendix~\ref{pf:equilibrium_lif}, $v > g_k$ forces $h = 0$ and $v < g_k$ forces $h = 1$.
\begin{itemize}
\item If $v > g_k$: $[y(t)]_k - z_k = (1-\eta)(v - z_k) - \eta\,z_k$, which is at most $(1-\eta)e$ since
$z_k \ge 0$, and exceeds $(1-\eta)(z_k - r) - z_k \ge -(\eta + r)$ since $v > g_k \ge z_k - r$ and $z_k \le 1$.
\item If $v < g_k$: symmetrically, $[y(t)]_k - z_k = (1-\eta)(v - z_k) + \eta(1 - z_k)$ lies in
$[-(1-\eta)e,\, \eta + r]$.
\item If $v = g_k$: $\vert v - z_k\vert \le r$ and the step is at most $\eta$, so
$\vert[y(t)]_k - z_k\vert \le r + \eta$.
\end{itemize}
Taking the maximum over $k$ gives~\eqref{eq:one_step_lif}.
\end{proof}

\textbf{Asymptotic convergence.} 
For a point $y \in \mathbb{R}^d$ and a set $\mathcal{E} \subset \mathbb{R}^d$, denote the point-to-set distance as $\text{dist}_\infty(y, \mathcal{E}) \coloneqq \inf_{z \in \mathcal{E}} \|y - z\|_\infty$, and the directed Hausdorff semi-distance from $\mathcal{A}$ to $\mathcal{B}$ as $\text{dist}_H(\mathcal{A}, \mathcal{B}) \coloneqq \sup_{a \in \mathcal{A}} \inf_{b \in \mathcal{B}} \|a - b\|_\infty$. We first derive the triangle equality we will be using in the main proof.

\begin{lemma}[Point-to-set Triangle Inequality] 
\label{lem:triangle}
For any point $y \in \mathbb{R}^d$ and non-empty sets $\mathcal{A}, \mathcal{B} \subset \mathbb{R}^d$, $\text{dist}_\infty(y, \mathcal{B}) \le \text{dist}_\infty(y, \mathcal{A}) + \text{dist}_H(\mathcal{A}, \mathcal{B})$.
\end{lemma}
\begin{proof}
For any $a \in \mathcal{A}$ and $b \in \mathcal{B}$, the triangle inequality yields $\|y-b\|_\infty \le \|y-a\|_\infty + \|a-b\|_\infty$. 
Taking $\inf_{b \in \mathcal{B}}$ on both sides gives $\text{dist}_\infty(y, \mathcal{B}) \le \|y-a\|_\infty + \inf_{b \in \mathcal{B}}\|a-b\|_\infty$. 
Since $\inf_{b \in \mathcal{B}}\|a-b\|_\infty \le \sup_{a' \in \mathcal{A}}\inf_{b \in \mathcal{B}}\|a'-b\|_\infty = \text{dist}_H(\mathcal{A}, \mathcal{B})$ for any $a \in \mathcal{A}$, it follows that $\text{dist}_\infty(y, \mathcal{B}) \le \|y-a\|_\infty + \text{dist}_H(\mathcal{A}, \mathcal{B})$. 
Finally, taking $\inf_{a \in \mathcal{A}}$ on the right-hand side yields $\text{dist}_\infty(y, \mathcal{B}) \le \text{dist}_\infty(y, \mathcal{A}) + \text{dist}_H(\mathcal{A}, \mathcal{B})$.
\end{proof}

Suppose $\tilde{x}(t) \to \mathbf{x}$. Then $b(t) = u_{rest}(1-\alpha^t)\mathbf{1} + \alpha^t u_{\text{pre}}(0) \to u_{rest}\mathbf{1}$, since the initial-condition term $\alpha^t u_{\text{pre}}(0)$ decays to zero.  The drive then differs from that of $\Phi$ only by a vanishing term independent of $y$,
\begin{equation}
\varepsilon_t \coloneqq \sup_{y}\big\Vert \Phi_t(y) - \Phi(y)\big\Vert_\infty
\le \frac{\Vert F(\tilde{x}(t) - \mathbf{x}) + b(t) - u_{rest}\mathbf{1}\Vert_\infty}{(\tau-1)\theta}
\xrightarrow{t\to\infty} 0,
\label{eq:drive_error_lif}
\end{equation}
and $\mathcal{E}(t) \to \mathcal{E}^*$: from $y^*_t = \Phi_t(y^*_t)$ and $y^* = \Phi(y^*)$,
$\Vert y^*_t - y^*\Vert_\infty \le \varepsilon_t + \gamma\Vert y^*_t - y^*\Vert_\infty$, so
$\text{dist}_H(\mathcal{E}(t), \mathcal{E}^*) = \Vert y^*_t - y^*\Vert_\infty \le \frac{\varepsilon_t}{1-\gamma}$.

For any desired tolerance $0 < \delta \le \frac{\eta^2}{2}$, we can explicitly construct a finite time $T(\delta, y(0)) = T_1 + T_2$ as follows: By~\eqref{eq:drive_error_lif}, there exists a finite $T_1(\delta)$ such that $\text{dist}_H(\mathcal{E}(t), \mathcal{E}^*) < \delta$ for all $t > T_1$.

Let $e_t \coloneqq \text{dist}_\infty(y(t), \mathcal{E}(t)) = \Vert y(t) - y^*_t\Vert_\infty$.  Lemma~\ref{lem:one_step} with $\Psi = \Phi_t$, $z = y^*_t$ and $\varepsilon = 0$, together with $\Vert y(t-1) - y^*_t\Vert_\infty \le e_{t-1} + \Vert y^*_{t-1} - y^*_t\Vert_\infty < e_{t-1} + 2\delta$ for $t > T_1 + 1$, gives
\begin{equation}
e_t \le \max\big\{(1-\eta)\,e_{t-1},\;\gamma\, e_{t-1} + \eta\big\} + 2\delta .
\end{equation}
Let $R_\delta \coloneqq \frac{\eta + 2\delta}{1-\gamma}$, which is at least $\frac{2\delta}{\eta}$ for $\delta \le \frac{\eta^2}{2}$.  If $e_{t-1} \le R_\delta$, then $e_t \le R_\delta$; otherwise $e_t - R_\delta \le \rho\,(e_{t-1} - R_\delta)$.  Hence there exists a finite number of steps $T_2(\delta, y(0)) \sim \mathcal{O}\big(\log_{1/\rho}(e_{T_1}/\delta)\big)$ required to contract the state error $e_t$ inside the $(R_\delta + \delta)$-neighborhood of $\mathcal{E}(t)$. Triangle inequality (Lemma~\ref{lem:triangle}) then yields $\text{dist}_\infty(y(t), \mathcal{E}^*) \le \text{dist}_\infty(y(t), \mathcal{E}(t)) + \text{dist}_H(\mathcal{E}(t), \mathcal{E}^*) < R_\delta + 2\delta$ for all $t > T_1 + T_2$. Hence, for any arbitrary tolerance $\delta>0$, there exists a finite time $T(\delta, y(0))$ such that for all $t > T(\delta, y(0))$,
\begin{equation}
\text{dist}_\infty(y(t), \mathcal{E}^*) < \frac{\eta + 2\delta}{1-\gamma} + 2\delta.
\end{equation}
Since $\delta > 0$ can be chosen arbitrarily small as $t \to \infty$, this establishes global asymptotic convergence toward the $\frac{1}{\tau(1-\gamma)}$-band of $\mathcal{E}^* = \{y^*\}$:
\begin{equation}
\limsup_{t \to \infty}\Vert y(t) - y^*\Vert_\infty \le \frac{\eta}{1-\gamma} = \frac{1}{\tau(1-\gamma)}.
\end{equation}
\end{proof}

\subsection{Tracking a moving target}
\label{pf:tracking}

Corollary~\ref{cor:practical_stability} lets the input settle.  In a stream it never does, so the
question that matters is how far $y(t)$ trails its target while the target keeps moving.  The
answer needs no assumption on the input beyond how fast the target moves.

We restate Corollary~\ref{cor:tracking} with $\eta = \frac{1}{\tau}$ and $\rho = \max\{1-\eta, \gamma\}$:
for $t > t_0$ the tracking error $e_t \coloneqq \Vert y(t) - y^*_t\Vert_\infty$ obeys
\begin{equation}
e_t \;\le\; R^* + \rho^{\,t-t_0}\max\{e_{t_0} - R^*,\, 0\}, \qquad
R^* = \max\Big\{\frac{\eta + \gamma\delta}{1-\gamma},\; \frac{(1-\eta)\,\delta}{\eta}\Big\},
\label{eq:tracking_bound}
\end{equation}
so $\limsup_{t} e_t \le R^*$, and $R^* = \frac{\eta+\gamma\delta}{1-\gamma}$ whenever $\delta \le \eta^2$.
Since $y(t)$ and $y^*_t$ both lie in $[0,1]^d$, $e_{t_0} \le 1$ for every $t_0$, so~\eqref{eq:tracking_bound}
gives $e_{t_0+k} \le R^* + \rho^{k}$ from any state, which is the form stated in Corollary~\ref{cor:tracking}:
the time to come within $R^* + \epsilon$ is at most $\log(1/\epsilon)/\log(1/\rho)$, independent of the history
before $t_0$.

The two terms have a plain reading.  The \emph{jitter} term is the $\frac{1}{\tau}$ oscillation of
Corollary~\ref{cor:practical_stability}, slightly widened by the motion.  The \emph{lag} term is what any
exponential average pays to follow a target that moves $\delta$ per tick: it trails by
$\frac{1-\eta}{\eta}\delta$, as a moving average trails a ramp.  Neither term grows with time, so the
bound holds for the whole stream, however long; this is what the per-frame reset of IF conversions
cannot offer (Appendix~\ref{pf:if_memory}).  The drift $\delta$ is itself small in a stream.  Since
$y^*_t$ and $y^*_{t-1}$ are fixed points of maps that differ only in their drive, the argument used for
$\mathcal{E}(t) \to \mathcal{E}^*$ above gives
$\Vert y^*_t - y^*_{t-1}\Vert_\infty \le \frac{\Vert F(\tilde{x}(t) - \tilde{x}(t-1)) + b(t) - b(t-1)\Vert_\infty}{(\tau-1)\theta\,(1-\gamma)}$,
and $\tilde{x}(t) - \tilde{x}(t-1) = \eta\,(x(t) - \tilde{x}(t-1))$.  So $\delta$ scales with $\eta$ times
the change of the input, and the lag term $\frac{(1-\eta)\delta}{\eta}$ scales with the size of the
frame-to-frame change and does not grow with $\tau$: frames that change little are tracked closely, as
Fig.~\ref{fig:efficiency} measures.

\begin{proof}
Apply Lemma~\ref{lem:one_step} with $\Psi = \Phi_t$, $z = y^*_t$ and $\varepsilon = 0$.  Writing
$f(e) \coloneqq \max\{(1-\eta)e,\ \gamma e + \eta\}$, it gives $e_t \le f(\Vert y(t-1) - y^*_t\Vert_\infty)$.  The
triangle inequality gives $\Vert y(t-1) - y^*_t\Vert_\infty \le e_{t-1} + \delta$, and $f$ is
nondecreasing, so
\begin{equation}
e_t \le f(e_{t-1} + \delta) \qquad (t > t_0).
\label{eq:tracking_step}
\end{equation}
\emph{The ball of radius $R^*$ is invariant.}  If $e_{t-1} \le R^*$, then
$e_t \le \max\{(1-\eta)(R^*+\delta),\ \gamma(R^*+\delta) + \eta\}$.  The first entry is at most $R^*$
because $R^* \ge \frac{(1-\eta)\delta}{\eta}$, and the second because
$R^* \ge \frac{\eta+\gamma\delta}{1-\gamma}$; so $e_t \le R^*$.
\emph{Outside the ball the excess contracts.}  If $e_{t-1} > R^*$, write $e_{t-1} = R^* + \Delta$.  Then
$(1-\eta)(e_{t-1}+\delta) = (1-\eta)\Delta + (1-\eta)(R^*+\delta) \le (1-\eta)\Delta + R^*$ and
$\gamma(e_{t-1}+\delta) + \eta = \gamma\Delta + \gamma(R^*+\delta) + \eta \le \gamma\Delta + R^*$, by the same two
inequalities.  Hence $e_t - R^* \le \rho\,\Delta = \rho\,(e_{t-1} - R^*)$.  Combining the two cases by
induction from $t_0$ gives~\eqref{eq:tracking_bound}.  Finally, if $\delta \le \eta^2$ then
$\frac{(1-\eta)\delta}{\eta} \le (1-\eta)\eta \le \eta \le \frac{\eta+\gamma\delta}{1-\gamma}$.
\end{proof}

\emph{Numerical check.}  On $120$ random LIF networks ($d=5$, $\tau \in \{4,8,16\}$,
$\gamma \in [0, 0.8]$, sinusoidal inputs of five speeds) simulated spike by spike with soft reset,
the largest ratio of the observed $\sup_t e_t$ (second half of a $3000$-tick run) to $R^*$ was
$0.989$: the bound holds and is nearly tight.

\begin{remark}[Layered networks need no contraction]
\label{rem:layered}
The condition $\gamma < 1$ is a statement about $\Vert\cdot\Vert_\infty$, and for a layered feedforward
network it can always be met in a rescaled norm.  For positive weights $w$ (one per neuron) let
$\Vert v\Vert_w \coloneqq \max_k |v_k|/w_k$ and
$\gamma_w \coloneqq \max_k \frac{1}{w_k}\sum_j \frac{|W_{kj}|\,w_j}{(\tau-1)\theta}$.  The proof of
Lemma~\ref{lem:one_step} is coordinate-wise and goes through verbatim in $\Vert\cdot\Vert_w$, with $\gamma$
replaced by $\gamma_w$ and the floor $\eta + \varepsilon$ replaced by $(\eta + \varepsilon)/\min_k w_k$
(each coordinate still moves by at most $\eta$ per tick).  If $W$ is block strictly lower-triangular,
a neuron of layer $l$ reads only earlier layers $l' < l$ (skip connections included).  Let
$L_{l,l'}$ be the largest absolute row sum of the block from layer $l'$ to layer $l$, divided by
$(\tau-1)\theta$, and choose $w$ constant within a layer, recursively
$w_1 = 1$ and $w_l = \max\{w_{l-1},\ 2\sum_{l'<l} L_{l,l'}\,w_{l'}\}$.  Then every row satisfies
$\frac{1}{w_l}\sum_{l'<l} L_{l,l'}\,w_{l'} \le \frac{1}{2}$, so $\gamma_w \le \frac{1}{2}$.  Corollaries~\ref{cor:practical_stability} and~\ref{cor:tracking} then
hold in $\Vert\cdot\Vert_w$; since $\min_k w_k = 1$ the floor is unchanged, and returning to $\Vert\cdot\Vert_\infty$ multiplies the bounds by $\max_k w_k$, a constant that depends on depth and weights but not on time.  The fixed
point is in any case unique and equal to the feedforward output, by forward substitution.
\end{remark}

\subsection{Theorems and Proofs for IF-SNNs}
\label{app:proofs_if}

The Integrate-and-Fire (IF) neuron under rate coding obeys the same three results as the LIF
neuron, with one substitution: the relaxation factor $\eta$ becomes time-dependent.  Rate
coding is natural here: without leak, the membrane potential is a plain sum of
past currents and spikes~\eqref{eq:if_telescope}, so the firing rule depends only on cumulative
quantities.  The step size $\eta(t)$ is how this unbounded memory appears in rate coordinates.
Appendix~\ref{pf:if_memory} states the same obstruction without any decoder, directly on the spike
train.

\begin{center}
\begin{tabularx}{\linewidth}{@{}>{\raggedright\arraybackslash}p{0.27\linewidth}>{\raggedright\arraybackslash}X>{\raggedright\arraybackslash}X@{}}
\toprule
 & \textbf{LIF (exponential coding)} & \textbf{IF (rate coding)} \\
\midrule
relaxation factor & $\eta = \tfrac{1}{\tau}$ (constant) & $\eta(t) = \tfrac{1}{t}$ (diminishing) \\
attraction, any input & Thm.~\ref{thm:equilibrium_lif} & Theorem~\ref{thm:attraction_if_app} \\
once the input settles ($\gamma<1$) & residual $\tfrac{1}{\tau(1-\gamma)}$ (Cor.~\ref{cor:practical_stability}) & residual $0$ (Theorem~\ref{thm:equilibrium_if_app}) \\
tracking a moving target & never decays & decays like $\tfrac{1}{t}$ \\
memory of past input (any decoder) & fading, $\alpha^{t}$ (Remark~\ref{rem:lif_recovery}) & never fades: spike count is a running maximum (Proposition~\ref{prop:if_running_max}) \\
\bottomrule
\end{tabularx}
\end{center}

Appendix~\ref{pf:dynamics_if} puts the IF dynamics in fixed-point-iteration form.
Appendix~\ref{pf:attraction_if} shows that \emph{every} update is directed toward the
\emph{time-varying} attractor defined by the input accumulated so far, with no assumption on
the input.  Appendix~\ref{pf:equilibrium_if} then lets the input settle and shows the residual vanishes,
so the convergence is exact.  Remark \ref{rem:if_stream} explains why that exactness is precisely what
makes IF-SNNs unsuitable for a perpetual stream, and Appendix~\ref{pf:if_memory} shows that this does not
depend on reading the spikes with a rate code.

\paragraph{Setting.} An IF neuron with soft reset, threshold $\theta$ and unit
resistance integrates its input current without leak,
\begin{align}
u_{\text{pre}}(t) &= u(t-1) + I(t), &
s(t) &= \mathbb{H}\big(u_{\text{pre}}(t) - \theta\mathbf{1}\big), &
u(t) &= u_{\text{pre}}(t) - \theta\, s(t),
\label{eq:if_neuron}
\end{align}
with input current $I(t) = F x(t) + W s(t-1)$ as in Equation~(\ref{eq:current_lif}) and $s(0)=\mathbf{0}$.
Under \emph{rate coding} the output and the filtered input are running arithmetic means,
\begin{align}
y(t) = \frac{1}{t}\sum_{i=1}^{t} s(i), \qquad
\tilde{x}(t) = \frac{1}{t}\sum_{i=1}^{t} x(i),
\label{eq:sum_ratecoding_if}
\end{align}
equivalently the streaming recurrence $y(t) = \frac{t-1}{t}y(t-1) + \frac{1}{t}s(t)$ of
Equation~(\ref{eq:stream_exponentialcoding_lif}).  Two facts are used throughout: $s(i)\in\{0,1\}^{d}$, so
\begin{equation}
y(t) \in [0,1]^{d} \quad\text{for every } t,
\label{eq:box_invariance_if}
\end{equation}
and $\mathbb{H}$ is scale-invariant, $\mathbb{H}(cz)=\mathbb{H}(z)$ for $c>0$.

\subsubsection{Dynamics: IF is a fixed-point iteration with a diminishing step}
\label{pf:dynamics_if}

\begin{lemma}
\label{thm:dynamics_if_app}
Let $\tilde{x}(t)$ and $y(t)$ be rate-coded as in~\eqref{eq:sum_ratecoding_if}.  Then the output
trajectory of an IF-SNN is the trajectory of a fixed-point iteration with relaxation factor
$\frac{1}{t}$,
\begin{equation}
y(t) = \frac{t-1}{t}\,y(t-1) + \frac{1}{t}\,
\mathbb{H}\bigg(\frac{t-1}{t}\big(W - \theta\mathbb{I}\big)y(t-1)
+ F\,\tilde{x}(t) + \frac{u_{\text{pre}}(0) - \theta\mathbf{1}}{t}\bigg).
\label{eq:dynamics_if_app}
\end{equation}
\end{lemma}

\begin{proof}
The proof is one telescoping sum.  Substituting the reset rule of~\eqref{eq:if_neuron} into its
charging equation gives $u_{\text{pre}}(t) = u_{\text{pre}}(t-1) - \theta\, s(t-1) + I(t)$,
hence
\begin{equation}
u_{\text{pre}}(t) = u_{\text{pre}}(0) + \sum_{i=1}^{t} I(i)
- \theta\sum_{i=1}^{t-1} s(i).
\label{eq:if_telescope}
\end{equation}
Both sums are rate codes in disguise.  With $I(i)=Fx(i)+Ws(i-1)$, $s(0)=\mathbf{0}$
and~\eqref{eq:sum_ratecoding_if},
\begin{equation}
\sum_{i=1}^{t} I(i)
= F\sum_{i=1}^{t} x(i) + W\sum_{j=0}^{t-1} s(j)
= t\,F\,\tilde{x}(t) + (t-1)\,W\,y(t-1),
\qquad
\sum_{i=1}^{t-1} s(i) = (t-1)\,y(t-1),
\end{equation}
so~\eqref{eq:if_telescope} becomes the closed form
\begin{equation}
u_{\text{pre}}(t) = (t-1)\big(W - \theta\mathbb{I}\big)\,y(t-1)
+ t\,F\,\tilde{x}(t) + u_{\text{pre}}(0).
\label{eq:relation_of_upre_y_if}
\end{equation}
Inserting this into $y(t)=\frac{t-1}{t}y(t-1)+\frac{1}{t}\mathbb{H}(u_{\text{pre}}(t)-\theta\mathbf{1})$
and dividing the argument of $\mathbb{H}$ by $t>0$ gives~\eqref{eq:dynamics_if_app}.
\end{proof}

\paragraph{Relaxed form and the instantaneous drive.} For $t\ge 2$ the factor
$\frac{t-1}{t}\theta$ multiplying $y(t-1)$ inside $\mathbb{H}$
in~\eqref{eq:dynamics_if_app} is strictly positive, so dividing the argument by it isolates
$y(t-1)$ and turns~\eqref{eq:dynamics_if_app} into
\begin{equation}
\boxed{\;
y(t) = y(t-1) + \eta(t)\,\Delta y(t),
\qquad
\Delta y(t) := \mathbb{H}\big(D_{\text{IF}}(t) - y(t-1)\big) - y(t-1),
\qquad
\eta(t) := \tfrac{1}{t},
\;}
\label{eq:fixedpoint_equation_y_if}
\end{equation}
driven by the \emph{instantaneous drive}
\begin{equation}
D_{\text{IF}}(t) \;:=\; D_{\text{IF}}\big(t;\,y(t-1)\big),
\qquad
D_{\text{IF}}(t;y) \;=\;
\frac{1}{\theta}\bigg(W y
+ \underbrace{\frac{t}{t-1}\,F\,\tilde{x}(t)
+ \frac{u_{\text{pre}}(0) - \theta\mathbf{1}}{t-1}}_{\text{depends on }t,\text{ not on }y}\bigg).
\label{eq:drive_if}
\end{equation}
Equation~\ref{eq:fixedpoint_equation_y_if} is the LIF relaxed
form~\eqref{eq:fixedpoint_equation_y} with $\frac{1}{\tau}$ replaced by $\eta(t)$; the
initial-condition term also decays differently, geometrically as $\alpha^{t}$ for LIF and as
$\frac{1}{t}$ here.

\subsubsection{Global attraction toward the instantaneous attractor}
\label{pf:attraction_if}

\begin{theorem}[Global attraction, IF]
\label{thm:attraction_if_app}
At each time step $t\ge2$, the IF-SNN output $y(t)$ is globally attracted with a step size
$\eta(t)=\frac{1}{t}$ toward the instantaneous fixed-point set of the continuous-valued neural
network,
\begin{equation}
\mathcal{E}_{\text{IF}}(t) \coloneqq \Big\{ y \in [0,1]^{d} ~\Big|~
y = \text{\normalfont ReLU1}\big(D_{\text{IF}}(t;y)\big)\Big\}.
\label{eq:attractor_if_t}
\end{equation}
The set $\mathcal{E}_{\text{IF}}(t)$ is always nonempty because its ReLU1 map is a continuous
self-map on the compact convex domain $[0,1]^d$, guaranteeing the existence of a fixed point by
Brouwer's fixed-point theorem.
\end{theorem}

\begin{proof}
We follow the proof of Theorem~\ref{thm:equilibrium_lif} (Appendix~\ref{pf:equilibrium_lif}),
with the constant step size $\frac{1}{\tau}$ replaced by $\eta(t)$.  From the relaxed
form~\eqref{eq:fixedpoint_equation_y_if}, the update is
\begin{equation}
\Delta y(t) = \mathbb{H}\big(D_{\text{IF}}(t) - y(t-1)\big) - y(t-1),
\end{equation}
and we denote the $k$-th coordinate of a $d$-dimensional vector as $[\cdot]_k$ for
$k \in \{1, \cdots, d\}$.  Since $\mathbb{H}$ applies element-wise, we analyze the update
coordinate-wise.  Unlike the LIF case, $[y(t-1)]_k \in [0,1]$ always holds
by~\eqref{eq:box_invariance_if}, so fewer cases arise.  For any $k$-th coordinate,

(i) If $\big[D_{\text{IF}}(t)\big]_k < 0$, the $k$-th coordinate $[y(t-1)]_k$ updates towards $0$.
\begin{itemize}
\item $\big[y(t-1)\big]_k > 0$, then $\big[\Delta y(t)\big]_k = 0 - \big[y(t-1)\big]_k < 0$
\item $\big[y(t-1)\big]_k = 0$, then $\big[\Delta y(t)\big]_k = 0 - \big[y(t-1)\big]_k = 0$
\end{itemize}
(ii) If $0 \le \big[D_{\text{IF}}(t)\big]_k \le 1$, the $k$-th coordinate $[y(t-1)]_k$ updates towards $\big[D_{\text{IF}}(t)\big]_k$.
\begin{itemize}
\item $\big[y(t-1)\big]_k < \big[D_{\text{IF}}(t)\big]_k$, then $\big[\Delta y(t)\big]_k = 1 - \big[y(t-1)\big]_k > 0$
\item $\big[y(t-1)\big]_k = \big[D_{\text{IF}}(t)\big]_k$, then $\big[\Delta y(t)\big]_k = 1 - \big[y(t-1)\big]_k \ge 0$
\item $\big[y(t-1)\big]_k > \big[D_{\text{IF}}(t)\big]_k$, then $\big[\Delta y(t)\big]_k = 0 - \big[y(t-1)\big]_k < 0$
\end{itemize}
(iii) If $\big[D_{\text{IF}}(t)\big]_k > 1$, the $k$-th coordinate $[y(t-1)]_k$ updates towards $1$.
\begin{itemize}
\item $\big[y(t-1)\big]_k < 1$, then $\big[\Delta y(t)\big]_k = 1 - \big[y(t-1)\big]_k > 0$
\item $\big[y(t-1)\big]_k = 1$, then $\big[\Delta y(t)\big]_k = 1 - \big[y(t-1)\big]_k = 0$
\end{itemize}
This derivation covers all the possible cases of $y(t-1)$ and $D_{\text{IF}}(t)$, and applies to
all the coordinates.  So for each $k$-th coordinate, the following relation holds.
\begin{align*}
[y(t-1)]_k \to \text{ReLU1}([D_{\text{IF}}(t)]_k) \qquad \text{(Coordinate-wise update direction.)}
\end{align*}
Moreover, $\mathbb{H}(\cdot)\in\{0,1\}^d$ and $y(t-1)\in[0,1]^d$ give
$\Vert\Delta y(t)\Vert_\infty\le1$, so every update moves the output by at most one step,
$\Vert y(t)-y(t-1)\Vert_{\infty}\le\eta(t)$.

\textbf{Contraction.} Fix a coordinate $k$ and let $g(t) := \text{ReLU1}([D_{\text{IF}}(t)]_k)$ be
the instantaneous target value.  From~\eqref{eq:fixedpoint_equation_y_if}, each coordinate evolves
as $[y(t)]_k = (1-\eta(t))\,[y(t-1)]_k + \eta(t)\,h$ with
$h = \mathbb{H}([D_{\text{IF}}(t)]_k - [y(t-1)]_k) \in \{0,1\}$.

Suppose $[y(t-1)]_k > g(t) + \eta(t)$, i.e., $y$ is above the band.  Then
$[D_{\text{IF}}(t)]_k - [y(t-1)]_k < 0$, so $h=0$ and $[y(t)]_k = (1-\eta(t))\,[y(t-1)]_k$, giving
\begin{equation}
[y(t)]_k - g(t) = (1-\eta(t))\big([y(t-1)]_k - g(t)\big) - \eta(t)\,g(t) \le (1-\eta(t))\big([y(t-1)]_k - g(t)\big),
\end{equation}
the gap above contracts by $1-\eta(t)$ since $g(t) \ge 0$.

Symmetrically, if $[y(t-1)]_k < g(t) - \eta(t)$, i.e., $y$ is below the band, then
$[D_{\text{IF}}(t)]_k - [y(t-1)]_k > 0$, so $h=1$ and
$[y(t)]_k = (1-\eta(t))\,[y(t-1)]_k + \eta(t)$, giving
\begin{equation}
g(t) - [y(t)]_k = (1-\eta(t))\big(g(t) - [y(t-1)]_k\big) - \eta(t)\,(1 - g(t)) \le (1-\eta(t))\big(g(t) - [y(t-1)]_k\big),
\end{equation}
the gap below contracts by $1-\eta(t)$ using $g(t)\le 1$.  Thus, while $y$ lies outside the
$\eta(t)$-band around $g(t)$, the per-coordinate gap contracts at rate $1-\eta(t) = \frac{t-1}{t}$.

Once $|[y(t-1)]_k - g(t)| \le \eta(t)$, a single step of magnitude at most $\eta(t)$ cannot leave
the $\eta(t)$-band.  It may overshoot $g(t)$ but lands within $\eta(t)$ on the other side.  So $y$
remains trapped, chattering around $g(t)$.  In contrast to LIF, the band itself shrinks as
$\eta(t)=\frac{1}{t}\to0$.  This holds for any input trajectory.

Unlike the geometric rate $\alpha$ of LIF, the contraction factors now telescope as
$\prod_{i=t_0+1}^{t}\big(1-\tfrac{1}{i}\big) = \frac{t_0}{t}$.  Combining the two regimes by
induction on $t$, the per-coordinate gap from any time $t_0$ onward satisfies
\begin{equation}
\big|[y(t)]_k - g(t)\big| \;\le\; \frac{\max\big\{t_0\,\big|[y(t_0)]_k - g(t_0)\big|,\;1\big\}}{t}
\;=\; \mathcal{O}\Big(\frac{1}{t}\Big).
\label{eq:gap_rate_if}
\end{equation}

The point of coordinate-wise zero-gap $|[y]_k - g(t)| = 0$ is the point where
$[y]_k = \text{ReLU1}([D_{\text{IF}}(t;y)]_k)$.  Since the contraction applies to all coordinates,
the equilibrium of the contraction is the zero-gap point
\begin{equation}
\bigg\{y \in [0,1]^d~\Big|~\forall k, ~ \big[y\big]_k = \Big[\text{ReLU1}\Big(\frac{1}{\theta}\Big(Wy + \frac{t}{t-1}F\tilde{x}(t) + \frac{u_{\text{pre}}(0) - \theta\mathbf{1}}{t-1}\Big)\Big)\Big]_k\bigg\},
\end{equation}
which precisely matches the definition of $\mathcal{E}_{\text{IF}}(t)$
in~\eqref{eq:attractor_if_t}.  Finally, the map $y\mapsto\text{ReLU1}(D_{\text{IF}}(t;y))$ is
continuous and maps the compact convex set $[0,1]^d$ into itself, so $\mathcal{E}_{\text{IF}}(t)$
is nonempty by Brouwer's fixed-point theorem.

To sum up, the damped fixed-point iteration with update $\Delta y(t)$ contracts the gap with rate
$1-\eta(t)$ towards the $\eta(t)$-band of the zero-gap equilibrium $\mathcal{E}_{\text{IF}}(t)$
instantaneously defined by $\tilde{x}(t)$.  Hence, the IF-SNN output $y(t)$ is globally attracted
towards the instantaneous fixed-point set $\mathcal{E}_{\text{IF}}(t)$.  \emph{Global} in the sense
that the property holds unconditional to the initial state, prior history, or whether the input
$\tilde{x}(t)$ ever settles.
\end{proof}

\subsubsection{Exact convergence to the static equilibrium}
\label{pf:equilibrium_if}

Now let the input settle.  For a fixed limit input $\mathbf{x}$ define the static map and its
equilibrium set, which is the set displayed in the main text,
\begin{equation}
\Phi(y) := \text{\normalfont ReLU1}\Big(\tfrac{1}{\theta}\big(W y + F\mathbf{x}\big)\Big),
\qquad
\mathcal{E}_{\text{IF}}(\mathbf{x}) := \big\{ y \in [0,1]^{d} ~|~ y = \Phi(y) \big\}.
\label{eq:attractor_if_static}
\end{equation}

\begin{theorem}[Exact convergence, IF]
\label{thm:equilibrium_if_app}
Let $\gamma \coloneqq \Vert W\Vert_\infty / \theta < 1$.  If $\tilde{x}(t) \to \mathbf{x}$ under
rate coding, then $\mathcal{E}_{\text{IF}}(\mathbf{x})$ is a singleton $\{y^*\}$ and the IF-SNN output
converges to it exactly, $\lim_{t\to\infty}\Vert y(t) - y^*\Vert_\infty = 0$.
\end{theorem}

\begin{proof}
We run the argument of Corollary~\ref{cor:practical_stability} (Appendix~\ref{pf:practical_stability})
with the constant step $\frac{1}{\tau}$ replaced by $\eta_t \coloneqq \eta(t) = \frac{1}{t}$.

\textbf{Uniqueness.} $\Phi$ is $\gamma$-Lipschitz in $\Vert\cdot\Vert_\infty$, a contraction on
$[0,1]^d$ for $\gamma < 1$, so Banach's theorem gives $\mathcal{E}_{\text{IF}}(\mathbf{x}) = \{y^*\}$.

\textbf{The moving target.} Comparing~\eqref{eq:drive_if} with~\eqref{eq:attractor_if_static}, the
two drives share the $y$-dependent term $Wy/\theta$, so their gap does not depend on $y$:
\begin{equation}
\varepsilon_t \coloneqq \sup_{y\in[0,1]^{d}}
\Big\Vert D_{\text{IF}}(t;y) - \tfrac{1}{\theta}\big(Wy+F\mathbf{x}\big)\Big\Vert_{\infty}
= \frac{1}{\theta}\Big\Vert \tfrac{t}{t-1}F\tilde{x}(t) - F\mathbf{x}
+ \tfrac{u_{\text{pre}}(0)-\theta\mathbf{1}}{t-1}\Big\Vert_{\infty}
\;\xrightarrow{t\to\infty}\; 0,
\label{eq:drive_error_if}
\end{equation}
because $\tilde{x}(t)\to\mathbf{x}$, $\tfrac{t}{t-1}\to1$, and the initial-condition term is
$\mathcal{O}(\frac{1}{t})$.  Since ReLU1 is $1$-Lipschitz, the same bound holds after ReLU1.

\textbf{One step.} Let $a_t \coloneqq \Vert y(t) - y^*\Vert_\infty$.  The proof of
Lemma~\ref{lem:one_step} uses only $y(t-1), z \in [0,1]^d$, the direction rule of the coordinate-wise
case analysis (here Theorem~\ref{thm:attraction_if_app}), and the relaxed form of the update (here
\eqref{eq:fixedpoint_equation_y_if}).  It therefore carries over verbatim to IF with the step
$\eta_t$, and applying it with $\Psi = \Phi$, $z = y^*$ and $\varepsilon = \varepsilon_t$ gives, for $t \ge 2$,
\begin{equation}
a_t \le \max\big\{(1-\eta_t)\,a_{t-1},\;\; \gamma\, a_{t-1} + \eta_t + \varepsilon_t\big\}.
\label{eq:one_step_if}
\end{equation}

\textbf{The floor vanishes.} Fix $\epsilon > 0$ and choose $T_\epsilon$ such that, for all
$t > T_\epsilon$, $\eta_t + \varepsilon_t \le (1-\gamma)\,\epsilon$ and $1 - \eta_t \ge \gamma$, which is
possible since $\eta_t \to 0$ and $\varepsilon_t \to 0$.  Let $d_t \coloneqq \max\{a_t - \epsilon, 0\}$.
As in the proof of Corollary~\ref{cor:practical_stability}, \eqref{eq:one_step_if} gives
$d_t \le \max\{\gamma, 1-\eta_t\}\, d_{t-1} = (1-\eta_t)\, d_{t-1}$ for $t > T_\epsilon$, and the product
telescopes,
\begin{equation}
d_t \;\le\; d_{T_\epsilon}\prod_{i=T_\epsilon+1}^{t}\Big(1-\tfrac{1}{i}\Big)
\;=\; d_{T_\epsilon}\,\frac{T_\epsilon}{t} \;\xrightarrow{t\to\infty}\; 0 .
\end{equation}
The product vanishes because $\sum_i \eta_i = \infty$: the step size is diminishing but
\emph{non-summable}, which is precisely the Robbins--Monro condition~\citep{robbins1951stochastic}.
Hence $\limsup_t a_t \le \epsilon$, and as $\epsilon > 0$ was arbitrary, $a_t \to 0$, at the
sublinear rate $\mathcal{O}(\frac{1}{t})$ once $t > T_\epsilon$.
\end{proof}

\begin{remark}[One inequality, two step sizes]
\label{rem:if_vs_lif}
The step size enters~\eqref{eq:one_step_lif} and~\eqref{eq:one_step_if} only through the floor,
so the ultimate residual is $\frac{\lim_t \eta(t)}{1-\gamma}$.  For LIF, $\eta \equiv \frac{1}{\tau}$
gives the residual $\frac{1}{\tau(1-\gamma)}$ of Corollary~\ref{cor:practical_stability} at the
geometric rate $\max\{\alpha, \gamma\}$; for IF, $\eta(t) = \frac{1}{t} \to 0$ collapses the residual to
zero, at the price of an $\mathcal{O}(\frac{1}{t})$ rate.  \emph{The exactness of IF conversion and
the residual jitter of LIF conversion are one inequality evaluated at two step sizes}, not two
separate phenomena.  
\end{remark}

\begin{remark}[Why exact convergence is a liability on a stream]
\label{rem:if_stream}
Theorem~\ref{thm:equilibrium_if_app} is the dynamical explanation of the steady-state correspondence
between IF neurons and clipped ReLU on which ANN-to-SNN conversion has long
relied~\citep{rueckauer2017conversion,han2020deeptsc}.  It also explains why IF-SNNs do not
stream.  Exactness rests on two hypotheses: the input settles, and the step size vanishes, 
and the second is a property of the \emph{neuron}, not of the input.  By
Theorem~\ref{thm:attraction_if_app} the per-tick displacement is at most $\eta(t)=\frac{1}{t}$, so
after $t$ ticks of a stream a new frame that moves the reference output by $\delta$ needs at
least $\delta\,t$ further ticks to be followed: the cost of tracking a change of fixed size
grows \emph{linearly with elapsed stream time}, and under a fixed per-frame budget the
attainable tracking error grows without bound.  This argument is phrased in rate coordinates;
Appendix~\ref{pf:if_memory} shows that the obstruction remains for any decoding of the spikes.  It explains
why the IF conversion pipelines we compare against reset globally every frame, discarding exactly
the state that makes incremental computation cheap.  The LIF constant step size trades the exact limit for a permanent
$\frac{1}{\tau}$ residual  and buys, in exchange, a tracking ability that
does not decay with time: the property the streaming framework of
Algorithm~\ref{alg:stream_processing} is built on.
\end{remark}

\subsubsection{Charge debt: why no decoder lets an IF neuron track a stream}
\label{pf:if_memory}

\paragraph{What we claim.} Tracking a stream means that the output stays close to the ANN
output of the recent input, with a lag that does not grow with how long the stream has run.
LIF-SNNs have this property for any input (Corollary~\ref{cor:tracking}).  IF-SNNs do not, whatever
decoder reads their spikes: there are inputs for which every decoder fails for an arbitrarily long
time.  The results above describe the IF neuron through its rate code; this section shows that
the obstruction does not depend on that choice.

\paragraph{Intuition: charge debt.} By~\eqref{eq:if_telescope} the IF membrane is an unweighted
sum of everything the neuron has received and paid out,
$u(t) = u(0) + \sum_{k\le t}\big(I(k) - \theta s(k)\big)$: an input from the distant past counts as
much as the newest one.  While the input stays in $[0,\theta]$ this does no harm, since every input
soon leaves the membrane as spikes.  Negative input cannot leave, because a neuron cannot emit
negative spikes.  It piles up as a \emph{debt}, and the neuron stays silent until new input repays
it, however long that takes; during that time no decoder reading its spikes, and no downstream
layer, can recover the correct value.  This is the common case in a converted network, since
negative input is exactly when the ReLU1 unit outputs $0$: in our detector $45\%$ of the IF
membranes are below $0$ at a tick on average (Appendix~\ref{app:indepth:porting}).  An LIF neuron instead
discounts an input received $t-k$ ticks ago by $\alpha^{t-k}$, the rate at which its target forgets
the stream through $\tilde{x}(t)$, so its memory, and its debt, fades.

\paragraph{Proof by counterexample, in four steps.}
(1)~For input $I \le \theta$ the IF spike count is exactly the running maximum of the cumulative
input (Proposition~\ref{prop:if_running_max}).
(2)~Two IF neurons, A with a long negative prefix followed by a positive input and B with a
negative input throughout, therefore emit the same (empty) spike train for a time that grows with
the prefix length $L$.
(3)~Their correct outputs differ during that time, but a decoder sees the same spikes for both, so
it is wrong on one of them (Corollary~\ref{cor:if_windup}).
(4)~Since $L$ is arbitrary, no decoder gives IF a lag bound that holds over a stream, whereas an
LIF neuron recovers within a time set by $\tau$ alone (Remark~\ref{rem:lif_recovery}).

\paragraph{Step 1: the spike count.} Consider one IF neuron~\eqref{eq:if_neuron} with scalar
input current $I(t)$, initial potential $u(0)\in[0,\theta)$, $\mathbb{H}(0)=1$, and write
\begin{equation}
S(t) \coloneqq \sum_{i=1}^{t} I(i), \qquad N(t) \coloneqq \sum_{i=1}^{t} s(i), \qquad S(0)=N(0)=0,
\end{equation}
for the cumulative input and the spike count.

\begin{proposition}[The spike count is a running maximum]
\label{prop:if_running_max}
If $I(t) \le \theta$ for all $t \ge 1$, then for all $t \ge 0$
\begin{equation}
N(t) \;=\; \max_{0 \le k \le t} \Big\lfloor \frac{u(0) + S(k)}{\theta} \Big\rfloor
\qquad\text{and}\qquad u(t) < \theta .
\label{eq:if_running_max}
\end{equation}
\end{proposition}

\begin{proof}
Write $M(t)$ for the right-hand side of~\eqref{eq:if_running_max}; $M(0) = \lfloor u(0)/\theta\rfloor = 0 = N(0)$
and $u(0) < \theta$.  Assume $N(t-1) = M(t-1)$ and $u(t-1) < \theta$.  By~\eqref{eq:if_telescope},
$u_{\text{pre}}(t) = u(0) + S(t) - \theta N(t-1)$, so the neuron fires at $t$ if and only if
$\lfloor (u(0)+S(t))/\theta \rfloor \ge N(t-1) + 1$.
\emph{If it fires}, then also $u_{\text{pre}}(t) = u(t-1) + I(t) < 2\theta$, i.e.\
$\lfloor (u(0)+S(t))/\theta \rfloor \le N(t-1)+1$.  Hence
$\lfloor (u(0)+S(t))/\theta \rfloor = N(t-1)+1 = N(t)$, which exceeds $M(t-1)$, so $M(t) = N(t)$;
and $u(t) = u_{\text{pre}}(t) - \theta < \theta$.
\emph{If it does not fire}, then $\lfloor (u(0)+S(t))/\theta \rfloor \le N(t-1) = M(t-1)$, so
$M(t) = M(t-1) = N(t)$, and $u(t) = u_{\text{pre}}(t) < \theta$.
\end{proof}

\paragraph{Reading the proposition.} Without the constraint that spikes are nonnegative and at most one per tick,
the count would simply be $\lfloor (u(0)+S(t))/\theta\rfloor$, all charge received so far divided by
the charge per spike.  The constraint $I \le \theta$ means this ideal count rises by at most one per tick,
so the only constraint that ever binds is that the count cannot go down.  The actual count is
therefore the running maximum of the ideal one: it rises only when the cumulative input exceeds
its earlier peak.  A worked example with $\theta = 1$, $u(0) = 0$:
\begin{center}
\small
\begin{tabular}{@{}l*{12}{c}@{}}
\toprule
$t$ & 1 & 2 & 3 & 4 & 5 & 6 & 7 & 8 & 9 & 10 & 11 & 12 \\
$I(t)$ & .6 & .6 & .6 & $-1$ & $-1$ & $-1$ & .6 & .6 & .6 & .6 & .6 & .6 \\
$\lfloor S(t)\rfloor$ & 0 & 1 & 1 & 0 & $-1$ & $-2$ & $-1$ & 0 & 0 & 1 & 1 & 2 \\
$N(t)$ & 0 & \textbf{1} & 1 & 1 & 1 & 1 & 1 & 1 & 1 & 1 & 1 & \textbf{2} \\
$u(t)$ & .6 & .2 & .8 & $-.2$ & $-1.2$ & $-2.2$ & $-1.6$ & $-1.0$ & $-.4$ & .2 & .8 & .4 \\
\bottomrule
\end{tabular}
\end{center}
After the input returns to $0.6$ at $t = 7$, the neuron stays silent for five ticks while it repays
its debt; a neuron that had seen only the positive input would have fired about three times in the
same interval.

\paragraph{Steps 2 and 3: the same spikes, different correct outputs.}

\begin{corollary}[No decoder gives IF a lag bound]
\label{cor:if_windup}
Let $u(0)=0$, $0 < a \le 1$, $c > 0$, $L \in \mathbb{N}$, and set $m \coloneqq \lfloor cL/(a\theta) \rfloor$.
Consider two IF neurons with input currents
\begin{equation}
I_A(t) = \begin{cases} -c, & t \le L,\\ a\theta, & t > L, \end{cases}
\qquad\qquad
I_B(t) = -c \quad\text{for all } t .
\end{equation}
Both neurons emit no spike for all $t \le L + m$, so their spike trains coincide up to $L+m$.
Yet on $(L, L+m]$ their correct outputs differ,
$\text{\normalfont ReLU1}(I_A(t)/\theta) = a$ while $\text{\normalfont ReLU1}(I_B(t)/\theta) = 0$.
Hence any decoder of the spike train, causal or not, of any memory length, has error at least
$a/2$ on one of the two neurons at every $t \in (L, L+m]$.  Since $m$ grows linearly in $L$ and $L$
is arbitrary, no decoder gives an IF neuron a lag bound that holds over a stream.
\end{corollary}

\begin{proof}
Both inputs satisfy $I \le \theta$, so Proposition~\ref{prop:if_running_max} applies.  For neuron $B$,
$S_B(k) = -ck < 0$ for all $k \ge 1$, so $N_B(t) = 0$.  For neuron $A$ and $k \le L + m$,
$S_A(k) \le -cL + (k-L)^{+} a\theta \le -cL + m a\theta \le 0$, so every term of the running
maximum is at most $0$ and $N_A(k) = 0$.  The spike trains are identical (all zero) up to $L+m$, so
any decoder returns the same value $\hat{y}$ for both; since the correct outputs differ by $a$,
$\max\{|\hat{y} - a|, |\hat{y} - 0|\} \ge a/2$.
\end{proof}

Neuron $B$ is what makes the statement hold for every decoder: a decoder that reads silence as $a$
would be right on $A$, but then it is wrong on $B$, which emits the same silence.  A neuron with
input $I = a\theta$ from the start, by contrast, has $N(t) = \lfloor at \rfloor$ and fires at rate $a$
immediately; neuron $A$ falls behind it only because of the debt it accumulated before $L$.  The
same construction with $I > \theta$ during the prefix gives the symmetric case, where excess charge
keeps the neuron firing after the drive has dropped.

\paragraph{Why one neuron suffices.} The statement is negative, so a single neuron is enough: a
neuron of the first hidden layer receives currents of exactly this form for suitable inputs.
Whether the mechanism is what limits a full IF network on a real stream is an empirical question,
answered in Appendix~\ref{app:indepth:porting}.

\paragraph{Step 4: the LIF neuron recovers.}

\begin{remark}[LIF recovers in a time independent of the history]
\label{rem:lif_recovery}
For the LIF neuron~\eqref{eq:lif_charge} with $u_{\text{rest}} = 0$, a current bounded below by
$I(t) \ge -c$ keeps the potential bounded below, $u(t) \ge \min\{u(0), -Rc\}$, because between
spikes $u$ is a convex combination of its previous value and $R\,I(t)$, and a reset only follows
$u_{\text{pre}} \ge \theta > 0$.  If the current then switches to a constant $I_{\text{in}}$ with
$R I_{\text{in}} > \theta$, then until the first spike $u(L+n) = \alpha^{n} u(L) + (1-\alpha^{n}) R I_{\text{in}}$,
and $\alpha^{n} \le e^{-n/\tau}$ shows that the first spike occurs within
\begin{equation}
n \;\le\; \Big\lceil \tau \ln \frac{R I_{\text{in}} + \max\{Rc, -u(0)\}}{R I_{\text{in}} - \theta} \Big\rceil
\end{equation}
ticks, a bound that does not depend on $L$.  This is the fading memory that the streaming
framework relies on: leak removes the debt that an IF neuron would carry forever.
\end{remark}

\paragraph{The obvious fix does not work.}

\begin{remark}[Bounding the IF membrane breaks the conversion]
\label{rem:if_bound}
The debt could be limited by clamping the IF membrane from below, $u(t) \leftarrow \max\{u(t), \beta\}$,
as done by \citet{kugele2020efficient}.  The clamp, however, invalidates the steady-state
correspondence of Theorem~\ref{thm:equilibrium_if_app} on which conversion rests.  Let $D(t) \ge 0$ be the
total charge discarded by the clamp up to time $t$.  The charge balance~\eqref{eq:if_telescope}
becomes $u(t) = u(0) + S(t) - \theta N(t) + D(t)$.  Whenever $u(t)$ stays bounded, as in the example below, this gives
\begin{equation}
\frac{N(t)}{t} \;=\; \frac{1}{\theta}\Big(\frac{S(t)}{t} + \frac{D(t)}{t}\Big) + \mathcal{O}\Big(\frac{1}{t}\Big).
\end{equation}
The rate is the mean input plus the mean discarded charge, and the latter depends on how the input
fluctuates rather than on its mean.  The rate is therefore no longer a function of the mean input,
and in a network the currents $W s(t-1)$ fluctuate from tick to tick even for a constant
stimulus.  For example, with $\beta = 0$, $u(0) = 0$ and input alternating between $-\theta$ and
$1.5\theta$, the charge below $0$ is discarded on every negative tick and the neuron fires on every positive tick:
the rate is $0.5$, whereas $\text{ReLU1}$ of the mean input is $0.25$, which an unclamped IF neuron
attains.  An upper clamp fails by the symmetric argument, with $D(t) \le 0$.  Leak, in contrast, bounds the history while
preserving the correspondence (Corollary~\ref{cor:practical_stability}).
\end{remark}

\newpage
\section{Algorithms}

\begin{figure}[H]
\vspace{-1.5em}
\centering
\begin{minipage}[t]{0.6\textwidth}
\begin{algorithm}[H] 
\caption{\modelname{} }
\label{alg:stream_processing}
\begin{algorithmic}[1]
\footnotesize %
\REQUIRE LIF-SNN $\mathcal{S}$; Stream $\{x_i\}$; Ticks $\{T_i\}$; Leak $\tau$; Init $y(0), z(0), u_n(0)$
\ENSURE Output stream $\{y_i\}$
\STATE Initialize LIF neurons $n \in \mathcal{S}$ with $u_n(0)$
\STATE $y \leftarrow y(0)$, $z \leftarrow z(0)$
\FOR{each frame $x_i$ in stream}
    \FOR{$t = 1$ \TO $T_i$}
        \STATE $x(t), z \leftarrow \text{StreamEncode}(x_i, z, \tau)$ \COMMENT{Algorithms~\ref{alg:float_encoding}--\ref{alg:spike_population_encoding}}
        \STATE $y \leftarrow \frac{\tau-1}{\tau} y + \frac{1}{\tau} \mathcal{S}(x(t))$ \COMMENT{decode}
    \ENDFOR
    \STATE $y_i \leftarrow y$
\ENDFOR
\end{algorithmic}
\end{algorithm}
\end{minipage}
\end{figure}

\paragraph{Stream encoders.} The SNN sees its input only through the exponentially coded input
$\tilde{x}(t)$ (Theorem~\ref{thm:equilibrium_lif}), so any encoder is valid as long as
$\tilde{x}$ tracks the frame $x$.  With $z$ the encoder's replica of $\tilde{x}$,
\begin{equation}
x - z \;\leftarrow\; \tfrac{\tau-1}{\tau}\,(x - z) \;+\; \tfrac{1}{\tau}\,\big(x - x(t)\big),
\label{eq:encoder_error}
\end{equation}
i.e., each encoder is a feedback controller on this error, and must emit $x$ on average once
$z = x$.  The float encoders (Algorithms~\ref{alg:float_encoding}--\ref{alg:float_lookahead_encoding})
are $x(t) = x + \kappa(x - z)$, contracting the error by $\frac{\tau-1-\kappa}{\tau}$ per tick.  The
spike encoders (Algorithms~\ref{alg:spike_encoding}--\ref{alg:spike_population_encoding}) quantize the
same controller: sigma-delta is the LIF dynamics~\eqref{eq:fixedpoint_equation_y} with drive $x$,
and population lookahead splits the lookahead current over $K \ge \tau$ neurons with graded
thresholds, each firing at most one binary spike per tick, keeping $\vert x - z\vert \le
\frac{1}{2\tau}$ for $x \in [0,1]^d$.

\begin{figure}[H]
\vspace{-1em}
\centering
\begin{minipage}[t]{0.48\textwidth}
\begin{algorithm}[H]
\caption{Sample-and-hold (float)}
\label{alg:float_encoding}
\begin{algorithmic}[1]
\footnotesize
\REQUIRE Frame $x$; State $z$; Leak $\tau$  \ENSURE Current $x(t)$; State $z$
\STATE $x(t) \leftarrow x$ \COMMENT{Stateless encoding}
\STATE $z \leftarrow \frac{\tau-1}{\tau} z + \frac{1}{\tau} x(t)$ \COMMENT{$x - z$ decays by $\frac{\tau-1}{\tau}$}
\RETURN $x(t),~z$
\end{algorithmic}
\end{algorithm}
\vspace{-1.2em}
\begin{algorithm}[H]
\caption{Proportional residual (float)}
\label{alg:float_residual_encoding}
\begin{algorithmic}[1]
\footnotesize
\REQUIRE Frame $x$; State $z$; Leak $\tau$; Gain $\kappa \in [0, \tau-1]$  \ENSURE Current $x(t)$; State $z$
\STATE $x(t) \leftarrow x + \kappa\,(x - z)$ \COMMENT{Residual feedback}
\STATE $z \leftarrow \frac{\tau-1}{\tau} z + \frac{1}{\tau} x(t)$ \COMMENT{$x - z$ decays by $\frac{\tau-1-\kappa}{\tau}$}
\RETURN $x(t),~z$
\end{algorithmic}
\end{algorithm}
\vspace{-1.2em}
\begin{algorithm}[H]
\caption{Lookahead (float)}
\label{alg:float_lookahead_encoding}
\begin{algorithmic}[1]
\footnotesize
\REQUIRE Frame $x$; State $z$; Leak $\tau$  \ENSURE Current $x(t)$; State $z$
\STATE $z \leftarrow \frac{\tau-1}{\tau} z$ \COMMENT{Lookahead}
\STATE $x(t) \leftarrow \tau\,(x - z)$ \COMMENT{Scaled residual}
\STATE $z \leftarrow z + \frac{1}{\tau} x(t)$ \COMMENT{$z = x$ at every tick}
\RETURN $x(t),~z$
\end{algorithmic}
\end{algorithm}
\end{minipage}
\hfill
\begin{minipage}[t]{0.48\textwidth}
\begin{algorithm}[H]
\caption{Sigma-delta (spike)}
\label{alg:spike_encoding}
\begin{algorithmic}[1]
\footnotesize
\REQUIRE Frame $x$; State $z$; Leak $\tau$  \ENSURE Spikes $x(t)$; State $z$
\STATE $x(t) \leftarrow \mathbb{H}(x - z)$ \COMMENT{Fire if under-coded}
\STATE $z \leftarrow \frac{\tau-1}{\tau} z + \frac{1}{\tau} x(t)$ \COMMENT{$\vert x - z\vert \lesssim \frac{1}{\tau}$}
\RETURN $x(t),~z$
\end{algorithmic}
\end{algorithm}
\vspace{-1.2em}
\begin{algorithm}[H]
\caption{Stochastic sigma-delta (spike)}
\label{alg:spike_stochastic_encoding}
\begin{algorithmic}[1]
\footnotesize
\REQUIRE Frame $x$; State $z$; Leak $\tau$; Dither $\sigma$  \ENSURE Spikes $x(t)$; State $z$
\STATE $\xi \sim \mathcal{N}(\mathbf{0}, \sigma^2\mathbb{I})$
\STATE $x(t) \leftarrow \mathbb{H}(x - z + \xi)$ \COMMENT{Dithered firing}
\STATE $z \leftarrow \frac{\tau-1}{\tau} z + \frac{1}{\tau} x(t)$
\RETURN $x(t),~z$
\end{algorithmic}
\end{algorithm}
\vspace{-1.2em}
\begin{algorithm}[H]
\caption{Population lookahead (spike)}
\label{alg:spike_population_encoding}
\begin{algorithmic}[1]
\footnotesize
\REQUIRE Frame $x$; State $z$; Leak $\tau$; Population $K$  \ENSURE Spikes $\{s_j^{\pm}\}_{j=1}^{K}$; State $z$
\STATE $z \leftarrow \frac{\tau-1}{\tau} z$ \COMMENT{Lookahead}
\STATE $r \leftarrow \tau\,(x - z)$ \COMMENT{Float lookahead current}
\FOR{$j = 1$ \TO $K$}
    \STATE $s_j^{\pm} \leftarrow \mathbb{H}(\pm r - j + \frac{1}{2})$ \COMMENT{Thermometer code}
\ENDFOR
\STATE $x(t) \leftarrow \sum_{j=1}^{K} (s_j^{+} - s_j^{-})$ \COMMENT{$= \text{round}(r)$}
\STATE $z \leftarrow z + \frac{1}{\tau} x(t)$ \COMMENT{$\vert x - z\vert \le \frac{1}{2\tau}$}
\RETURN $\{s_j^{\pm}\},~z$
\end{algorithmic}
\end{algorithm}
\end{minipage}
\vspace{-1em}
\end{figure}

\begin{algorithm}[H]
\caption{ANN $\to$ LIF-SNN conversion}
\label{alg:lif_conversion}
\begin{algorithmic}[1]
\REQUIRE ANN $g$ with ReLU1 activations; leak $\tau > 1$; threshold $\theta = 1$
\ENSURE LIF-SNN $\mathcal{S}$, codec initializer $y(0)$, initial $u_n(0)$ for all neurons $n \in \mathcal{S}$
\STATE $a_\tau(x) := (\tau - 1)x + 1$
\STATE \textit{// Main conversion}
\STATE $\mathcal{S} \leftarrow \textsc{Copy}(g)$
\FORALL{DEQ layers $l \in L \subseteq \mathcal{S}$}
    \STATE $f \leftarrow \text{forward function of } l$
    \FORALL{ReLU1 activations $\sigma \in f$}
    \STATE $n \leftarrow \textsc{LIF}(\tau,\theta)$
    \STATE $\sigma \leftarrow n \circ a_\tau$
    \ENDFOR
    \STATE l $\leftarrow$ \textsc{Recurrent-SNN}$(f)$
\ENDFOR
\FORALL{ReLU1 activations $\sigma \in \mathcal{S}-L$}
    \STATE $n \leftarrow \textsc{LIF}(\tau,\theta)$
    \STATE $\sigma \leftarrow n \circ a_\tau$
\ENDFOR
\STATE \textit{// Bias correction: pre-charge $u(0)$ with it}
\STATE Disable firing of every neuron $n \in \mathcal{S}$
\STATE Temporarily hook each neuron $n$ to store its influx current $b_n$
\STATE $y(0) \leftarrow \mathcal{S}(\mathbf{0})$
\FORALL{neurons $n \in \mathcal{S}$}
    \STATE $u_n(0) \leftarrow b_n$ 
\ENDFOR
\STATE Re-enable firing of neurons
\RETURN $\mathcal{S}$
\end{algorithmic}
\end{algorithm}
\subsection{Comparison of Stream Encoders}
\label{app:encoder_comparison}

\begin{figure}[H]
\centering
\includegraphics[width=\linewidth]{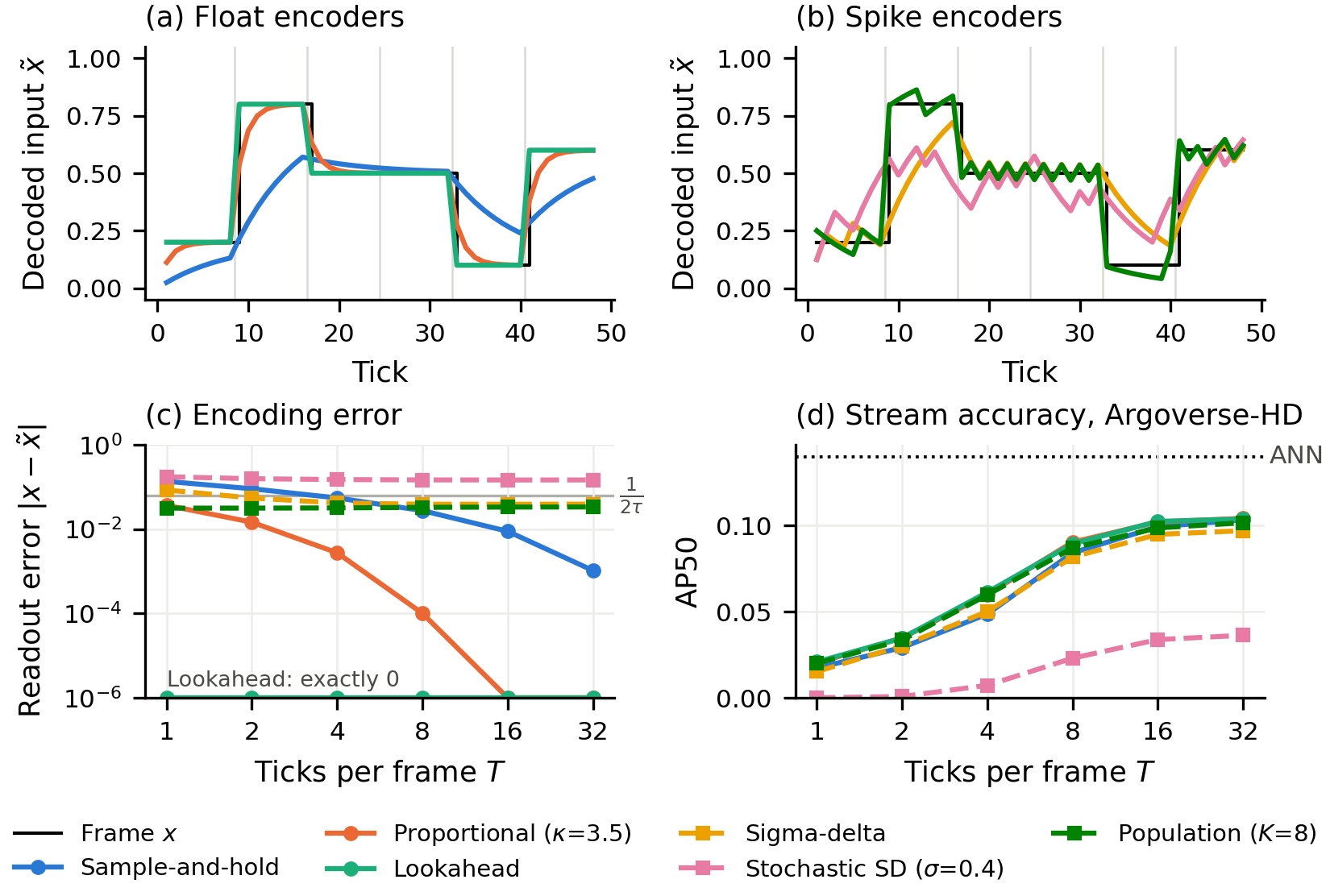}
\caption{Stream encoders ($\tau = 8$).  (a,b) The decoded input $\tilde{x}$ of one coordinate on
a staircase of six frames with $T = 8$ ticks each (grey lines mark frame boundaries).  (c) Input
readout error $\vert x - \tilde{x}\vert$ at the last tick of each frame against the tick budget,
on a correlated random stream; values are clipped at $10^{-6}$.  (d) AP50 of the converted
detector streaming one Argoverse-HD sequence ($901$ frames) against the tick budget; the dotted
line is the dense ANN.}
\label{fig:encoder_comparison}
\end{figure}

We compare the encoders of Algorithms~\ref{alg:float_encoding}--\ref{alg:spike_population_encoding}
at two levels, with $\tau = 8$, $\kappa = 3.5$, $\sigma = 0.4$, and $K = 8$
(Figure~\ref{fig:encoder_comparison}).
\emph{Input side} (a--c): by~\eqref{eq:encoder_error} the SNN sees the input only through
$\tilde{x}$, so encoding quality is $\vert x - \tilde{x}\vert$ at the tick a frame is read out,
measured on $d = 512$ coordinates over $64$ frames of a random walk in $[0,1]$ (per-frame step
$\mathcal{N}(0, 0.1^2)$, reflected at the boundaries).  \emph{Stream accuracy} (d): the converted
YOLOv8-style detector of Appendix~\ref{app:obj_det_eval} streams the first Argoverse-HD validation
sequence ($901$ frames) without reset, and we report AP50 over all frames.  Every encoder runs on
raw pixels in $[0,1]$ and feeds the network the normalized current $(x(t) - \mu)/\sigma_{\text{img}}$,
which for spike encoders amounts to binary spikes on pixel channels through the normalization
folded into the first layer; sample-and-hold is then identical to the streaming arm of the main
experiments.

\begin{itemize}
\item \textbf{Float encoders} follow the contraction $\big(\frac{\tau-1-\kappa}{\tau}\big)^T$ of
the gain family, reaching an input error of $2.7\times10^{-2}$ (sample-and-hold),
$1\times10^{-4}$ (proportional), and exactly $0$ (lookahead) at $T = 8$.  On the stream, the
faster input translates into faster tracking at small budgets: at $T = 4$, lookahead and
proportional reach AP50 $0.061$ and $0.060$ against $0.049$ for sample-and-hold, and all three
converge to $\approx 0.104$ by $T = 32$.
\item \textbf{Spike encoders} settle at an input error floor set by quantization rather than by
$T$: $0.039$ for sigma-delta and $0.033$ for population lookahead, which stays below
$\frac{1}{2\tau}$ at every tick at the cost of extra spikes only on frame changes
($0.73$ spikes per coordinate per tick at $T = 1$, $0.52$ at $T = 32$, against $0.51$ for
sigma-delta).  On the stream, population lookahead tracks its float counterpart within
$0.004$ AP50 at every budget, while sigma-delta follows sample-and-hold at small $T$ and levels
off lower ($0.097$ at $T = 32$), the residue of its quantization floor.
\item \textbf{Stochastic dithering} is the one encoder the detector does not tolerate: its
input noise leaves AP50 at $0.036$ even at $T = 32$.
\end{itemize}

This sequence is harder than average: the dense ANN reaches AP50 $0.140$ on it against $0.296$ on
the full validation set, so the numbers above are meaningful relative to one another and to that
reference rather than in absolute terms.

\section{Baseline Methods and Implementation Details}
\label{app:baselines}

This appendix describes how \modelname{} is implemented (Appendix~\ref{app:baselines:ours}), how the
baselines are implemented (Appendix~\ref{app:baselines:conv}--Appendix~\ref{app:baselines:direct}), and the
protocol that is shared by every row of Table~\ref{tab:snn_comparison_revised}
(Appendix~\ref{app:baselines:protocol}).  All methods are implemented in a single PyTorch codebase on
top of \emph{spikingjelly}~\citep{doi:10.1126/sciadv.adi1480}, so that every arm of a comparison shares the same data
pipeline, the same decoder, and the same metric code.  Each baseline is implemented directly
from the equations of its own paper; where a published method offers several settings we use
the one its authors report as best.

\subsection{\modelname{}}
\label{app:baselines:ours}

\paragraph{Convertible ANN.} Our framework requires only that the reference ANN be a
composition of affine operators and the bounded activation
$\text{ReLU1}(x)$. Concretely, every architecture we use is built from
convolutions, linear layers, batch normalization, average pooling, and $\text{ReLU1}$.
Operators that are not affine (e.g.\ max-pooling, layer normalization, softmax attention) are
replaced by affine equivalents before training, so no approximation is introduced at
conversion time.  
Task-specific post-processing that contains no learnable parameters --
non-maximum suppression for detection, CTC collapse for speech, arg-max for classification --
is executed outside the converted graph, exactly as in the ANN.

\paragraph{Conversion.} Conversion follows Algorithm~\ref{alg:lif_conversion}. The model is
traced into a graph, each Conv--BatchNorm pair is folded into a single affine operator, and
each $\text{ReLU1}$ site is replaced by the composition $n \circ a_\tau$ of an input rescaling
$a_\tau(x) = (\tau-1)x + 1$ and a soft-reset LIF neuron with $\theta=1$,
$u_{\text{rest}}=0$, $R=1$.  Because $\text{ReLU1}$ is bounded by construction, the firing
threshold is fixed at $1$ and \emph{no activation statistics, calibration set, or per-layer
search is required}; this is the only structural difference from the conversion baselines in
Appendix~\ref{app:baselines:conv}, all of which fit at least one parameter per layer from data.  The
single data-independent step is the membrane pre-charge of
Algorithm~\ref{alg:lif_conversion}: firing is disabled, the zero input is propagated once, and
each neuron's resting influx is stored as its initial membrane potential $u_n(0)$ together with
the output codec state $y(0)$.  This is a one-off correction of the constant term $b(t)$ in
Equation~(\ref{eq:relation_of_upre_y}); it is computed once at conversion time and never re-applied.

\paragraph{Streaming inference.} Streaming inference follows
Algorithm~\ref{alg:stream_processing}.  The membrane potentials of all neurons and the
input/output codec states are initialized once at the beginning of a stream and then persist
across frames; there is no global reset, no per-frame re-initialization, and no cross-frame
communication other than the neuronal state itself.  Each frame $k$ is allotted $T$ ticks, and
the prediction for frame $k$ is read from the output codec at the last tick of that frame.
Inputs are encoded with the sample-and-hold scheme of
Algorithm~\ref{alg:float_encoding}: at each tick the encoder emits the current frame
unchanged, and the change between frames reaches the SNN through the exponentially coded input
$\tilde{x}(t)$, which follows the new frame at the leak rate $\frac{\tau-1}{\tau}$.

\paragraph{Hyperparameters.} The framework has two hyperparameters: the leak $\tau$ and the
per-frame tick budget $T$.  We use $\tau = 8$ throughout (except the Best column), and sweep $T \in \{2,4,8,16,32,64\}$,
which is the axis reported in Table~\ref{tab:snn_comparison_revised}. For the \textbf{Best} column, we use $\tau=16$ and measure the accuracy in $T>128$, just like all the other baseline methods. An optional
\emph{warm-up} grants the first frame of a stream $4T$ extra ticks so that the very first
prediction is not read from a cold membrane; its cost is amortized over the whole stream and it
is disabled unless stated. The only exception is the Argoverse-HD entry of \modelname{} at $T=16$ in Table~\ref{tab:snn_comparison_revised}, which uses a warm-up of $64$ ($=4T$) ticks.  No other tuning is performed, and the same $\tau$ is used for all
three tasks.

\subsection{ANN-to-SNN conversion baselines}
\label{app:baselines:conv}

These methods share our setting most closely: they start from a trained ANN and produce an SNN
without direct SNN training.  

\begin{itemize}
\item \textbf{ReLU-Norm}~\citep{rueckauer2017conversion} (with the max-norm variant of
  \citet{diehl2015thresholdbalancing}): converts an \emph{unbounded}-ReLU ANN by normalizing each layer by a scale
  $\lambda_l$ estimated on a calibration set, so that weights and biases become
  $\tilde{W}_l = W_l \lambda_{l-1}/\lambda_l$ and $\tilde{b}_l = b_l/\lambda_l$.  We implement
  both the max-norm ($\lambda_l = \max a_l$) and the robust percentile variant, normalizing one
  layer at a time and re-measuring after each so that the rescaling is correct when a
  normalization layer sits between the affine layer and the activation.  We report the better
  of the two.
\item \textbf{RTS}~\citep{deng2021optimal}: threshold balancing plus a single closed-form bias
  shift $b_l \leftarrow b_l + \theta/(2T)$ that minimizes the half-step mismatch between
  the ANN's clipped activation and the SNN's step response.  Its source activation is the
  threshold-ReLU, which is exactly $\text{ReLU1}$ at threshold $1$, so no retraining is
  required.  Note that $T$ enters the stored parameters, so a converted model is tied to the
  tick budget it was converted for.
\item \textbf{RMP-SNN}~\citep{hao2023reducingresmembranepotential}: soft reset
  ($V \leftarrow V - V_{\text{th}}$) instead of hard reset, together with a layer-by-layer
  threshold balancing $V_{\text{th}}^l \leftarrow \alpha \max_l$ with $\alpha \le 0.6$, each
  layer frozen before the next is estimated.
\item \textbf{Spike-Calib}~\citep{li2021freelunch}: post-hoc calibration of the converted IF-SNN.
  We implement both the \emph{light} (bias) and \emph{advanced} (bias, initial potential and
  weights) variants, and follow the paper's greedy Algorithm~1: the $T$-step simulation is
  re-run before each layer is calibrated, so every layer is corrected against the signal it
  actually receives from the already-corrected layers upstream.  BatchNorm folding (their
  Eq.~19) is already performed by our shared conversion front-end.
\item \textbf{QCFS}~\citep{buqcfs}: a training-time method.  The source ANN is trained with the
  quantization-clip-floor-shift activation
  $a^l = \lambda^l \,\text{clip}\big(\tfrac{1}{L}\lfloor z^l L/\lambda^l + \phi \rfloor, 0,
  1\big)$ with a trainable per-layer $\lambda^l$ and the paper's recommended shift
  $\phi = 1/2$, and is then converted by $\theta^l \leftarrow \lambda^l$,
  $v^l(0) \leftarrow \theta^l \phi$.  We train a separate ANN per task with $L=8$.
\item \textbf{SlipReLU}~\citep{jiang2023sliprelu}: also training-time.  The activation is the
  convex combination
  $c\,\theta\,\text{clip}(z/\theta + \delta_1, 0, 1) + (1-c)\,\theta\,\text{clip}(\tfrac{1}{N}
  \lfloor Nz/\theta + \delta\rfloor, 0, 1)$, converted by $V_{\text{th}} \leftarrow \theta$,
  $v(0) \leftarrow V_{\text{th}} \delta$.  We use the authors' best reported settings
  $\delta_1 = 0$, $\delta = 1/2$, $c = 0.4$, and train a separate ANN per task.
\item \textbf{DCGS}~\citep{huang2025differential}: the multi-threshold (MT) neuron, carrying
  $2n$ thresholds $\lambda_p = \theta/2^{p-1}$ for $p = 1,\dots,n$ together with their
  negatives, and emitting at most one graded spike per time step whose magnitude is the
  $\lambda_p$ closest to the membrane potential; $n=1$ degenerates to an IF neuron with an added
  negative threshold.  We use $n = 4$, and report the reported row under rate, since differential coding failed to achieve meaningful accuracy on all tasks.  Two
  threshold settings are run and we report the better: the per-layer threshold iteration of
  their Algorithm~1, with the fitted thresholds passed per site rather than collapsed to a
  single global value, and the fixed $\theta = 1$ that a bounded activation admits by
  construction.
\item \textbf{AdaFire} and \textbf{AdaFire+SSC}~\citep{wang2025adaptive}: an adaptive-firing
  neuron emitting up to $\phi$ spikes per step, so its per-step output range is
  $[0, V_{\text{th}}\phi]$, with $\phi$ chosen per layer on the paper's
  sensitivity/energy criterion.  \textbf{+SSC} adds Sensitivity Spike Compression, which raises
  the per-layer threshold by a factor $\rho$ so that consecutive spikes merge into fewer,
  higher-amplitude ones.  We do not implement the third component (Input-aware Adaptive
  Timesteps), an entropy-based early exit over a \emph{bounded} sample, because this evaluation
  scores a perpetual stream at a fixed per-frame budget.
\item \textbf{LIF-Reset}: our own converted LIF-SNN, run under the conventional protocol --
  membranes and codec states are reset at every frame boundary and the frame is simulated for
  $T$ ticks.  This is the direct ablation of streaming: identical weights, identical neurons,
  identical decoder, the reset being the only difference.
\end{itemize}

\subsection{Directly-trained SNN baselines}
\label{app:baselines:direct}

These methods train a spiking network end-to-end with surrogate gradients. For methods that propose specific spiking neural architectures, we use the authors' released architectures  and their released
weights except DVS-Gesture task. For the DVS-Gesture task, we train all the models with our task setup.

\begin{itemize}
\item \textbf{Training rules.} Rectangular/boxcar surrogate gradients (STBP,
  \citealp{wu2018spatio}); the online rules OTTT~\citep{Xiao2022OnlineTT} and
  SLTT~\citep{meng2023sltt}, whose memory is constant in $T$; the TET loss~\citep{deng2022temporal},
  which scores every tick rather than the time average; and ASGL~\citep{wang2023asgl}, whose
  analog branch is itself a bounded ReLU.
\item \textbf{Spike-driven Transformer (SDT)}~\citep{yao2023sdt} (gesture): spike-driven
  self-attention (SDSA), which replaces softmax attention with a Hadamard product and
  column-wise sums of binary spike matrices, with membrane shortcuts between blocks.  We use a
  depth-4, width-256 skeleton of $4.15$M parameters, comparable to GestureNet ($3.96$M), train it
  with the same stream windows, augmentation and EMA schedule as our reference ANN, and report
  the EMA checkpoint.  Its forward pass steps through the $16$ frames of a window, so its time
  steps are tied to its frame binning and we report it at this design point, $T=16$.
  Re-binning a window into $T \ne 16$ frames, with the input rescaled to the trained per-frame
  intensity, moves the model off its training distribution and lowers accuracy on both sides
  ($21.9$--$82.6$ for $T \in \{2, \dots, 64\}$).  Spikformer~\citep{zhou2022spikformer} and
  Meta-SpikeFormer~\citep{yao2024sdtv2}, trained identically on the same skeleton, reached
  $89.4$ and $88.8$; we list the Spike-driven Transformer, which ties Spikformer overall and is
  slightly more accurate on the gesture classes (11 classes) ($91.3$ vs.\ $91.2$).
\item \textbf{EMS-YOLO}~\citep{su2023deep} (detection): an ordinary binary LIF neuron
  ($V[t{+}1] = \tau V[t](1 - X[t]) + \sum_j W_{ij}X_j[t{+}1]$, $X[t{+}1] = H(V[t{+}1] -
  V_{\text{th}})$) with a rectangular surrogate, in the EMS-ResNet block that places a LIF
  before every convolution -- including on the shortcut -- so that every convolution receives a
  binary tensor.  We evaluate the authors' COCO weights.
\item \textbf{SpikeYOLO}~\citep{luo2024integer} (detection): the I-LIF neuron
  $\text{clip}(\text{round}(U), 0, D)$ with the meta SNN blocks, trained with integer values and
  deployed with each integer step expanded into $D$ binary sub-steps.  We evaluate the authors'
  COCO weights in both the integer and the expanded-binary inference modes.
\end{itemize}

\subsection{Common evaluation protocol}
\label{app:baselines:protocol}

\paragraph{Shared reference ANN.} Every post-hoc conversion baseline whose precondition our
ANN satisfies is converted from the \emph{same} trained checkpoint as \modelname{}, so that the
comparison prices what each method loses in conversion rather than differences in the source
model.  Methods whose activation function is part of their contribution (QCFS, SlipReLU) require their
own source ANN; we train one per task with the identical architecture, schedule, and
augmentation, changing only the activation.  Methods that need unbounded ReLU statistics
(ReLU-Norm, Spike-Calib, AdaFire) are given an unbounded-ReLU ANN trained the same way,
plus the calibration split they require.

\paragraph{Reset convention.} Every baseline is run under the protocol its paper specifies,
which for all conversion baselines means a per-frame reset: the network state is cleared, the
frame is simulated for $T$ ticks, and the prediction is decoded.  \modelname{} is run without
any reset after the start of the stream.  The \emph{LIF-Reset} row isolates this single
difference on our own converted network.  In Figure~\ref{fig:ablation_study} and Appendix~\ref{app:indepth:porting}, we additionally report
what happens when the reset is simply removed from the published conversions.

\paragraph{Compute.} All models we train ourselves are trained and evaluated on a single
NVIDIA RTX~2080~Ti (12\,GB).

\section{Zero-shot Transfer for Streaming Object Detection}
\label{app:obj_det_eval}

\subsection{Datasets}

\paragraph{COCO 2017 (training).} The reference ANN is trained for 80-class object detection on
COCO 2017~\citep{lin2014coco}: $118{,}287$ \texttt{train2017} images and $5{,}000$
\texttt{val2017} images.  We use only the box annotations.  COCO is the source domain and is
used for training and for reporting the ANN's frame-wise quality; no streaming evaluation is
performed on it.

\paragraph{Argoverse-HD (streaming evaluation).} Streaming detection is evaluated on
Argoverse-HD~\citep{li2020streaming}, which pairs the front-center ring camera of
Argoverse~1.1~\citep{chang2019argoverse} with dense 2D box annotations at 30\,fps.  We use the
full validation split: $24$ sequences and $15{,}062$ annotated frames, $\sim\!627$ frames per
sequence on average, at a native resolution of $1920\times1200$.  Argoverse-HD annotates eight
categories -- \emph{car, truck, bus, person, bicycle, motorcycle, traffic light, stop sign} --
all of which are COCO categories with the same definitions, so a COCO-trained detector is
scored directly, with no fine-tuning and no modification of the detection head.  The
annotations are dense and are \emph{not} temporally interpolated, so every frame is an
independent ground-truth observation.

\paragraph{Why this pair.} We chose a zero-shot transfer setting for two reasons.  First, COCO
is the benchmark on which SNN detectors are trained and compared, but it has no streaming
evaluation; Argoverse-HD provides a rigorous 30\,fps streaming benchmark but has no SNN
detection literature.  Second, the two share their label space, so transfer requires no head
surgery and no retraining, and the comparison therefore measures streaming inference rather
than adaptation.  The same argument is made by \citet{li2020streaming}, whose own baselines are
likewise off-the-shelf COCO models evaluated without training on Argoverse.  We report ordinary
per-frame COCO AP rather than the streaming-AP metric of \citet{li2020streaming}: sAP scores a
detector against the world at the moment its output is emitted and therefore requires a
wall-clock latency model, whereas the quantity under study here is the tick budget a frame
needs, which is a hardware-independent count.

\subsubsection{Licenses}

COCO images are made available under the Flickr Terms of Use and its annotations under the
Creative Commons Attribution 4.0 License.  Argoverse~1.1 is released by Argo~AI under the
Creative Commons Attribution-NonCommercial-ShareAlike 4.0 License, and the Argoverse-HD
annotations are released by \citet{li2020streaming} for non-commercial research use.  All
datasets are used within their licensed terms, for non-commercial academic research.

\subsection{Model}

The reference ANN is a convertible YOLOv8-style detector~\citep{jocher2023ultralytics}: a CSP backbone, a PAN--FPN
neck, and a DFL detection head with task-aligned assignment, at three pyramid levels with
strides $(8,16,32)$.  We use the \texttt{v8m}-scale configuration -- stage widths
$(48,96,192,384,576)$ and backbone depths $(2,4,4,2)$ -- giving $25.05$M parameters and $39$
sequential spiking layers along the deepest path.

Three substitutions make the model convertible, and each one is applied \emph{before} training,
so the trained ANN and the converted SNN describe exactly the same function:

\begin{itemize}
\item \textbf{SiLU $\rightarrow$ ReLU1.} Every activation in the backbone, neck and head is the
  bounded $\text{ReLU1}$, which is what lets the converted network use a fixed unit threshold
  with no data-driven calibration.
\item \textbf{MaxPool $\rightarrow$ AvgPool.} The SPPF block's max-pooling is the only operator
  in the architecture with no affine equivalent; we use average pooling in its place.  (An exact
  reconstruction of max-pooling from $\text{ReLU1}$ units is also available and gives the same
  function at a higher layer count; we do not use it in the reported model.)
\item \textbf{Post-processing kept outside the graph.} Sigmoid decoding, DFL integration,
  coordinate de-normalization and NMS are parameter-free and run on the decoded output, exactly
  as in the ANN, and therefore never enter the converted network.
\end{itemize}

Baselines that require a different activation (QCFS, SlipReLU) or an unbounded one (ReLU-Norm,
Spike-Calib, AdaFire) are trained as separate arms of the same architecture, schedule and
augmentation, changing only the activation; the directly-trained SNN detectors (EMS-YOLO,
SpikeYOLO) use their authors' published architectures and weights.

\subsection{Training Setup}

Training follows the standard YOLOv8 recipe, adapted only where the convertibility constraint
requires it.  The model is trained from scratch on COCO \texttt{train2017} for $300$ epochs at
an input resolution of $640$, with SGD (learning rate $0.005$, final learning-rate factor
$0.01$, Nesterov momentum $0.937$, weight decay $5\times10^{-4}$) and a cosine schedule, after a
warm-up of $10{,}992$ optimizer steps ($\approx 3$ epochs) over which the momentum is ramped
from $0.8$.  The effective batch size is $32$, realized as batch $16$ with two gradient
accumulation steps.  Augmentation is mosaic ($p=1.0$, disabled for the final $10$ epochs),
horizontal flip ($p=0.5$), HSV jitter $(0.015, 0.7, 0.4)$, and random affine applied on the
composed mosaic canvas with translation $0.1$ and scale $0.5$.  The loss is the YOLOv8
composite -- CIoU box loss, distribution focal loss, and BCE classification loss -- with gains
$(7.5, 1.5, 0.5)$ and task-aligned assignment.  We use mixed precision, TF32 matmuls, and an EMA
of the weights; the EMA weights are the ones converted.  Checkpoints are selected on
\texttt{val2017} mAP.

\subsection{Evaluation Setup}

\paragraph{Input pipeline.} Each frame is letterboxed to $640\times640$, scaled to $[0,1]$ and
normalized with the ImageNet statistics used during training.  Frames
are presented in their recorded order at the native 30\,fps and no frame is skipped, since the
inter-frame interval is the quantity the streaming claim depends on.

\paragraph{Streaming protocol.} For \modelname{}, the converted SNN is instantiated once per
sequence, its membrane and codec states are initialized once, and the $24$ sequences are then
processed frame by frame with $T$ ticks per frame.  The prediction for a frame is read at the
last tick allotted to that frame.  For the reset baselines, the state is cleared at every frame
boundary and the frame is simulated for $T$ ticks, which is the protocol every published
conversion method specifies.  The dense ANN is run on the same frames .

\paragraph{Metric.} We report COCO AP@50 computed with \texttt{pycocotools}, with a score
threshold of $0.001$, NMS IoU $0.7$ and at most $300$ detections per frame, which is the
standard COCO convention.  The dense ANN and the streaming arms are scored over all $15{,}062$
annotated frames of the validation split; the per-frame-reset conversion baselines are scored on
the same $24$ sequences under the protocol their own papers specify.  Every arm is scored by the
same metric code, and each comparison reported in the paper is made over one common frame set.
Efficiency is reported as described in Appendix~\ref{app:baselines:protocol}.

\section{Streaming Automatic Speech Recognition (ASR)}
\label{app:asr_eval}

\subsection{Datasets}

We train and evaluate on LibriSpeech~\citep{panayotov2015librispeech}, the standard English
read-speech corpus derived from LibriVox audiobooks and sampled at 16\,kHz.  Training uses the
full 960\,h partition (\texttt{train-clean-100}, \texttt{train-clean-360},
\texttt{train-other-500}), $280{,}663$ utterances totalling $958.2$\,h after dropping
utterances longer than $17$\,s.  \texttt{dev-clean} is used for validation and model selection,
and \texttt{test-clean} and \texttt{test-other} are used for the reported word error rates.

Unlike the usual offline protocol, which presents a whole pre-segmented utterance to the model
at once, we stream each utterance to the model in fixed-size chunks and require a prediction
after every chunk, so that the model never sees future audio beyond its own bounded look-ahead.

\subsubsection{Licenses}

LibriSpeech is distributed under the Creative Commons Attribution 4.0 License, and is used here
for non-commercial academic research.

\subsection{Model}

The reference ANN is a windowed SummaryMixing Conformer encoder~\citep{parcollet2024summarymixing}
with a CTC head, adapted so that the whole encoder is a composition of affine operators and
$\text{ReLU1}$.  SummaryMixing is the natural starting point for a convertible streaming
encoder: it replaces self-attention's data-dependent $T\times T$ mixing with a \emph{constant}
one -- a uniform mean over time -- which is an affine operator, whereas softmax attention is not.

\paragraph{Front-end.} 80-bin log-mel filterbank features at a 10\,ms hop, normalized with a
\emph{fixed global} CMVN fitted once on the training set and frozen (per-utterance CMVN is
neither causal nor foldable into the network), followed by a convolutional subsampling of
$\times 4$, giving one encoder frame every 40\,ms.

\paragraph{Window assembly.} The encoder consumes one window per call, formed by concatenating
the previous chunk and the current chunk, $[\,\text{prior chunk} \,\Vert\, \text{current
chunk}\,]$.  This is the design decision that keeps the encoder feed-forward: the published
streaming SummaryMixing variant accumulates its summary statistic \emph{across} calls, which is
a cross-call recurrence, whereas taking the mean over a two-chunk window makes the same summary
a reduction \emph{inside} a single forward pass -- an ordinary \texttt{AvgPool1d}.  With a chunk
of $32$ encoder frames and a left context of $32$ frames, the window spans $2.56$\,s and the
model emits a $1.28$\,s chunk of frame posteriors per call.  The first window of a stream is
zero-padded on the left and its mean is renormalized accordingly.

\paragraph{Block.} Each of the 18 encoder blocks applies two per-frame branches
$f(X)$ and $s(X)$ (Linear\,+\,ReLU1), averages $s(X)$ over the window with a single
\texttt{AvgPool1d} whose kernel equals the window length, broadcasts the resulting summary back
to every frame, concatenates it with $f(X)$, and projects with a further Linear\,+\,ReLU1.  A
depthwise convolution module (kernel 31 or 63) and a feed-forward sandwich complete the block,
as in a Conformer.  The model dimension is $256$; the encoder has $\approx\!25$--$28$M
parameters and $128$--$182$ spiking sites depending on the arm.

\paragraph{Substitutions for convertibility.} GELU/Swish are replaced by $\text{ReLU1}$;
LayerNorm by masked BatchNorm, since LayerNorm computes per-sample statistics at run time and
therefore cannot be folded into the preceding affine layer; and the streaming variant's
cumulative masked mean by the windowed \texttt{AvgPool1d} described above.  The head is CTC, as
in the source recipe; CTC collapse and beam search run on the decoded posteriors, outside the
converted graph.

\subsection{Training Setup}

The encoder is trained with CTC on the 960\,h partition for $40$--$53$ epochs, with Adam
(peak lr $8\times10^{-4}$, $5$k--$15$k warm-up steps, then inverse-square-root decay), dropout
$0.1$, weight decay $5\times10^{-4}$, gradient clipping, and mixed precision.  The effective
batch size is $14$ utterances $\times$ gradient accumulation, with a length-bucketed sampler.
Augmentation is speed perturbation plus DropFreq/DropChunk on the waveform, and SpecAugment on
the features where indicated.  We keep an EMA of the weights (decay $0.9999$) and average the
final $10$ epoch checkpoints, as is standard for this recipe.  Two vocabularies are used: a
29-symbol character set and a $5000$-token unigram sub-word model; and some arms add an
auxiliary attention decoder to the CTC loss during training only.  \emph{The auxiliary decoder
is discarded before conversion and is never part of the deployed or converted model}; it acts
purely as a training-time regularizer.  Training goes through exactly the same window assembly
that inference uses, so the model is never trained on context it will not have at deployment;
this equivalence is asserted by a unit test on every commit.

\subsection{Evaluation Setup}

\paragraph{Streaming protocol.} Each utterance is streamed chunk by chunk with 40ms inter-chunk interval.  For \modelname{},
the converted SNN's state is initialized once at the beginning of the utterance and then
persists across chunk boundaries; each chunk is allotted $T$ ticks and the frame posteriors for
that chunk are read at its final tick.  The first chunk optionally receives a warm-up of extra
ticks.  For the reset baselines the state is cleared at every chunk boundary, which is the
protocol those methods assume.  We use $\tau = 8$ throughout, as on the other two tasks.

\paragraph{Decoding and metric.} Frame posteriors are decoded with CTC greedy decoding, or with
a CTC prefix beam search combined with an $n$-gram language model where stated; decoding is
identical for every arm, so it never contributes to a difference between arms.  We report
corpus-level word error rate (WER) on \texttt{test-clean} and \texttt{test-other}.  

\section{Gesture Recognition on Event Streams}
\label{app:gesture_eval}

\subsection{Datasets}

We use DVS128 Gesture~\citep{amir2017lowpower}, the standard neuromorphic gesture benchmark,
recorded with a DAVIS128 event camera at $128\times128$ spatial resolution.  It contains
$1{,}342$ gesture instances over $11$ classes (10 gestures plus an \emph{other} class),
performed by $29$ subjects under three lighting conditions, and distributed as $122$
\emph{continuous} recordings with interval labels.  We use the official subject split: $98$
recordings for training and $24$ for test.

\paragraph{Streaming, not clip-wise.} Most recent work evaluates this dataset clip-wise: each
pre-segmented gesture is cut out in advance, binned by constant event count, and classified as
a whole.  That protocol is not causal -- constant-count binning needs the end of the clip to
place its bins, and cutting the clip out presupposes that the gesture has already been
segmented.  We therefore evaluate on the \emph{unsegmented} recordings, as the original
benchmark does~\citep{amir2017lowpower}, and require the model to predict on the fly from the
events it has received so far.  Concretely, we slide a $2$\,s window over each recording and
score every window position.  

A $2$\,s window is the geometry that follows from the data.  Gestures last $6.53$\,s on average
and the gaps between them $1.84$\,s, so a window as long as a gesture straddles two gestures in
over half of all positions, whereas a $2$\,s window is clean in $56.3\%$ of positions and spans
two gestures in only $7.4\%$.  Because the gestures are repetitive, a $2$\,s window still
contains complete motion cycles.  Since a streaming evaluation also encounters rest periods, the
stream task carries a $12$th \emph{null} class. 

\subsubsection{Licenses}

The DVS128 Gesture dataset is released by IBM Research under the Creative Commons Attribution
4.0 License.  We use the distributed archive as published, verified by its MD5 checksum, and
apply only the corrected label file for one recording released with the dataset mirror (one
gesture's end time is recorded after the next gesture's start time; the corrected file resolves
the overlap).  The dataset is used for non-commercial academic research.

\subsection{Model}

The reference ANN is \emph{GestureNet}, a convertible TCN-style network of $3.96$M parameters
with $22$ spiking sites:

\begin{center}
\begin{tabular}{@{}ll@{}}
\toprule
stage & operator \\
\midrule
input            & $[B, T, C, 128, 128]$ event histogram \\
stem             & $5\times5$ conv, stride 2 $\rightarrow$ BN $\rightarrow$ ReLU1 \\
spatial backbone & 4 stages of residual blocks, $\text{AvgPool2d}(2)$ between stages \\
spatial pooling  & $\text{AvgPool2d}$ with a kernel fixed at build time \\
temporal         & 2 \texttt{Conv1d} blocks over the frame axis \\
readout          & $\text{AvgPool1d}(T)$ $\rightarrow$ Linear \\
\bottomrule
\end{tabular}
\end{center}

Two design decisions carry the model.  First, the spatial backbone is applied frame-wise with
shared weights, with $T$ folded into the batch dimension, so the number of spiking sites is
$O(\text{depth})$ rather than $O(\text{depth}\times T)$ -- which is what makes a multi-frame
window affordable at a small tick budget.  Second, the time axis is mixed by convolutions rather
than by a mean: four of the eleven classes are chirality pairs (left/right hand, clockwise vs.\
counter-clockwise) that differ only in temporal order, and any order-invariant reduction maps a
gesture and its time-reverse to the same representation.  $\text{AvgPool1d}$ is retained, but
only after the temporal convolutions have made the representation order-sensitive.  The head is
a bare linear layer with no terminal activation.  The same constraint rules out horizontal flip
and temporal reversal as augmentations: both map a class onto its partner's label.

\paragraph{Input representation.} Events are accumulated into a voxel grid, without
normalization or clamping.  For each event $(t,x,y,p)$ falling in the window, the bin
index is $\lfloor (t - t_{\text{start}})/\Delta \rfloor$ and the corresponding cell
$[\text{bin}, p, y, x]$ is incremented.  The $2$\,s window is divided into $16$ equal bins
of $125$\,ms, giving $[16, 2, 128, 128]$ for the two event polarities.  Beside those
disjoint bins we carry a running mean of them,
$x_{\text{cum}}[k] = \frac{1}{k+1}\sum_{j \le k} x[j]$, concatenated along the channel axis
 to give $[16, 4, 128, 128]$.  The two halves hold different things and are kept together
rather than one replacing the other: the disjoint bins carry instantaneous ON/OFF polarity
structure, which is what separates the rotation classes, while accumulation greys that
structure out as both polarities pile onto the same pixels; the running mean in turn
 carries the longer-range shape of the gesture, which no single $125$\,ms bin contains.
The cumulative sum is an invertible linear map along the frame axis, so nothing is lost
relative to the disjoint bins and the traced graph stays affine\,+\,ReLU1 -- the only
structural change is that the stem takes $4$ input channels instead of $2$.

\subsection{Training Setup}

The model is trained on $2$\,s windows drawn from the training recordings at random offsets, so
that every training window is distinct.  We use AdamW with a OneCycle schedule, label smoothing
$0.1$, dropout $0.1$, gradient clipping at $1.0$, and standard BatchNorm initialization, for
$200$ epochs.  Augmentation is random spatial roll and rotation only, for the chirality reason
above.  We keep an EMA of the weights and evaluate both the raw and the EMA checkpoints.  The
reported configuration is run with $3$ seeds; the measured seed spread is $0.51$\,pp, below the
eval-to-eval oscillation within a single run.  Since one test clip is $0.347$\,pp on the
$288$-clip protocol, we treat differences below $\sim\!1.4$\,pp as not meaningful and do not
draw conclusions from them.

\subsection{Evaluation Setup}

\paragraph{Streaming protocol.} Each test recording is streamed as a sequence of $2$\,s windows
at a $0.25$\,s stride.  For \modelname{}, the converted SNN's state is initialized once per
recording and persists across windows, with $T$ ticks per window; the prediction is read at the
final tick of each window.  Reset baselines clear their state at every window.  A window is
labelled by the class at its center; windows falling in a rest period take the \emph{null}
label.  We use $\tau = 8$, as on the other two tasks.

\paragraph{Metrics.} The primary metric is per-window classification accuracy on the stream.  We additionally
report two protocols that connect to the literature.  (i) \emph{Per-decision accuracy in the
style of the original benchmark}~\citep{amir2017lowpower}: decisions are taken on a sliding
window at a $0.1$\,s stride at every position inside a labelled gesture, the null column is
dropped, and the arg-max is taken over the $11$ gesture classes, with an optional majority-vote
horizon standing in for their sliding-window filter; we also report the $10$-category variant
that excludes \emph{other}, mirroring their own two numbers.  (ii) \emph{Clip accuracy} on the
standard $288$-clip test protocol, obtained by sliding the same $2$\,s windows inside each test
clip, averaging the logits and taking the arg-max over classes $1$--$11$; this is the number
comparable to the published leaderboard.  Efficiency is reported as in
Appendix~\ref{app:baselines:protocol}.

\section{Experimental Setups for the In-depth Analyses}
\label{app:indepth}

All four studies in the in-depth analysis are run on Argoverse-HD with the detector and the streaming
protocol described in Appendix~\ref{app:obj_det_eval}, at $640$ input resolution and $\tau=8$.  Unless
stated otherwise they use the full validation split ($24$ sequences, $15{,}062$ frames) and
never stride the stream, since the inter-frame interval is the independent variable in two of
the four studies.

\subsection{Integration with IF conversion methods}
\label{app:indepth:porting}

\paragraph{Question.} Our framework and the advanced IF conversion methods optimize different
things -- ours makes the trajectory track a moving reference across frames, theirs makes a
single bounded simulation more accurate -- so it is worth asking whether the two compose.

\paragraph{Porting.} For each published conversion method we take the converted network exactly
as that method produces it, using the same implementation that produces its row in
Table~\ref{tab:snn_comparison_revised}, and run it under two protocols that differ by exactly one
call: the method's own per-frame reset, and our streaming protocol in which the state is
initialized once per sequence and then persists.  Nothing else changes -- same weights, same
calibration, same decoder, same frames, same score threshold -- so any difference is attributable
to the reset alone.  Methods that pre-charge each neuron to a fixed initial potential
(e.g.\ $\theta/2$ for QCFS and SlipReLU) store that value as the neuron's \emph{reset} value, so
it is re-applied by the reset call and only by the reset call; AdaFire likewise clears its burst
state per frame.  

\paragraph{Measurement.} All arms are scored over one identical frame set -- whole sequences,
never strided -- with AP@50 at a score threshold of $0.001$, and sweep $T \in \{8,16,32,64,128\}$
(plus $T=256$ for the IF arm, whose published protocol needs a far larger budget).  The dense
ANN on the same frames provides the ceiling.  Because the per-frame-reset baselines and the
streaming arm are scored on the same frames, the ratios are meaningful even where the absolute
values differ from the full-validation numbers of
Table~\ref{tab:snn_comparison_revised}.  Some of these baselines require a source ANN trained
with their own activation, or an unbounded-ReLU network plus calibration data, so the comparison
necessarily crosses checkpoints; this is structural rather than a choice, and is itself part of
the point, since a bounded activation needs nothing fitted.

\paragraph{Is the IF failure a decoder problem?} The streaming arm above reads each frame as
the mean of its own $T$ ticks, a tumbling window.  To separate the neuron from the decoder we keep
each converted network, frames and protocol fixed and change only the decoder of the output layer:
an exponential moving average with the same $\tau = 8$ as LIF-Stream, never reset, and a sliding
window of the last $16$ ticks across frame boundaries.  We do this for four IF conversions: QCFS and
SlipReLU, which train their own source ANN; ReLU-Norm~\citep{rueckauer2017conversion}, which
normalizes an unbounded-ReLU ANN (99.9th percentile on $32$ calibration frames) and converts it to
soft-reset IF neurons with threshold $1$; and AdaFire, whose neuron may emit up to four spikes per
tick.  We also add the mechanism of \citet{kugele2020efficient}, a lower bound $u \ge 0$ on every IF
membrane, applied after each tick.  They apply it to a plain IF conversion, which corresponds to
our ReLU-Norm row.

\begin{center}
\small
\begin{tabular}{@{}llccc@{}}
\toprule
\multicolumn{2}{@{}l}{Argoverse-HD seqs.\ 5, 19, 10 (1{,}374 frames), AP50} & $T=4$ & $T=8$ & $T=16$ \\
\midrule
QCFS & per-frame reset (published protocol) & 0.7 & 1.7 & 4.6 \\
     & no reset, tumbling-window decoder & 0.8 & 0.7 & 0.7 \\
     & no reset, EMA decoder ($\tau = 8$) & 0.6 & 0.7 & 0.7 \\
     & no reset, sliding-window decoder ($16$ ticks) & 0.6 & 0.6 & 0.7 \\
\midrule
SlipReLU & per-frame reset (published protocol) & 0.4 & 0.4 & 1.1 \\
     & no reset, tumbling-window decoder & 0.2 & 0.3 & 0.5 \\
     & no reset, EMA decoder ($\tau = 8$) & 0.2 & 0.3 & 0.5 \\
     & no reset, sliding-window decoder ($16$ ticks) & 0.2 & 0.4 & 0.5 \\
\midrule
ReLU-Norm & per-frame reset (published protocol) & 0.0 & 0.0 & 0.0 \\
     & no reset, tumbling-window decoder & 0.0 & 0.0 & 0.0 \\
     & no reset, EMA decoder ($\tau = 8$) & 0.0 & 0.0 & 0.0 \\
     & no reset, sliding-window decoder ($16$ ticks) & 0.0 & 0.0 & 0.0 \\
\midrule
AdaFire & per-frame reset (published protocol) & 0.0 & 0.0 & 0.4 \\
(multi-spike) & no reset, tumbling-window decoder & 0.5 & 0.7 & 0.8 \\
     & no reset, EMA decoder ($\tau = 8$) & 0.4 & 0.6 & 0.8 \\
     & no reset, sliding-window decoder ($16$ ticks) & 0.4 & 0.6 & 0.8 \\
\midrule
\multicolumn{5}{@{}l}{\emph{Membrane bound $u \ge 0$ as in \citet{kugele2020efficient}, no reset, EMA decoder}} \\
ReLU-Norm & plain IF conversion, the setting of \citet{kugele2020efficient} & 0.0 & 0.0 & 0.0 \\
\midrule
\multicolumn{2}{@{}l}{LIF-Stream (ours)} & \textbf{12.1} & \textbf{18.4} & \textbf{22.9} \\
\bottomrule
\end{tabular}
\end{center}

With every conversion the three decoders agree within $0.2$ AP50 and stay below $1$ AP50: the
decoder is not what fails.  On average, $45$--$50\%$ of the IF membranes are below zero at a tick
and are clipped by the bound, so this charge debt is the normal state of these networks rather
than a corner case (Appendix~\ref{pf:if_memory}).  On the plain IF conversion that \citet{kugele2020efficient}
use, the bound does not help at these budgets: ReLU-Norm stays at $0$ even with a per-frame reset,
since it needs hundreds of ticks per frame (Table~\ref{tab:snn_comparison_revised}), whereas
\citet{kugele2020efficient} use $15$--$60$ simulation steps per frame.  At any budget the bound is
also not a principled fix: it discards charge whenever the input dips below it, which biases the
firing rates away from the ReLU1 correspondence that the conversion relies on (Remark~\ref{rem:if_bound}).

\subsection{Inter-frame prediction efficiency}
\label{app:indepth:frame_ticks}

\paragraph{Question.} A uniform budget charges every frame the same $T$.  If frames differ in how
many ticks they actually need, the distribution of the \emph{needed} budget is the quantity the
efficiency claim should be stated in, and the gap between it and a flat $T$ is what incremental
computing buys.

\paragraph{Defining ``settled''.} The settle tick of frame $k$ is measured on the \emph{detections},
not on the raw readout:
\begin{equation}
A_k[t] = F_1\big(\mathrm{dets}(y_t),\, \mathrm{dets}(y_{T_{\max}})\big),
\qquad
\mathrm{settle}(k) = \min\{\,t : A_k[t'] \ge 1-\epsilon \text{ for all } t' \ge t\,\},
\end{equation}
where detections are matched by class at IoU $\ge 0.3$ and $F_1$ is computed over detections
above a confidence of $0.3$.  This is what ``the answer has stopped changing'' means for a
detector, and it is insensitive to the slow residual drift that a threshold on the float readout
would pick up.  The stability threshold ($0.3$) is deliberately distinct from the scoring
threshold ($0.001$, the COCO convention): at the scoring threshold the output carries a long tail
of low-confidence boxes whose ordering never stabilizes.  
$T_{\max}=64$ is used as the reference budget.

\paragraph{Complexity label.} The per-frame complexity on the $x$-axis is computed
\emph{independently of the tick count}, otherwise the plot would be a tautology.  We use the
dense ANN's own frame-to-frame output change,
$\mathrm{drift}(k) = \Vert y_{\text{ANN}}(k) - y_{\text{ANN}}(k{-}1)\Vert /
\Vert y_{\text{ANN}}(k{-}1)\Vert$, measured in the same flattened head space; the ANN arm is run
anyway as the accuracy ceiling, so this costs nothing extra.  Raw pixel change is reported as a
secondary, interpretability-only label.

\subsection{Finer temporal resolution helps streaming SNNs}
\label{app:indepth:time_resolution}

\paragraph{Question.} Tick cost is normally quoted per frame, which silently treats the frame rate
as given.  It is a design choice, and the two protocols respond to it in opposite ways: with a
per-frame reset every frame pays the full transient, so cost per second is $T \times \text{fps}$,
exactly linear in the frame rate with no property of the video entering; with a persisting state a
higher frame rate means a smaller change between consecutive frames, so each frame needs fewer
ticks.

\paragraph{Setup.} We re-use the settle-tick definition of Appendix~\ref{app:indepth:frame_ticks}
unchanged -- same detection-space criterion, same matching threshold, same $\epsilon$ grid -- and
sweep the sampling rate by striding the source video: strides $1,2,3,4,5,6,10$ of the native
30\,fps stream give $30, 15, 10, 7.5, 6, 5, 3$\,fps.  Each point is the full validation split at
that rate.  Cost per second of video is $\text{fps} \times \overline{\mathrm{settle}}(\text{fps})$.
The reset arm needs no run: its frames are independent, so its cost is $T \times \text{fps}$
analytically, and we use the tick budget at which it attains the same accuracy as the streaming
arm on the same data.  Striding is what the streaming protocol otherwise forbids, precisely
because it widens the inter-frame gap and removes the redundancy persistence exploits; here that
is the independent variable.  Consequently the numbers in this study are comparable only across
its own sweep, the 30\,fps column being the one it shares with the main streaming results.

\paragraph{Reporting.} We fit a power law to cost per second against frame rate for each arm and
report the exponent, together with the censored fraction at every point; points whose censoring is
large enough to understate the mean are excluded from the figure rather than reported with a
caveat.  The claim the measurement supports is the ratio of the two exponents -- reset grows
linearly with frame rate by construction, while the streaming arm grows sublinearly -- so the
advantage widens as the stream is sampled more finely.

\subsection{Qualitative visualization of the inference trajectory}
\label{app:indepth:qualitative}

For Figure~\ref{fig:qualitative} we project the dense ANN's per-frame outputs onto their own top
two principal directions and place every SNN tick of every arm on that same basis, so that all
arms are shown in one coordinate system defined by the reference model rather than by any of the
spiking arms.  The projection is computed on a single Argoverse-HD sequence, with the detector
and protocol of Appendix~\ref{app:obj_det_eval}; distances quoted in the text are measured in that
2-D projection between an arm's final tick for a frame and that frame's ANN target.  The arms
shown are the published IF conversion protocol, the LIF conversion under a per-frame reset, and
our \modelname{}, each at its own tick budget, so that the figure shows what a given budget buys
rather than a budget-matched comparison alone.

\subsection{Synaptic operations and energy at matched accuracy}
\label{app:indepth:ops}

Ticks are the unit of Table~\ref{tab:snn_comparison_revised}, but a tick does not cost the same
for every arm.  We therefore count synaptic operations per frame with the counter of the DVS leg,
unchanged: a layer driven by spikes pays one accumulate (AC) per spike and synapse, a layer driven
by analog values (the stem) pays a multiply-accumulate (MAC).  MACs arise only in the stem, because the float encoder feeds each frame to it as analog values at every tick (App.~\ref{app:baselines}); they scale with $T$ and account for $3$--$5\%$ of the energy in every row.  A spike-based stream encoder removes them, since the stem then also receives binary spikes and pays ACs instead; this costs little accuracy, as the population-lookahead spike encoder stays within $0.4$ AP50 of its float counterpart at every tick budget (Appendix~\ref{app:encoder_comparison}).  Counts are averaged over the first
$60$ consecutive frames of Argoverse-HD sequence $5$; for a reset arm one frame is its full $T$-tick
simulation, for the streaming arm the sequence is run once and divided by the number of frames.
Energy uses the common $45$\,nm figures, $4.6$\,pJ per MAC and $0.9$\,pJ per AC,
which are an assumption about hardware rather than a measurement.  AP50 is from the three
sequences of Appendix~\ref{app:indepth:porting}.

\begin{center}
\small
\begin{tabular}{@{}lccccc@{}}
\toprule
Arm & $T$ & AP50 & AC (G) & MAC (G) & energy (mJ) \\
\midrule
\modelname{} (ours) & 8 & 18.4 & 128 & 1.1 & 120 \\
\modelname{} (ours) & 16 & 22.9 & 257 & 2.1 & 241 \\
\modelname{} (ours) & 32 & 25.9 & 515 & 4.2 & 483 \\
LIF, per-frame reset & 64 & 24.8 & 1069 & 8.5 & 1001 \\
QCFS, per-frame reset & 64 & 15.7 & 828 & 8.5 & 784 \\
QCFS, per-frame reset & 128 & 22.1 & 1631 & 17.0 & 1546 \\
SlipReLU, per-frame reset & 64 & 10.8 & 845 & 8.5 & 799 \\
SlipReLU, per-frame reset & 128 & 19.0 & 1679 & 17.0 & 1590 \\
AdaFire, per-frame reset & 64 & 22.2 & 1599 & 8.5 & 1478 \\
AdaFire, per-frame reset & 128 & 29.4 & 3243 & 17.0 & 2997 \\
\bottomrule
\end{tabular}
\end{center}

At matched accuracy, \modelname{} at $T=16$ uses $6.1\times$ less energy than AdaFire, the
strongest baseline, at $T=64$, and $6.4\times$ less than QCFS at $T=128$; it uses $6.6\times$ less than SlipReLU at $T=128$ while being more accurate ($22.9$ vs.\ $19.0$); at $T=8$ it comes within $0.6$ AP50 of SlipReLU at
$T=128$ with $13\times$ less energy; and at $T=32$ it uses $2.1\times$ less than the same LIF network with a
per-frame reset at $T=64$.  The saving comes from the smaller tick budget; AdaFire's burst spikes are
counted as multiple events.  At $T=128$ AdaFire reaches $29.4$ AP50, above the $25.9$ of
\modelname{} at $T=32$, at six times its energy.

\subsection{Sensitivity to the leak $\tau$}
\label{app:indepth:tau}

$\tau$ sets the step size $\frac{1}{\tau}$ of Lemma~\ref{thm:dynamics_lif}.  We use $\tau = 8$ for every
task and budget up to $64$ ticks.  The sweep below repeats the streaming arm of
Appendix~\ref{app:indepth:porting} (same detector and frames) with $\tau \in \{4, 8, 16\}$.

\begin{center}
\small
\begin{tabular}{@{}lccc@{}}
\toprule
AP50, Argoverse-HD seqs.\ 5, 19, 10 & $\tau=4$ & $\tau=8$ & $\tau=16$ \\
\midrule
$T = 8$ & 3.2 & \textbf{18.4} & 14.0 \\
$T = 16$ & 3.7 & \textbf{22.9} & 20.7 \\
\bottomrule
\end{tabular}
\end{center}

A larger $\tau$ moves the output more slowly, so with few ticks per frame it lags the target, and
the gap closes as $T$ grows ($14.0 \to 20.7$).  A smaller $\tau$ is not simply faster: at $\tau = 4$
accuracy collapses and does not recover with more ticks, which matches the full-validation run
($1.8$ AP50 at $T=8$).  Two effects may contribute, and we have not separated them: the $\frac{1}{\tau}$ band of
Corollary~\ref{cor:practical_stability} is then $0.25$, a quarter of the ReLU1 range, and $\tau$ also sets the
input gain of the conversion, $(\tau-1)z + 1$ (Appendix~\ref{app:baselines}), so $\tau=4$ is a
different value-to-rate mapping rather than a leakier $\tau=8$.  In either case $\tau$ is a real
hyperparameter.  We fixed $\tau = 8$ in advance, independently of the evaluation sets, and used it
for every task without tuning; the sweep confirms it as the best of the three values.

\subsection{Scene cuts, frame drops and input noise}
\label{app:indepth:robust}

The streams of Table~\ref{tab:snn_comparison_revised} are continuous recordings.  We stress the
streaming arm (Appendix~\ref{app:indepth:porting}: same detector, $\tau=8$, sequences 5, 19, 10)
in three ways, never resetting the state after the start of a stream.
\emph{Scene cuts}: the three sequences are concatenated into one stream, so the network meets two
abrupt cuts.  \emph{Frame drops}: every frame except the first is dropped independently with
probability $0.3$ (seeded) and the rest are streamed back to back; the arm is scored on the kept
frames and compared with the unperturbed stream on the same frames.  \emph{Noise}: i.i.d.\ Gaussian
noise of standard deviation $0.1$ (normalised input units) on every frame; the dense ANN is scored on
the same noisy frames.

\begin{center}
\small
\begin{tabular}{@{}lcc@{}}
\toprule
AP50 & $T=8$ & $T=16$ \\
\midrule
Each sequence its own stream (reference) & 18.4 & 22.9 \\
Scene cuts: one stream, no reset at the cuts & 18.3 & 22.9 \\
\quad first 10 frames after each cut: carried state / fresh start & 5.9 / 4.3 & 12.0 / 11.0 \\
Frame drops ($p=0.3$) / unperturbed stream on the same frames & 15.4 / 18.1 & 21.2 / 22.6 \\
Noise $\sigma=0.1$ (dense ANN: $31.0$, clean $32.5$) & 17.0 & 21.2 \\
\bottomrule
\end{tabular}
\end{center}

A cut costs nothing overall, and the frames right after it are recognised slightly \emph{better}
when the network starts from the previous scene's state than from a fresh one: as
Theorem~\ref{thm:equilibrium_lif} predicts, attraction holds from any state, so a stale state is no
worse a starting point than an empty one.  Dropping frames enlarges the jump between consecutive
frames, which raises the lag term of Corollary~\ref{cor:tracking}; accuracy falls by $15\%$ at $T=8$ and by
$6\%$ at $T=16$, where more ticks absorb the larger jumps.  Input noise costs the streaming SNN
$7$--$8\%$, close to the $5\%$ it costs the dense ANN.

\subsection{Empirical Validation of our theory in Figure~\ref{fig:snn_trajectory_visualization}}
\label{app:theory_validation_setup}
We use a small, fully controlled network. The reference ANN is an MLP of seven affine layers of width 64 (12-dimensional input, 16-dimensional output, 22.7K parameters) whose only nonlinearity is ReLU1. It is trained on a teacher–student task in which inputs $x$ and targets $y$ are generated from a shared two-dimensional latent $z$: the input is a random linear lift of the latent into 12 dimensions followed by a tanh squash, and the target is a bank of sinusoids of independent random projections of that same latent, each with its own phase. 
\begin{align}
\mathbf{z} &\sim \mathcal{N}\!\left(\mathbf{0},\, 0.8^{2}\mathbf{I}_2\right)
  && \text{shared latent} \\
\mathbf{x} &= 0.5 + 0.4\,\tanh\!\left(\mathbf{A}\mathbf{z}\right) \in (0.1, 0.9)^{12}
  && \text{network input} \\
\mathbf{y} &= 0.5 + 0.45\,\sin\!\left(\mathbf{W}\mathbf{z} + \boldsymbol{\varphi}\right) \in (0.05, 0.95)^{16}
  && \text{regression target}
\end{align}
where $A_{ij}\sim\mathcal{N}(0,\tfrac12)$, $W_{ij}\sim\mathcal{N}(0,1.6^2)$ and $\varphi_i\sim\mathcal{U}(0,2\pi)$ are sampled once and held fixed, and $\tanh$, $\sin$ act elementwise.
Training is 1.5K Adam steps (batch 256, learning rate $3\times10^{-3}$, fixed seed). The evaluation input stream is the latent traversing a Lissajous curve over 24 frames. We convert the network with LIF neurons and feed the input 32 ticks per frame in a single streaming pass, and membrane state persists across the entire 768-tick stream. All distances are measured in the full 16-dimensional readout space, with PCA used only for visualization.

\end{document}